%% file: main.tex
\pdfoutput=1

\documentclass[acmtog]{acmart}

\acmSubmissionID{1884}
\usepackage{booktabs}
\usepackage{multirow}
\usepackage{subfigure}
\usepackage{soul}
\usepackage{xcolor}
\usepackage[ruled]{algorithm2e}

\usepackage{comment}
\input{custom_commands}

\addeditor{antoine}{AG}{0.0, 0.5, 0.0}
\addeditor{diego}{DG}{0.0, 0.0, 0.8}
\addeditor{bingchen}{BG}{0.9, 0., 0.9}
\addeditor{nissim}{NM}{0.9, 0.4, 0.}
\addeditor{maks}{MO}{0.3, 0.4, 0.95}
\addeditor{george}{GD}{0.0, 0.4, 0.9}
\addeditor{todo}{TODO}{0.9, 0.0, 0.0}

\showeditsfalse

\SetAlFnt{\small}
\SetAlCapFnt{\small}
\SetAlCapNameFnt{\small}
\SetAlCapHSkip{0pt}

\acmJournal{TOG}

\acmJournal{TOG}
\acmYear{2025} \acmVolume{44} \acmNumber{6} \acmArticle{1} \acmMonth{12}\acmDOI{10.1145/3763339}

\begin{document}
\title{MILo: Mesh-In-the-Loop Gaussian Splatting for Detailed and Efficient Surface Reconstruction}


\author{Antoine Gu\'edon}
\email{antoine.guedon@enpc.fr}
\authornote{Both authors contributed equally to the paper.}

\author{Diego Gomez}
\authornotemark[1]
\affiliation{
\institution{\'Ecole Polytechnique}
\country{France}
}
\email{diego.gomez@polytechnique.edu}

\author{Nissim Maruani}
\affiliation{
\institution{Inria, Université Côte d'Azur}
\country{France}
}
\email{nissim.maruani@inria.fr}

\author{Bingchen Gong}
\affiliation{
\institution{\'Ecole Polytechnique}
\country{France}
}
\email{bingchen.gong@polytechnique.edu}

\author{George Drettakis}
\affiliation{
\institution{Inria, Université Côte d'Azur}
\country{France}
}
\email{George.Drettakis@inria.fr}

\author{Maks Ovsjanikov}
\affiliation{
\institution{\'Ecole Polytechnique}
\country{France}
}
\email{maks@lix.polytechnique.fr}

\input{sections/0_abstract}

%


%
%


\input{figures/teaser}

\maketitle

\input{sections/1_introduction}
\input{sections/2_related_work}
\input{sections/3_method}

\input{sections/4_experiments}
\input{sections/5_conclusions_and_limitations}

\bibliographystyle{ACM-Reference-Format}
\bibliography{main}

\input{sections/X_supplemental}
\end{document}

%% file: custom_commands.tex
\usepackage{dsfont}
\usepackage{etoolbox}
\usepackage{color}

\makeatletter
\DeclareRobustCommand*\cal{\@fontswitch\relax\mathcal}
\makeatother

\newif\ifshowedits

\newcommand{\addeditor}[3]{%
  \definecolor{#1color}{rgb}{#3}
  \expandafter\newcommand\csname #1\endcsname[1]{%
  \ifshowedits
    {\color{#1color} ##1}%
  \else
    {##1}%
  \fi
  }%
  \expandafter\newcommand\csname #1rmk\endcsname[1]{%
  \ifshowedits
    {\color{#1color} {\bf [#2: ##1]}}
  \fi
  }%
  \expandafter\newcommand\csname #1rpl\endcsname[2]{%
  \ifshowedits
    {\color{#1color} ##1 \sout{##2}}
  \else
    {##1}
  \fi
  }%
}

\newcommand{\createtextvar}[1]{
  \expandafter\newcommand\csname #1\endcsname{%
  {\text{#1}}
}%
}
\newcommand{\mycomment}[1]{}

\newcommand{\calL}{{\cal L}}

\newcommand{\IR}{{\mathds{R}}}

\newcommand{\vcomment}[1]{}

\usepackage{colortbl}

\definecolor{yellow}{rgb}{1, 1, 0.7}
\definecolor{orange}{rgb}{1, 0.85, 0.7}
\definecolor{tablered}{rgb}{1, 0.7, 0.7}
\definecolor{red}{rgb}{1, 0, 0}

\definecolor{tablethree}{rgb}{0.7, 1, 1}
\definecolor{tabletwo}{rgb}{0.7, 0.85, 1}
\definecolor{tableone}{rgb}{0.7, 0.7, 1}

\newcommand{\best}{\cellcolor{tablered}}
\newcommand{\sbest}{\cellcolor{orange}}
\newcommand{\tbest}{\cellcolor{yellow}}

\newcommand{\bestbis}{\cellcolor{tableone}}
\newcommand{\sbestbis}{\cellcolor{tabletwo}}
\newcommand{\tbestbis}{\cellcolor{tablethree}}

\newcommand{\method}{MILo}

%% file: sections/0_abstract.tex
\begin{abstract}

While recent advances in Gaussian Splatting have enabled fast reconstruction of high-quality 3D scenes from images, extracting accurate surface meshes remains a challenge. Current approaches extract the surface through costly post-processing steps, resulting in the loss of fine geometric details or requiring significant time and leading to very dense meshes with millions of vertices. More fundamentally, the \textit{a posteriori} conversion from a volumetric to a surface representation limits the ability of the final mesh to preserve all geometric structures captured during training.
We present \method, a novel Gaussian Splatting framework that bridges the gap between volumetric and surface representations by differentiably extracting a mesh from the 3D Gaussians. We design a fully differentiable procedure that constructs the mesh---including both vertex locations and connectivity---at every iteration directly from the parameters of the Gaussians, \emph{which are the only quantities optimized during training.}

Our method introduces three key technical contributions:
(1)~a bidirectional consistency framework ensuring both representations---Gaussians and the extracted mesh---capture the same underlying geometry during training;
(2)~an adaptive mesh extraction process performed at each training iteration, which uses Gaussians as differentiable pivots for Delaunay triangulation; 
(3)~a novel method for computing signed distance values from the 3D Gaussians that enables precise surface extraction while avoiding geometric erosion.  

Our approach can reconstruct complete scenes, including backgrounds, with state-of-the-art quality while requiring an order of magnitude fewer mesh vertices than previous methods. 

Due to their light weight and empty interior, our meshes are well suited for downstream applications such as physics simulations and animation.

The code for our approach and an online gallery are available at 
{\color[HTML]{037D99}
\url{https://anttwo.github.io/milo/}
}.

\end{abstract}

%% file: figures/teaser.tex
\begin{teaserfigure}
    \vspace{-0.3cm}
    \vspace{0.2cm}
    \includegraphics[width=\textwidth, trim={0 1.75cm 0 0cm}, clip]{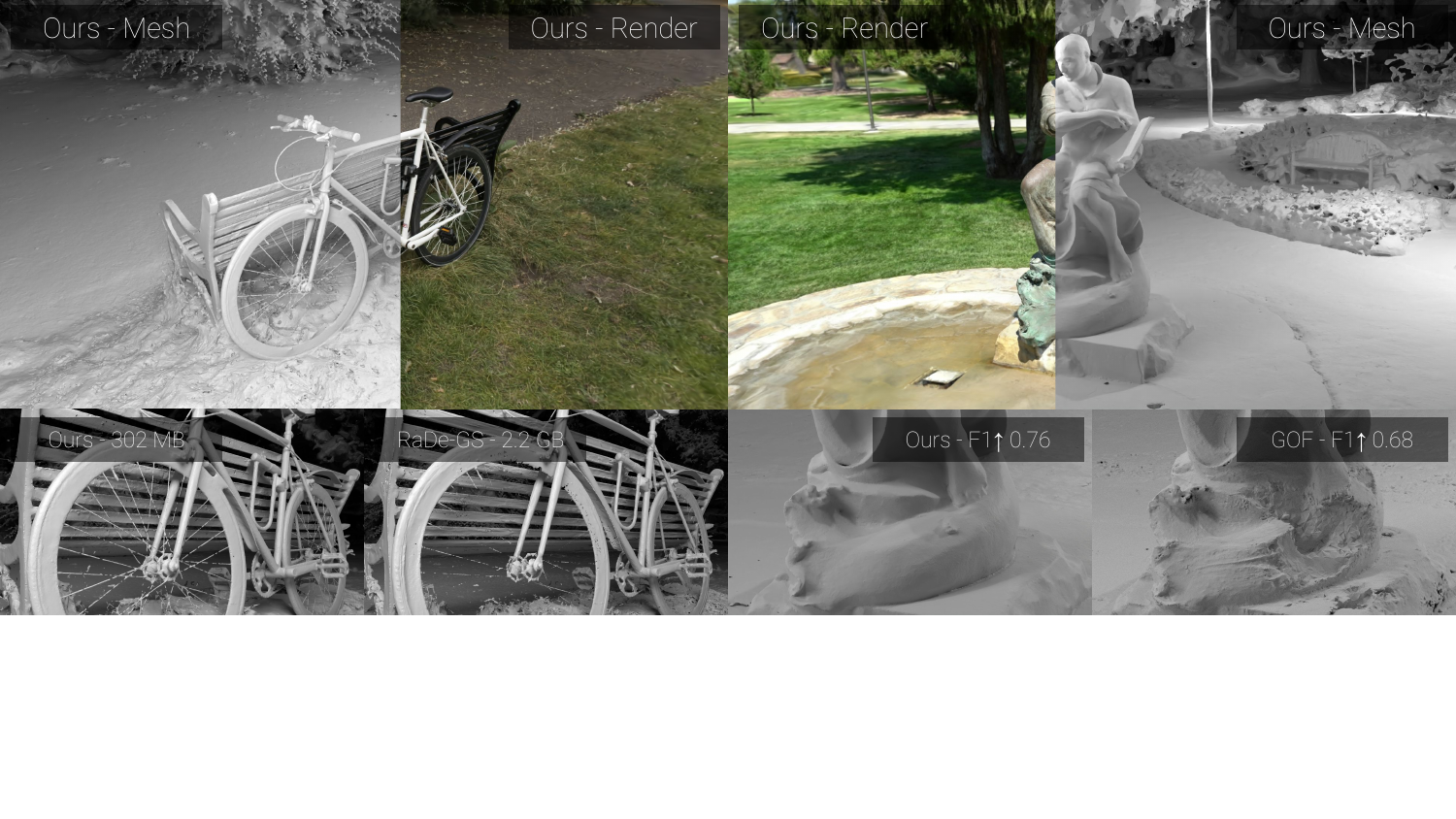}
    \vspace{-1.9cm}
    \caption{
        \textbf{Mesh-in-the-Loop Gaussian Splatting.} 
        Our method introduces a novel differentiable mesh extraction framework that operates during the optimization of 3D Gaussian Splatting representations. At every training iteration, we differentiably extract a mesh---including both vertex locations and connectivity---directly from Gaussian parameters. This enables gradient flow from the mesh to Gaussians, allowing us to promote bidirectional consistency between volumetric (Gaussians) and surface (extracted mesh) representations. This approach guides Gaussians toward configurations better suited for surface reconstruction, resulting in higher quality meshes with significantly fewer vertices. 
        In this example, our method reconstructs the entire bicycle scene---including background---with almost 10 times fewer vertices than previous methods while preserving fine geometric details.
        Our framework can be plugged into any Gaussian splatting representation, increasing performance while generating an order of magnitude fewer mesh vertices. \method~makes the reconstructions more practical for downstream applications like physics simulations and animation.
    }
    \label{fig:teaser}
\end{teaserfigure}

%% file: sections/1_introduction.tex
\section{Introduction}

Recent methods~\cite{li2023neuralangelo, Huang2DGS2024, Yu2024GOF, zhang2024rade, Reiser2024SIGGRAPH, guedon2024sugar,chen2023neusg} for reconstructing surface meshes from images commonly adopt a two-stage pipeline: Optimize a volumetric representation via differentiable rendering---typically using Neural Radiance Fields (NeRFs) or 3D Gaussian Splatting---then extract a surface mesh during \emph{postprocessing} by defining a surface, usually as an isosurface.  None of these methods considers the reconstructed mesh \emph{during optimization}.

This strategy offers no guarantee that the final mesh will be consistent with the volumetric representation---fine details \emph{that are present when rendering the representation} can be lost.

This postprocessing step collapses complex volumetric information into a surface, often leading to the loss of fine or semitransparent structures, and can introduce artifacts in the extracted mesh. Moreover, Gaussian Splatting and NeRF-based methods are known to adjust their opacity and view-dependent colors independently of the geometry. This allows them to fit the training images more precisely, but often at the expense of geometric consistency. \emph{``Cheating"} leads to hallucinated structures such as floaters or cavities, which are \emph{particularly hard to resolve during mesh extraction}. Fixing them is non-trivial, as the underlying volumetric representation has already absorbed these inconsistencies in its optimization. Several works have highlighted this issue in challenging settings—such as low-information~\cite{warburg2023nerfbusters, Yang2023FreeNeRF, gomez2025fourierffewshotnerfsprogressive,guedon2024matcha} or high specularities \cite{verbin2022refnerf}—yet a general and robust solution remains elusive.

Crucially, this process provides \textit{no feedback during optimization}, leaving mesh quality undefined until training is complete. This means that there is no guarantee that the optimized Gaussian representation will give rise to a high quality mesh capturing all relevant structures. For instance, fine details such as bicycle spokes can disappear or be inaccurately represented in the final mesh \textit{even if visually they exist in the volumetric representation}.

Moreover, many existing methods \cite{yariv2021volume_volsdf, wang2021neus, Huang2DGS2024, zhang2024quadratic} focus on object-centric reconstructions, evaluating only the quality of the foreground geometry. Similarly, benchmarks like DTU~\cite{jensen2014large} or Tanks and Temples~\cite{knapitsch2017tanks}  offer ground truth solely for foreground objects, discouraging full-scene reconstructions that have backgrounds. 

In this work, we introduce a novel pipeline for reconstructing compact, high-fidelity meshes of complete 3D scenes using Gaussian Splatting~\cite{kerbl3Dgaussians}, addressing all of the aforementioned limitations. Our core idea is to generate a mesh at every training iteration from a set of points entangled with the Gaussians, which we call \textbf{Gaussian Pivots}.

This setup enables direct gradient backpropagation from the mesh to the Gaussian parameters, allowing us to impose consistency between the surface mesh, volumetric field, and input data. Thanks to the Pivots, Gaussians serve as an implicit parameterization of an explicit mesh, progressively optimized to yield high-quality geometry, vertex positions, and connectivity.

As a result, our approach \textbf{sidesteps fixed-topology constraints}, given that the connectivity is dynamically updated as Gaussians move around the scene. On the other hand, the mesh provides a geometric prior that regularizes Gaussians to address \emph{``cheating"}, mitigating undesirable geometric artifacts. This tight, bidirectional coupling enables effective mesh-based regularization, enhancing both fidelity and compactness. In other words, \textbf{both representations help each other}.

To further control the complexity of outputs, we propose novel densification and simplification strategies inspired by~\cite{fang2024mini}. 

We achieve state-of-the-art geometric quality while using an order of magnitude fewer vertices than competing methods, making our surfaces far more practical for downstream applications. Inspired in previous works which focus on surface-based view synthesis methods \cite{Reiser2024SIGGRAPH} we use a proxy metric based on mesh rendering to evaluate mesh quality in background regions where no ground-truth geometry exists, training a neural color field over the mesh and comparing rendered images across methods. This decoupling rendering fidelity from mesh resolution and helps us to evaluate reconstruction quality across the entire scene. Our work makes the following contributions.
\begin{itemize}
    \item We introduce the \emph{first radiance field pipeline in which extracting a surface mesh is an integral part of the optimization}, leveraging expressive 3D Gaussian Splatting to parameterize and jointly refine both representations.
    \item We propose mesh-based regularization strategies that improve geometric quality, especially for thin structures.
    \item Our method achieves state-of-the-art results in mesh quality and compactness across multiple complex 3D scenes, improving upon previous approaches in both scalability and visual fidelity.
    \item \diego{We adapt a metric introduced in the context of surface-based view synthesis and use it as an evaluation protocol to assess full-scene geometry, even in the absence of ground-truth 3D models. We use this protocol to demonstrate that our method achieves state-of-the-art results in mesh quality and compactness across multiple complex 3D scenes,}
\end{itemize}

We provide detailed definitions of 3D Gaussian Splatting and Delaunay Triangulation in the supplementary material.

%% file: sections/2_related_work.tex
\section{Related Work}
\label{sec:related work}

\input{figures/architecture}

\paragraph{Novel View Synthesis.} Early novel view synthesis (NVS) techniques modeled scenes as continuous volumetric fields, notably with NeRF~\cite{mildenhall2020nerf} introducing a multilayer perceptron to learn a radiance field. Subsequent works explored alternative representations, including hash-based encodings~\cite{mueller2022instantngp}, discrete voxel grids~\cite{yu2021plenoxels}, and low-rank tensor decompositions~\cite{chen2022tensorf}, as well as solutions focused on antialiasing~\cite{barron2021mip, barron22mipnerf360, barron2023zipnerf}.

Recently, 3D Gaussian Splatting~\cite{kerbl3Dgaussians} has redefined the NVS landscape by enabling faster, high-quality synthesis from point-based primitives. This sparked a wave of follow-up work aiming to improve efficiency and scalability, including Mip-Splatting~\cite{yu2024mip}, Mini-Splatting v1 and v2~\cite{fang2024mini_v0, fang2024mini}, MCMC-Gaussians~\cite{kheradmand20243d}, BetaSplatting~\cite{liu2025deformable}, and Taming 3DGS~\cite{mallick2024taming}. Additional work like 3DGUT~\cite{wu20243dgut} and 3DGRT~\cite{moenne20243d} extend these models to arbitrary camera systems and differentiable ray tracing, respectively.

Despite their impressive rendering performance, none of these methods aim at explicit surface mesh reconstruction, limiting their applicability in tasks requiring geometric reasoning or downstream editing. \\

\vspace*{-4mm}
\paragraph{Surface reconstruction from images.} Recovering explicit surface meshes exclusively from images is a longstanding challenge. \diego{Well-established} volumetric approaches relying on implicit functions~\cite{wang2021neus, yariv2021volume_volsdf, li2023neuralangelo} proved efficient for reconstructing accurate surfaces. \diego{Training an implicit function is a strong regularization that allows to obtain smooth surfaces, thus these methods excel at extracting the foreground of object-centric scenes.}

\diego{However, these methods generally require very long training times (often exceeding 24 hours) before mesh extraction can be performed. The success of Gaussian Splatting has naturally inspired an emerging line of work that trains splats jointly with implicit functions, notably Gaussian-UDF \cite{li2025gaussianudf}, GS-Pull \cite{zhang2024gspull}, and GSDF \cite{yu2024gsdf}. By combining these two representations, such methods mitigate the time bottleneck by querying the implicit function through strategies that exploit the positions of the Gaussian splats. While this improves training efficiency, it also \textbf{exacerbates scalability issues}, since two representations must be optimized simultaneously. Moreover, although neural SDFs excel at reconstructing isolated objects, their expressivity remains bounded by the MLP architecture \cite{li2025gaussianudf,zhang2024gspull,yu2024gsdf}. Consequently, while these approaches can in principle be applied to more complex scenes, they tend to underperform beyond single-object or foreground-dominant settings.}

Moreover, these volumetric representations typically cannot achieve real-time rendering, limiting their practical applications. Some approaches address this by first optimizing the volumetric representation, extracting a mesh, and then baking the rendering capabilities into the mesh to enable real-time rendering~\cite{yariv23baked, Reiser2024SIGGRAPH}. \diego{Thus, these methods are split into two stages, which \textbf{prevents any refinement of the mesh}, while still exhibiting lengthy optimization times.}

Other methods based on discrete structures~\cite{munkberg22nvdiffrec, shen23flexicubes} also proved efficient to some extent, but focus on individual objects and \textbf{are not scalable to large complex scenes}.

\diego{Recent methods} relying on Gaussian Splatting offer promising \diego{efficient} alternatives \diego{\cite{wang2023neus2,dai2024high} namely}: 2DGS~\cite{huang20242d}, RaDe-GS~\cite{zhang2024rade}, NeuSG~\cite{chen2023neusg}, VCR-GauS~\cite{chen2024vcr}, and Quadratic Gaussian Splatting~\cite{zhang2024quadratic} explore different rendering or regularization strategies that rely on Truncated Signed Distance Fields (TSDF), while SuGaR~\cite{guedon2024sugar} and Gaussian Opacity Fields~\cite{yu2024gaussian} propose their own mesh extraction procedures.

\diego{All previously highlighted methods perform mesh extraction \emph{after} optimization, treating it as a post-processing step. Some of these go slightly further, adopting} a two-stage approach where they first optimize a volumetric representation, extract a mesh, and then refine it with differentiable rendering~\cite{yariv23baked,guedon2024sugar,guedon2024gaussianfrosting}. \diego{This line of work, however, neither avoids the initial extraction issues \textbf{nor adjusts the mesh topology}, which stays fixed during the refinement process}. This separation, \diego{in both cases}, introduces potential inconsistencies: \textbf{there is no guarantee that the extracted surface accurately reflects the underlying volumetric representation}. Since Gaussians are continuous by nature, naive isosurfacing often results in geometric artifacts such as over-inflation or erosion. These artifacts are particularly noticeable around thin structures (see Fig.~\ref{fig:teaser}). Moreover, \diego{all of these} methods typically produce overly dense meshes (up to tens of millions of vertices) that are difficult to scale to large scenes.

\diego{In contrast, our method efficiently leverages Gaussians to scale to full-scene reconstruction, and is the \textbf{first to integrate mesh extraction directly into the optimization loop}. This design naturally enables the refinement of both vertex positions and mesh topology, ensuring consistency with the underlying volumetric representation (Gaussian splats)}.

\vspace*{-3mm}
\paragraph{Voronoi $\&$ Delaunay-based methods.} Our work also exploits the Voronoi diagram and specifically its dual the Delaunay triangulation, which are classical geometric constructions with deep theoretical and practical relevance~\cite{aurenhammer1991voronoi}. These structures have long been applied to surface reconstruction problems~\cite{amenta_new_1998, amenta_power_2001, dey_tight_2003}, with explicit guarantees under specific sampling conditions. More recently, they have also gained traction in machine learning contexts: in 2D vision~\cite{williams_voronoinet_2020}, 3D geometry~\cite{maruani_voromesh_2023, maruani_ponq_2024}, or even novel view synthesis~\cite{elsner_adaptive_2023, govindarajan_radiant_2025}. While GOF~\cite{Yu2024GOF} applies Delaunay triangulation as a post-processing step for mesh extraction, Radiant Foam~\cite{govindarajan_radiant_2025} incorporates it into its training pipeline, but focuses solely on view synthesis and not surface reconstruction. In contrast, \emph{our work is the first to leverage Delaunay triangulation in-the-loop with the explicit goal of producing high-quality meshes, enabling direct supervision of surface geometry throughout training.} \\

%% file: figures/architecture.tex
\begin{figure*}[!ht]
\centering
\includegraphics[width=0.9\linewidth]{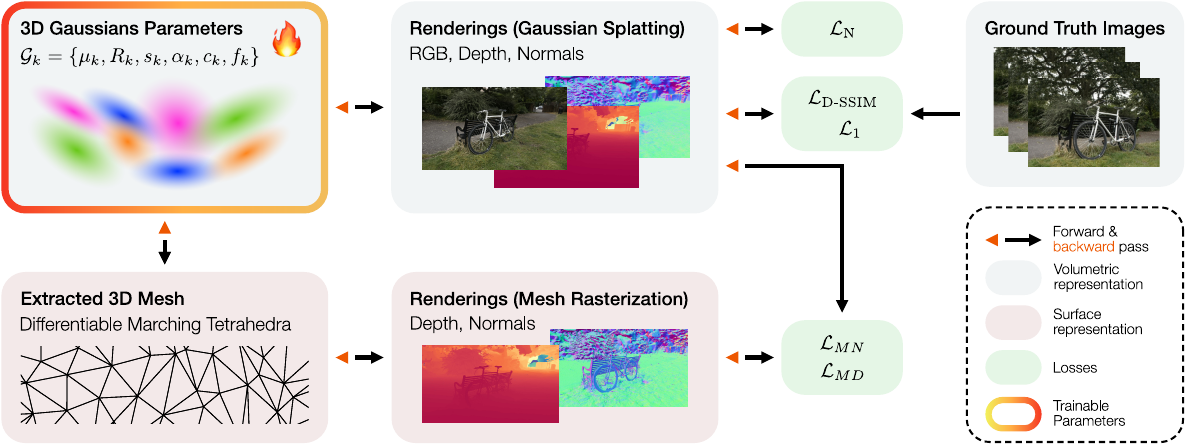}
\caption{
\textbf{MILo Pipeline.} We use Gaussians as pivots for the parametrization of a mesh we can differentiably extract at every iteration. Supervision on this mesh allows us to impose a soft prior on the Gaussians which results in better reconstructed surfaces
}
\label{fig:architecture}
\end{figure*}

%% file: sections/3_method.tex
\section{Overview}
\label{sec: overview}

Our method aims to address the fundamental challenge of extracting high-quality surface meshes from 3D Gaussian Splatting representations. Rather than treating mesh extraction as a post-processing step, we integrate it directly into the optimization process and ensure that both the volumetric and surface representations are consistent. By extracting a mesh at every training iteration and back-propagating gradients to the Gaussians, we guide them toward configurations optimized for accurate surface extraction. See Fig.~\ref{fig:architecture} for a visual overview of our method.

At each training iteration, our pipeline consists of five main steps:

\begin{enumerate}
    \item Fetch the \textit{trainable} Delaunay vertices derived from the Gaussian Pivots (Sec.~\ref{sec:delaunay_sampling}).
    \item Update the Delaunay triangulation (Sec.~\ref{sec:delaunay_sampling}).
    \item Fetch the \textit{trainable} signed distance values for each Delaunay vertex (Sec.~\ref{sec:computing_sdf}).
    \item Apply GPU-based differentiable Marching Tetrahedra to extract the mesh (Sec.~\ref{sec:diff_mt}).
    \item Back-propagate image-based and consistency losses to the Gaussian parameters, by simultaneously rendering the extracted mesh and the 3D Gaussians. (Sec.~\ref{sec:mesh_in_the_loop_optimization}).
\end{enumerate}

During optimization, we enforce geometric consistency between the mesh and the Gaussians by rendering and comparing depth and normal maps with both representations, encouraging the extracted surface to match the geometry encoded in the 3D Gaussians. This bidirectional consistency framework not only improves the quality of the extracted mesh but also encourages Gaussians to converge toward better, solid surfaces with multi-view consistent geometry.

\input{tables/resource_requirements}

\section{Differentiable Mesh Extraction}
\label{sec: differentiable_mesh_extraction}

In this section, we provide more details on the different steps of our differentiable mesh extraction, allowing gradients to flow from the vertices of the mesh back to the parameters of the Gaussians during optimization.

\subsection{Sampling Delaunay Vertices from Gaussians}\label{sec:delaunay_sampling}

The first step for extracting a surface mesh is to identify appropriate points in 3D space that will serve as the vertices of our Delaunay triangulation. While a naive approach would consist in using the \textbf{centers} of \textbf{all} Gaussians as Delaunay vertices, this simple strategy presents two significant limitations:

\begin{enumerate}
    \item The \textbf{centers} of Gaussians are typically located on or near the surface of the scene, whereas for effective application of marching tetrahedra, one needs Delaunay vertices straddling the target surface.
    \item Using \textbf{all} Gaussians as Delaunay vertices can be computationally expensive for large scenes. 
\end{enumerate}

To address the first issue, we follow the strategy proposed by Gaussian Opacity Fields~(GOF)~\cite{Yu2024GOF}: we sample 9 points per Gaussian, including its center and 8 corner points aligned with its principal axes. These are obtained by scaling and rotating the unit bounding box vertices and center $\{b_0, b_1...b_8\}$:
\begin{align}
p_{k, i} &= \mu_k + R_k \times (s_k \odot b_i) \ \ \text{for} \ \ i=0 ...8 \>,
\end{align} 
where $\odot$ is the Hadamard product.

This sampling strategy ensures that the triangulation adapts to the anisotropic nature of the primitives, creating a more accurate representation of the surface.

To address the scalability challenge, we propose to sample Delaunay vertices only from Gaussians that are most likely to be located near the surface. To do so, we leverage the \emph{importance-weighted sampling} introduced in Mini-Splatting2~\cite{fang2024mini}, which ranks Gaussians based on their contribution to rendering across all training views, as reflected by the average magnitude of their blending coefficients along camera rays. We use these importance scores as sampling probabilities, allowing us to select a subset of Gaussians that effectively preserve the geometric structure and fine details of the scene. 

While only selected Gaussians receive direct gradient updates from the mesh regularization, since we impose consistency between the 3DGS and mesh rendering (see section \ref{sec:mesh_in_the_loop_optimization}), all Gaussians are constrained in practice. The result is a computationally efficient and geometrically accurate triangulation. 

\input{tables/surface_metric_tandt}

We propose two variants of our method:
\begin{itemize}
    \item \textbf{Base model:} 
    We sample a set of Gaussians using Importance-weighted sampling~\cite{fang2024mini} and remove all other Gaussians from the scene. We extract a mesh at each iteration from the remaining Gaussians. This results in a lightweight set of Gaussians---between 0.1M and 0.5M
    depending on the complexity of the scene---and a correspondingly lightweight mesh that still captures fine details.
    
    \item \textbf{Dense model:} 
    We also sample a set of Gaussians using Importance-weighted sampling~\cite{fang2024mini}, but we do not remove the other Gaussians from the scene: We maintain the large set of Gaussians (typically between 2M and 5M) for Gaussian Splatting but use only the sampled ones as pivots for generating the Delaunay vertices. The number of Delaunay vertices is approximately the same as the base model, still resulting in a lightweight mesh; however, having more Gaussians helps in learning better SDF values at these vertices. This model results in longer optimization times, but better performance.
\end{itemize}
Once we obtain the Delaunay vertices, we compute their \textbf{Delaunay triangulation} which provides a tetrahedralization required for the following step.

\subsection{Computing Signed Distance Values}\label{sec:computing_sdf}

To compute a surface mesh from the previously obtained Delaunay triangulation, we rely on the Marching Tetrahedra algorithm~\cite{Doi1991marchingtet}. This algorithm, which we will describe in Sec.~\ref{sec:diff_mt}, requires a tetrahedral grid augmented with scalar values, typically derived from a Signed Distance Field (SDF). To apply it, we must assign a (signed) scalar value to each Delaunay vertex. Although these values do not originate from a true SDF, we refer to them as SDF values for simplicity.

\input{figures/gaussian_tetrahedralization}

We propose a simple yet effective strategy: \textbf{augmenting each Gaussian $\mathcal{G}_k$ with 9 optimizable SDF values $f_k \in \mathbb{R}^9$}---one for each Delaunay vertex. Importantly, these SDF values are decoupled from the Gaussians’ other parameters---such as opacity, scale, and rotation---allowing for localized control over the extracted isosurface level. This decoupling proves highly beneficial for accurately capturing fine surface details and ensuring strong consistency between the extracted mesh and the underlying volumetric representation. For an even faster convergence, we devise a custom initialization algorithm: please refer to the supplementary material for more details.  An illustration of our approach is provided in Fig~\ref{fig:gaussian_tetrahedralization}.

\subsection{Differentiable Marching Tetrahedra}
\label{sec:diff_mt}

Once the Delaunay vertices and the SDF values are computed, we apply Marching Tetrahedra~\cite{Doi1991marchingtet} to extract a triangle mesh. For each tetrahedron with vertices having SDF values of opposite signs (indicating that the surface intersects the tetrahedron), the algorithm computes the intersection points of the surface with the edges of the tetrahedron. These 3 (or 4) intersection points form the vertices $\{v_n\}$ of the final mesh, and they are trivially connected with 1 (or 2) triangle faces. Specifically,
given two Delaunay vertices $p_{k,i}$ and $p_{k', j}$ of a tetrahedron with SDF values $f_{k, i}$ and $f_{k', j}$ of opposite signs, the position of $v_{n}$ is given by:
\begin{equation}
    v_{n} = \frac{f_{k,i} p_{k', j} - f_{k', j} p_{k,i}}{f_{k,i} - f_{k', j}} \> ,
    \label{eq: marching_tetrahedra}
\end{equation}

Note that this mesh extraction process allows gradients to flow from the vertices of the extracted mesh back to the Gaussians through (a) the learnable SDF values and (b) the coordinates of the Delaunay vertices, computed from the means and covariances of the Gaussians.

\section{Mesh-in-the-Loop Optimization}
\label{sec:mesh_in_the_loop_optimization}

\input{figures/results}

We build on top of previous works~\cite{kerbl3Dgaussians,Huang2DGS2024,Yu2024GOF,zhang2024rade}, whose optimization is based on successive iterations of rendering the scene with 3D Gaussians and comparing the resulting image to the target image from the captured dataset. Our approach can be plugged into any Gaussian-based method able to render depth maps and normal maps through differentiable rasterization~\cite{Huang2DGS2024,Yu2024GOF,zhang2024rade}. Please note that depending on the method, the definition of the rendered depth and normal maps may differ.

A key advantage of our approach lies in the ability to extract a triangle mesh during optimization, perform any differentiable operation on the surface, and backpropagate the resulting gradients to the Gaussians. To couple the two representations, we rely on differentiable mesh rendering: we directly compare depth and normal maps rendered from both representations to enforce geometric consistency between the Gaussians and the extracted surface. Below, we describe the different loss functions involved during  optimization.

\subsection{Volumetric rendering}

To optimize the Gaussians, we rely on the same volumetric rendering loss $\calL_{\text{vol}}$ as~\cite{Yu2024GOF,Huang2DGS2024,zhang2024rade}, which consists of two photometric terms and a regularization term:
\begin{equation}
    \calL_{\text{vol}} = (1 - \lambda_{\text{RGB}}) \calL_{1} + \lambda_{\text{RGB}} \calL_{\text{D-SSIM}} + \lambda_{\text{N}} \calL_{\text{N}} \> ,
    \label{eq: volumetric_rendering_loss}
\end{equation}

where $\calL_{1}$ is the L1 loss, $\calL_{\text{D-SSIM}}$ is a D-SSIM term and $\calL_{\text{N}}$ is a normal consistency loss defined as the following sum over pixels $i$:

\begin{equation}
    \calL_{\text{N}} = \sum_{i} \left(1 - \textbf{N}(i) \cdot \tilde{\textbf{N}}(i)\right) \> ,
    \label{eq: gaussian_normal_consistency_loss}
\end{equation}

where $\textbf{N}_i$ is the expected normal at pixel $i$ computed from volumetric rendering, and $\tilde{\textbf{N}}_i$ is the normal direction at pixel $i$ obtained by applying finite difference on the rendered depth map, following~\cite{Huang2DGS2024,Yu2024GOF,zhang2024rade}. This term encourages Gaussians to align with their neighbors and drastically reduces the noise in the rendered depth and normal maps.

\subsection{Volume-to-Surface Consistency}

To enforce consistency between the geometry encoded in (a) the Gaussians as a volumetric representation and (b) the extracted surface mesh, we introduce the following loss:

\begin{equation}
    \calL_{\text{mesh}} = 
    \lambda_{\text{MD}} \calL_{\text{MD}}
    + \lambda_{\text{MN}} \calL_{\text{MN}}  \>.
    \label{eq: mesh_consistency_loss}
\end{equation}

Here $\calL_{\text{MD}}$ is a depth consistency loss defined as follows:

\begin{equation}
    \calL_{\text{MD}} = \sum_{i} 
        \log\left(1 + |D(i) - D_{\text{M}}(i)|\right)
    \> ,
    \label{eq: mesh_depth_consistency_loss}
\end{equation}

with $D$ the depth map rendered from the Gaussians and $D_{\text{M}}$ the depth map rendered from the mesh. The normal consistency loss $\calL_{\text{MN}}$ is defined as

\begin{equation}
    \calL_{\text{MN}} = \sum_{i} \left(1 - \tilde{\textbf{N}}(i) \cdot N_{\text{M}}(i)\right) \> ,
    \label{eq: mesh_normal_consistency_loss}
\end{equation}

with $N_{\text{M}}(i)$ the normal of the face of the mesh rasterized at pixel~$i$.
\subsection{Regularization}

Despite their effectiveness, the previous losses are not sufficient for reconstructing optimal meshes, and two key challenges remain.

\paragraph{Erosion.} During optimization, if all SDF values inside a tetrahedron become positive, fine details of the geometry can be eroded or even lost. Due to the sharp nature of mesh rasterization, it becomes extremely difficult for the representation to recover this geometry, even with antialiasing applied to mesh renderings. Once a region is eroded, the gradient signal becomes weak or non-existent, making it challenging for optimization to restore the missing surface.

To address this issue, we propose a regularization term $\calL_{\text{erosion}}$ that encourages geometry to be preserved:

\begin{equation}
    \calL_{\text{erosion}} = \sum_{g \in G_{\text{Del}}} \max(0, f_{\mu_g}) \> ,
    \label{eq: anti_erosion_loss}
\end{equation}

where $G_{\text{Del}}$ is the set of Gaussians sampled for the Delaunay triangulation and $f_{\mu_g}$ is the SDF value of the center of the $g$-th Gaussian. As mentioned in Section~\ref{sec: differentiable_mesh_extraction}, the centers of all Gaussians sampled for the triangulation are used as Delaunay vertices: this regularization term aims to include the centers of these Gaussians inside the surface by encouraging their SDF values to become negative, preventing erosion in all areas where at least one Gaussian has been sampled. \diego{$\calL_{\text{erosion}}$ is applied only to the centers of selected Gaussian pivots, not to all tetrahedral vertices, preventing collapse. This ensures regularization is effective without harming mesh integrity.}

\paragraph{Interior artifacts.} The previous losses do not provide adequate constraints on occluded parts of the scene. As a consequence, the interior of the surface mesh, which does not receive proper supervision through depth and normal renderings, tends to produce chaotic structures and internal cavities, rather than being empty as it should be. These interior artifacts can impact downstream applications such as mesh editing or physics simulations.

To prevent such artifacts inside the mesh, we introduce a novel loss based on a feedback loop between the SDF values and the extracted mesh. This loss aims to enforce occluded points located inside the mesh to have negative SDF values. Specifically, after extracting the surface using the Gaussians and their SDF values as described in section~\ref{sec:diff_mt}, 
we use the mesh to build an occupancy label~$o_p \in \{0, 1\}$ for each Delaunay site $p$, indicating whether these points are inside or outside the visible portion of the surface mesh as described below. We finally enforce the Delaunay vertices labeled as inside to have negative SDF values, without enforcing any additional constraint for Delaunay vertices labeled as outside---as these points already receive supervision from the renderings.

This regularization term is computed as follows:

\begin{equation}
    \calL_{\text{interior}} = \sum_p H(\sigma(-f_p), o_p) \cdot o_p \> ,
    \label{eq: interior_loss}
\end{equation}

where $H$ is the cross-entropy loss, $\sigma$ is the sigmoid function, $f_p$ is the SDF value of the Delaunay site $p$, and $o_p$ is the occupancy label of $p$.

To compute the occupancy labels $o_p$, we first render depth maps of the mesh from all training viewpoints. Then, for any Delaunay site $p$, we classify it as inside the mesh if it located ``behind'' all depth maps containing $p$ in their field of view. Updating the occupancy labels requires a few seconds and can be performed every 200 iterations, resulting in a computationally efficient and effective regularization term for producing meshes with clean interiors.

Our full optimization loss is finally defined as:

\begin{equation}
    \calL = \calL_{\text{vol}} + \calL_{\text{mesh}} + \calL_{\text{reg}} \> ,
    \label{eq: final_loss}
\end{equation}

with

\begin{equation}
    \calL_{\text{reg}} = \lambda_{\text{erosion}} \calL_{\text{erosion}} + \lambda_{\text{interior}} \calL_{\text{interior}} \> ,
    \label{eq: regularization_loss}
\end{equation}

where $\lambda_{\text{erosion}}$ and $\lambda_{\text{interior}}$ are hyperparameters controlling the strength of the anti-erosion and interior regularization, respectively.

%% file: tables/resource_requirements.tex
\begin{table}[t]
    \centering
    \caption{\textbf{Resource requirements for different methods}. We compare the computational resources required for training on high-resolution images, as well as the resulting model sizes across different approaches. All measurements are averaged across scenes from the Tanks \& Temples dataset.
    Our method achieves high-quality reconstruction with significantly lower resource requirements, particularly regarding mesh complexity. The base version of our model uses up to 10x fewer Gaussians than competing methods depending on the scene, producing meshes with fewer vertices and triangles while preserving details and achieving better accuracy, making it more suitable for downstream applications.}
    \vspace{-0.2cm}
    \resizebox{0.98\linewidth}{!}{
    \begin{tabular}{@{}l|ccc|ccc}
     \multirow{2}{*}{} & \multicolumn{3}{c|}{Training Resources} & \multicolumn{3}{c}{Output Mesh} \\
     & \#Gaussians (M) & GPU Mem (GB) & Time & \#Vertices & \#Triangles & Size (MB) \\
     \hline
     2DGS & 0.98 & 4.7 GiB & 29m & 16.39 M & 21.68 M & 557.1 \\
     GOF & 1.55 & 10.6 GiB & 93m & 16.49 M & 33.17 M & 600.0 \\
     RaDe-GS & 1.56 & 12.4 GiB & 42m & 14.75 M & 29.59 M & 592.0 \\
     \hline
     Ours (base) &  0.28 &  10.0 GiB &  50m &  4.36 M &  8.97 M &  179.6 \\
     Ours (dense) & 2.11 &  16.5 GiB &  110m &  6.89 M &  13.79 M &  276.1 \\
    \end{tabular}
    }
    \label{tab:resource_requirements}
    \vspace{-0.2cm}
\end{table} 

%% file: tables/surface_metric_tandt.tex
\begin{table*}[!ht]
\centering
\caption{\textbf{Quantitative comparison on the Tanks \& Temples Dataset~\cite{Knapitsch2017}}. We report the F1-score and average optimization time. All results are evaluated with the official evaluation scripts. $\text{Ours}_{\text{X}}$ uses the differentiable rasterizer from X. 
VCR-GauS~\cite{chen2024vcr} relies on a pre-trained normal estimation model.
Best results for implicit methods are highlighted in blue; best results for explicit methods are highlighted in red. \diego{When a method fails (OOM), we report the mean of the successful scenes, denoted by a superscript asterisk ($x^*$).}
Our method achieves the best F1 score among explicit representations, even though it outputs lighter meshes than other approaches.
}
\vspace{-0.2cm}
\resizebox{0.98\linewidth}{!}{
\begin{tabular}{@{}l|cccccc|cccccc|c|ccc}
 & \multicolumn{6}{c@{}|}{Implicit} & \multicolumn{9}{c@{}}{Explicit} \\ 
 & \diego{Gaussian UDF} & \diego{GS-Pull} & \diego{GSDF} & NeuS & Geo-Neus & Neuralangelo & SuGaR & 3DGS & 2DGS & GOF & RaDe-GS & QGS & VCR-GauS & \textbf{$\text{Ours}_{\text{RaDe-GS}}$ (base)} & \textbf{$\text{Ours}_{\text{GOF}}$ (base)} & \textbf{$\text{Ours}_{\text{RaDe-GS}}$ (dense)}\\ 
 \hline
Barn & 0.27 & \sbestbis  0.60 & 0.16 &  0.29 &  \tbestbis 0.33 &  \bestbis 0.70  & 0.14 & 0.13 & 0.36 & 0.51 & 0.43 & 0.43 & \sbest 0.62 & 0.55 & \tbest \diego{0.59} & \best \diego{0.64} \\
Caterpillar & 0.17 & \bestbis 0.37 & 0.11 & \tbestbis 0.29 & 0.26 & \sbestbis 0.36 & 0.16 & 0.08 & 0.23 & 0.41 & \tbest 0.32 & 0.31 & 0.26 & \sbest 0.38 & \best \diego{0.39} & \sbest \diego{0.38} \\
Courthouse & 0.03 & \tbestbis 0.16 & OOM & \sbestbis 0.17 & 0.12 &  \bestbis 0.28 & 0.08 & 0.09 & 0.13 & \tbest 0.28 & 0.21 & 0.26 & 0.19 & \tbest 0.28 & \sbest \diego{0.29} &\best \diego{0.31}\\
Ignatius & 0.38 & 0.71 & 0.34 & \sbestbis 0.83 & \tbestbis 0.72 &  \bestbis 0.89 & 0.33 & 0.04 & 0.44 & 0.68 & 0.69 & \best 0.79 & 0.61 & 0.73 & \sbest \diego{0.78} & \tbest \diego{0.76} \\
Meetingroom & 0.17 & \tbestbis 0.22 & 0.01 & \sbestbis 0.24 &  0.20 &  \bestbis 0.32 &  0.15 & 0.01 & 0.16 &\best 0.28 & \tbest 0.25 & \tbest 0.25 & 0.19 & \tbest 0.25 & \sbest \diego{0.26} & \best \diego{0.28} \\
Truck & 0.30 & \bestbis 0.52 & 0.25 & \tbestbis 0.45 & \tbestbis 0.45 & \sbestbis 0.48 &  0.26 & 0.19 & 0.26 & \tbest 0.59 & 0.51 & \sbest 0.60 & 0.52 & \sbest 0.60 & \best \diego{0.62} & \tbest \diego{0.59}\\ 
\hline
Mean & 0.22 & \sbestbis 0.43 & $0.18^*$ & \tbestbis 0.38 & 0.35 & \bestbis 0.50 & 0.19 & 0.09 & 0.30 & \tbest 0.46 & 0.40 & 0.44 & 0.40 & \sbest 0.47 & \best \diego{0.49} & \best \diego{0.49} \\
Time & \tbestbis 90m & \bestbis 37.6m & \sbestbis 70m & >24h & >24h & >24h & 73~m & \best 7.9~m & \tbest 12.3 ~m & 69m & \sbest 11.5~m & 120 m & 53 m & 50 m & \diego{150 m} &110 m\\ 
\end{tabular}
}
\label{tab:surface_metrics_tandt}
\vspace{-0.2cm}
\end{table*}

%% file: figures/gaussian_tetrahedralization.tex
\begin{figure}
    \centering
    \includegraphics[width=.7\linewidth]{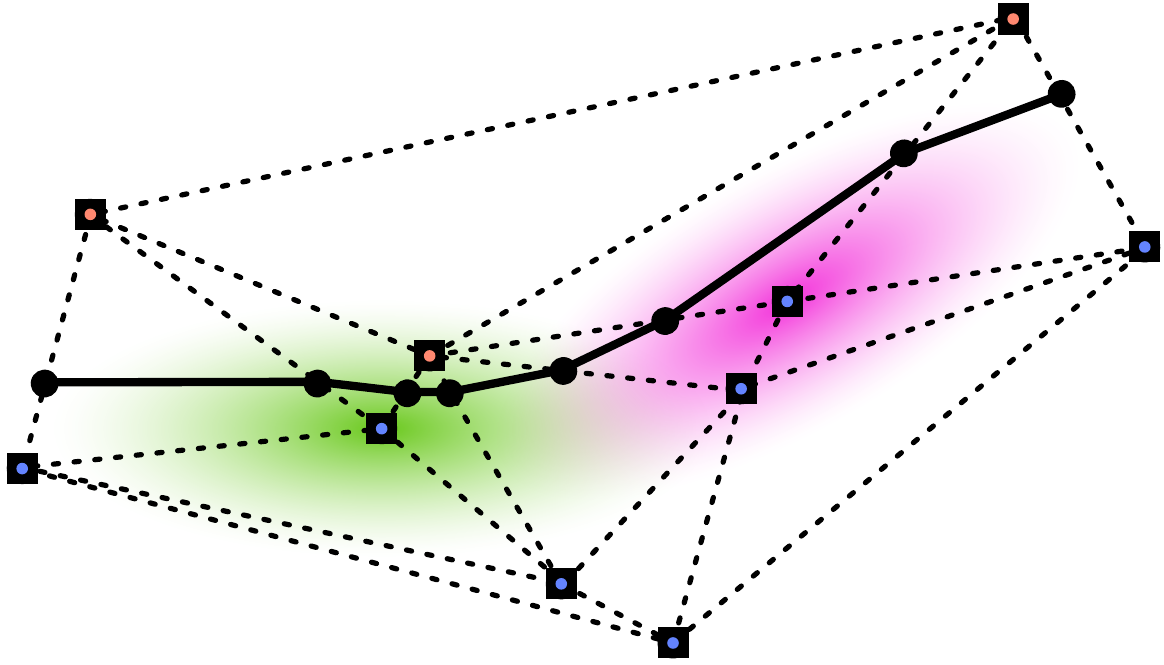}
    \caption{\textbf{Overview of Gaussian Pivots in 2D.} Two Gaussians each generate Delaunay vertices (black squares) with continuous SDF values (red and blue dots). Their Delaunay triangulation (dashed lines) is iso-surfaced with Marching Tetrahedra, which places final mesh vertices (black dots) on edges where sign changes occur, resulting in the extracted mesh (bold lines).}
    \label{fig:gaussian_tetrahedralization}
    \vspace{-0.2cm}
\end{figure}

%% file: figures/results.tex
\begin{figure*}[t]
    \centering
    \resizebox{\linewidth}{!}{
        \includegraphics[width=\linewidth, height=0.5625\linewidth]{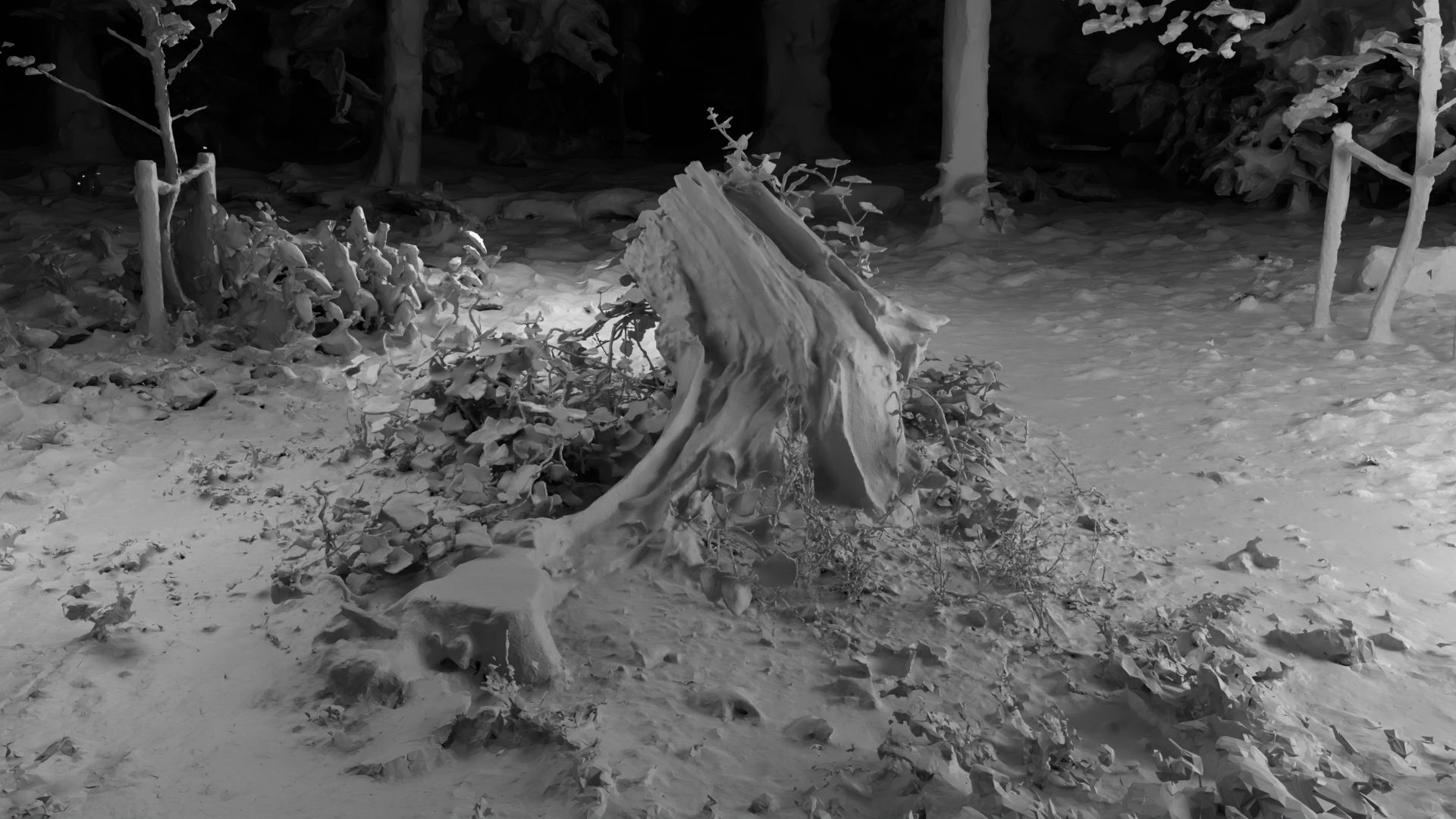}
         \includegraphics[width=\linewidth, height=0.5625\linewidth]{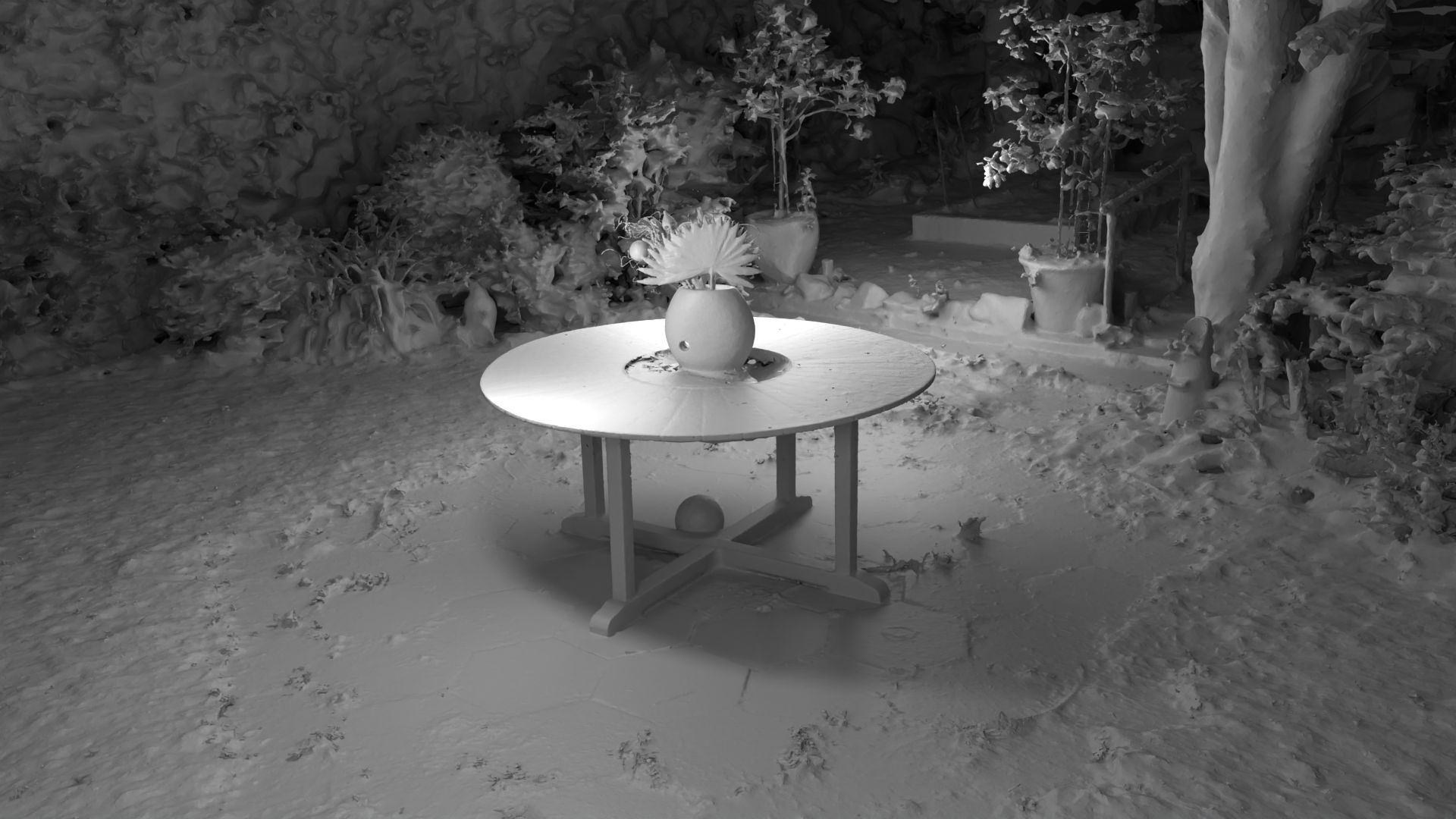}
         \includegraphics[width=\linewidth, height=0.5625\linewidth]{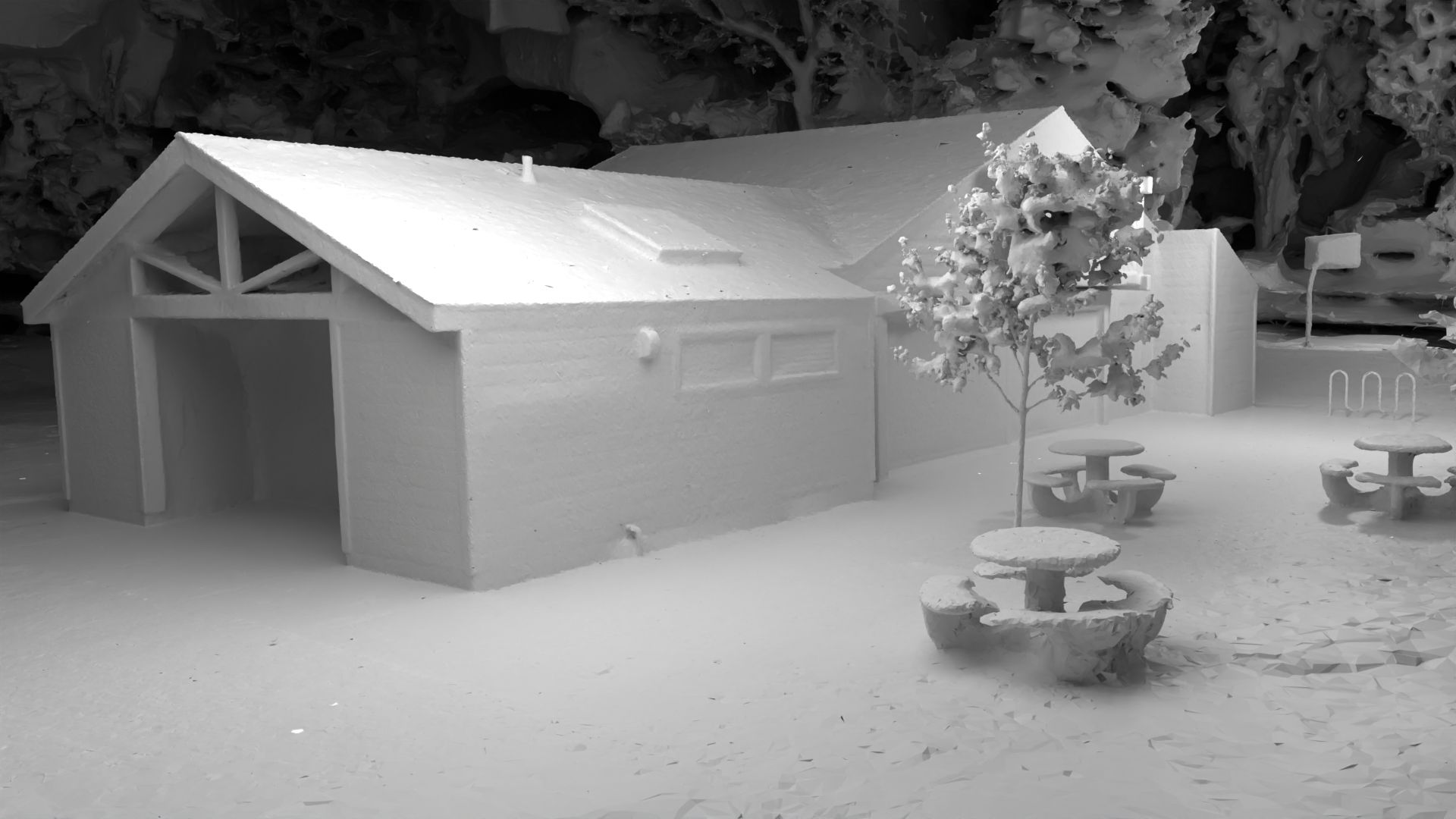}
    }
    \vspace*{-4mm}
    \caption{\textbf{Examples of surface reconstruction results on the Tanks and Temples and MipNeRF~360 datasets}. Our method produces meshes with clean surfaces and very fine details, while being lighter than previous methods  (Stump: \textit{320MB}, Garden: \textit{301MB}, Barn: \textit{313MB})}
    \label{fig:results}
    \vspace*{-3mm}
    \end{figure*} 

%% file: sections/4_experiments.tex
\input{figures/qualitative_comparison}
\input{tables/surface_metric_dtu}

\section{Experiments}
\label{sec:experiments}

In this section, we evaluate our method on various datasets and compare it with state-of-the-art approaches for surface reconstruction from multi-view images. We first describe our experimental setup, then present quantitative and qualitative results for surface reconstruction quality, followed by ablation studies and applications in mesh editing and animation. We show that the obtained meshes are clean, accurate and light as showcased in Fig.~\ref{fig:results}.

\paragraph{Implementation Details.} Our method is implemented in PyTorch, building upon the RaDe-GS codebase for Gaussian Splatting with depth and normal maps rendering.
We train the Gaussian representation using standard 3D Gaussian Splatting techniques with aggressive densification and importance-weighted sampling inspired by Mini-Splatting2~\cite{fang2024mini}. We provide more details on the optimization procedure, SDF normalization, mesh rendering and Delaunay triangulation in the supplementary material.

\def \backgroundRGBmetric {Mesh-Based Novel View Synthesis}

\paragraph{Datasets.} We evaluate our method on several standard datasets: (1) the Tanks and Temples (T\&T) dataset~\cite{Knapitsch2017}, which contains complex real-world scenes with varying scales and challenging geometry; (2) the DTU dataset~\cite{jensen2014large}, which provides scans of real objects with ground truth geometry; (3) the Mip-NeRF 360 dataset~\cite{barron22mipnerf360}, which contains unbounded and complex scenes; (4) and DeepBlending~\cite{hedman18deepblending} which contains real-world scenes with challenging geometry. For T\&T, we follow previous works~\cite{Huang2DGS2024,Yu2024GOF,zhang2024rade} and use the scenes Barn, Caterpillar, Ignatius, Courthouse, Meetingroom, and Truck. For DTU, we use the same 15 scenes as in previous works~\cite{Huang2DGS2024,Yu2024GOF,zhang2024rade}.

\paragraph{Metrics.} For quantitative evaluation of surface reconstruction quality, we follow standard practice~\cite{Yu2024GOF,zhang2024rade} and report F1-score for T\&T and Chamfer Distance for DTU. For novel view synthesis on Mip-NeRF 360 and DeepBlending, we report PSNR, SSIM, and LPIPS metrics. 

No dataset provides ground truth geometry for background objects. To mitigate this issue we use \backgroundRGBmetric\; (see Section. \ref{sec:mesh_background_evaluation}) to measure the visual consistency between the extracted mesh and reference views, which contain background information. This evaluation measures the ability of reconstructed meshes to accurately represent full scenes, including background objects, while requiring only a set of ground truth images.

\subsection{Resources Requirements.} 

\diego{The addition of mesh-based constraints inevitably leads to increased computational load. However, we observed that only a subset of Gaussians is critical for producing an effective tetrahedral structure. 

Building on this, inspired by the sampling strategy introduced in Mini-Splatting2 \cite{fang2024mini}, we employ it for a distinct purpose: selecting the Gaussians which will spawn the Gaussian pivots that serve as Delaunay vertices. Unlike Mini-Splatting2, our dense model preserves all Gaussians throughout optimization, and the sampling criterion is used exclusively to identify those most suitable for mesh construction rather than to prune the representation. 

This targeted selection yields lighter and more accurate meshes, significantly reducing the computational cost of triangulation, Marching Tetrahedra, and mesh rasterization. Table~\ref{tab:resource_requirements} provides a comprehensive comparison of the computational resources required by concurrent methods.}

\paragraph{Memory Requirements.} All experiments were conducted on a single NVIDIA RTX 4090 GPU with 24GB of VRAM. The base model contains between 0.1 and 0.5 million Gaussians, requiring 10GB of VRAM during training. The dense model generally contains between 2 and 4 million Gaussians requiring up to 17GB.

\paragraph{Time Requirements.} A complete training run for the base model on a single GPU takes 25 minutes for bounded scenes from DTU and between 40 and 50 minutes for unbounded scenes from Tanks and Temples and Mip-NeRF 360, while the dense model requires up to 2 hours for unbounded scenes.

\paragraph{Storage Requirements.} Our base model uses significantly fewer Gaussians (0.1-0.5M) and produces meshes with fewer vertices compared to previous approaches. Despite using fewer resources, our method achieves superior performance in terms of surface reconstruction quality. The dense variant of our model maintains a larger set of Gaussians but still produces meshes with fewer vertices and triangles than competing methods, making it more suitable for downstream applications that require efficient mesh representations.

\subsection{Surface Reconstruction}

Table~\ref{tab:surface_metrics_tandt} presents the quantitative results on the Tanks and Temples dataset. Our method consistently outperforms previous approaches in terms of F-score, demonstrating the effectiveness of our mesh-in-the-loop optimization strategy. The dense variant of our method achieves the best results among all Gaussian-based approaches.

We provide results on the DTU dataset, consisting of small, object-centric scenes. Table~\ref{tab:surface_metrics_dtu} shows the results on this dataset. \diego{Note that, DTU consists of isolated objects in highly controlled scenes, where standard post-hoc mesh extraction is already effective. MILo is particularly effective for complex full-scene reconstruction. Still, our mesh-in-the-loop optimization maintains competitive performance in terms of Chamfer Distance (CD) on DTU.} 
\diego{In line with established practice, we evaluate performance using F-Score on the Tanks and Temples dataset and Chamfer Distance (CD) on DTU.}

\input{tables/rendering_metrics_mesh}
\input{figures/tnt_cumulative}

Figure~\ref{fig:qualitative_comparison} shows qualitative comparisons of our method with previous approaches on selected scenes from the Tanks and Temples and Mip-NeRF 360 datasets. Our method produces meshes with significantly cleaner surfaces and better preservation of fine details. In particular, our approach effectively addresses the erosion problem present in previous methods, resulting in more complete reconstructions of thin structures and complex geometry: see Fig.~\ref{fig:tnt_cumulative} for a detailed analysis.

\subsection{Mesh-Based Novel View Synthesis}
\label{sec:mesh_background_evaluation}

Despite extensive research on mesh reconstruction from images, current evaluation benchmarks remain limited. Common datasets—DTU, MipNeRF 360, and Tanks \& Temples—all have notable drawbacks: MipNeRF 360 lacks ground-truth geometry; DTU features overly controlled, object-centric scenes; and Tanks \& Temples provides sparse ground-truth limited to foreground regions. Moreover, evaluation protocols to measure the alignment of extracted surfaces with ground truth images remain uncommon in surface reconstruction research.

We highlight this as an important issue in surface reconstruction. Inspired by previous works which evaluate surface-based view synthesis methods \cite{Reiser2024SIGGRAPH}, we use \backgroundRGBmetric\;as an evaluation method for meshes that relies solely on ground-truth RGB images.
The core intuition is that better geometry should yield better mesh-based renderings: for each test view, we render the scene using the mesh and compare the resulting image to the corresponding ground-truth image.

By measuring the visual consistency between mesh renderings and reference views, this metric allows us to evaluate:
\begin{itemize}
    \item Geometric artifacts and misalignment between the reconstructed surface and the ground truth images (\textit{e.g.}, surface erosion or inflation);
    \item Mesh completeness, since missing geometry usually results in degraded rendering performance;
    \item Background reconstruction, even when 3D ground truth is unavailable.
\end{itemize}

To render the mesh from a given viewpoint, we first rasterize the triangles using Nvdiffrast~\cite{Laine2020nvdiffrast}. 
Then, we associate a color with each pixel based on the rasterized triangle. 

A naive approach would rely on vertex colors, assigning RGB values to each mesh vertex and interpolating those values to determine a pixel's color. However, this method biases the metric towards dense meshes, as image quality would suffer for sparse, yet accurate, meshes due to limited color resolution.

To overcome this, we decouple color from mesh resolution by employing a neural color field $F_\text{color} : \IR^3 \rightarrow [0,1]^3$ for texturing the mesh. Specifically, to obtain the rendered color for any pixel $p$, we first calculate the 3D location $P\in \IR^3$ of the triangle point rasterized onto the pixel $p$. This is achieved by backprojecting the depth value using camera parameters. We then query the neural color field at this backprojected surface point to retrieve its RGB value $F_\text{color}(P)$. Unlike vertex colors, this mesh texturing process stores color values in a neural field rather than per-vertex, ensuring color assignment is independent of mesh resolution.

In practice, for each mesh under evaluation, we first train the neural field using only the training views. We adopt TensoRF\cite{chen2022tensorf} as the backbone representation and optimize it for 5k iterations. We then render test views for evaluation.
This metric effectively quantifies the alignment between mesh geometry and the image data. \diego{Note that, this evaluation protocol assumes dense testing views, as achieving good mesh-based rendering performance on a sparse test set does not necessarily translate into high quality of the underlying geometry.}

Table~\ref{tab:mesh_rendering_metrics} shows how \backgroundRGBmetric\;allows us to compare against other state-of-the-art methods. We use the differentiable rasterizer from RaDe-GS for training Ours (base). Our approach outperforms 2DGS \cite{huang20242d} and RaDe-GS \cite{zhang2024rade} across all metrics. GOF \cite{Yu2024GOF} produces highly detailed meshes, yielding strong PSNR results, but our method achieves comparable PSNR while outperforming in SSIM and LPIPS. This suggests that our meshes are not only accurate but also exhibit fewer visual artifacts and less noise compared to those from GOF. We include a qualitative comparison in the supplementary material.

\diego{Moreover, the performance on Tanks \& Temples under the \backgroundRGBmetric \; metric (Table~\ref{tab:mesh_rendering_metrics}) shows strong correlation with the F1 score computed against ground-truth geometry (Table~\ref{tab:surface_metrics_tandt}). For example, among the common baselines, 2DGS and RaDe-GS rank fourth and third, respectively, under both metrics. Ours (base) and GOF also perform consistently, with the Ours (base) achieving a slightly higher F1 score and GOF performing marginally better on \backgroundRGBmetric. These results indicate that our proposed metric is well aligned with ground-truth geometric accuracy.}

\subsection{Novel View Synthesis}

While our primary focus is on surface reconstruction, we also provide in supplementary material an evaluation of the novel view synthesis quality of our optimized Gaussians. 
Our method maintains competitive rendering quality compared to previous approaches, demonstrating that our mesh-in-the-loop optimization does not compromise the visual fidelity of the Gaussian representation.

\input{figures/ablation_depth_only_vs_mesh_only}

\subsection{Ablation Study}

\input{tables/ablation_components}

To evaluate the contribution of each component, we conduct ablation studies on the Tanks and Temples dataset (see Tab.~\ref{tab:ablation_f1_only}).
Adding depth supervision via the mesh loss ($+\mathcal{L}_{\text{MD}}$) substantially boosts $F_1$-score, while further including the normal loss ($+\mathcal{L}_{\text{mesh}}$) slightly reduces it. Still, Fig.~\ref{fig:depth_only_vs_mesh_only} shows that combining both is critical for removing noise from meshes.

Introducing the anti-erosion loss ($+\mathcal{L}_{\text{mesh}} + \mathcal{L}_{\text{erosion}}$) improves reconstruction by preserving thin structures and fine details—especially in challenging regions like fences or vegetation—by penalizing surface erosion. Our loss also acts as a mechanism to recover geometry lost due to SDF sign flips yielding a modest $F_1$-score gain (Fig.~\ref{fig:qualitative_comparison}).

Fig.~\ref{fig:interior_comparison} highlights the benefit of our interior regularization, which eliminates internal cavities and artifacts, producing watertight surfaces suitable for downstream tasks like physics simulation.

Combining all components leads to the highest $F_1$-score, validating the effectiveness of our full method. We also demonstrate its plug-and-play nature by integrating it into GOF’s pipeline.

\input{figures/interior_comparison}

\vspace{20pt}
\diego{
\subsection{Analysis of Mesh Extraction}

As discussed in Section~\ref{sec:related work}, many existing approaches to mesh extraction from images rely on the TSDF algorithm. These methods typically follow a simple yet seemingly effective pipeline: they first estimate accurate depth maps using a variety of techniques, and then fuse these maps into a mesh via TSDF. However, we argue that TSDF fusion does not scale well to large-scale reconstructions. This limitation arises from its reliance on a fixed 3D grid, which drives up memory requirements as resolution increases, and from its tendency to oversmooth geometric detail.

In Figure~\ref{fig:mesh_extraction_comparison}, we present a quantitative comparison between our extraction method and TSDF fusion. For TSDF fusion, we optimized Gaussians with the same densification strategy as MILo, but removed the mesh-in-the-loop regularization term and employed the same loss function as in \cite{Huang2DGS2024,zhang2024rade,Yu2024GOF} to avoid bias. TSDF fusion requires substantial memory and is prone to out-of-memory failures; consequently, we exclude the Courthouse scene from this benchmark, as TSDF fusion could not be run across all mesh resolutions due to memory constraints.
As shown in Figure~\ref{fig:mesh_extraction_comparison}, our method consistently achieves higher reconstruction quality, demonstrating the robustness of our learnable SDF strategy.
We highlight that, TSDF fusion requires \textbf{iterating over all training viewpoints for each extraction}, leading to a long and computationally expensive process that \textbf{prevents integration into the training loop of Gaussian Splatting representations.}

By contrast, MILo leverages learnable SDF values and a scalable pivot set to enable lightweight and efficient mesh extraction, making it practical to incorporate mesh extraction directly into the training loop.
}

\input{figures/mesh_extraction_comparison}

%% file: figures/qualitative_comparison.tex
\begin{figure*}[!ht]

\vspace*{7mm}

\centering

\resizebox{\linewidth}{!}{
    \includegraphics[width=0.20\linewidth]{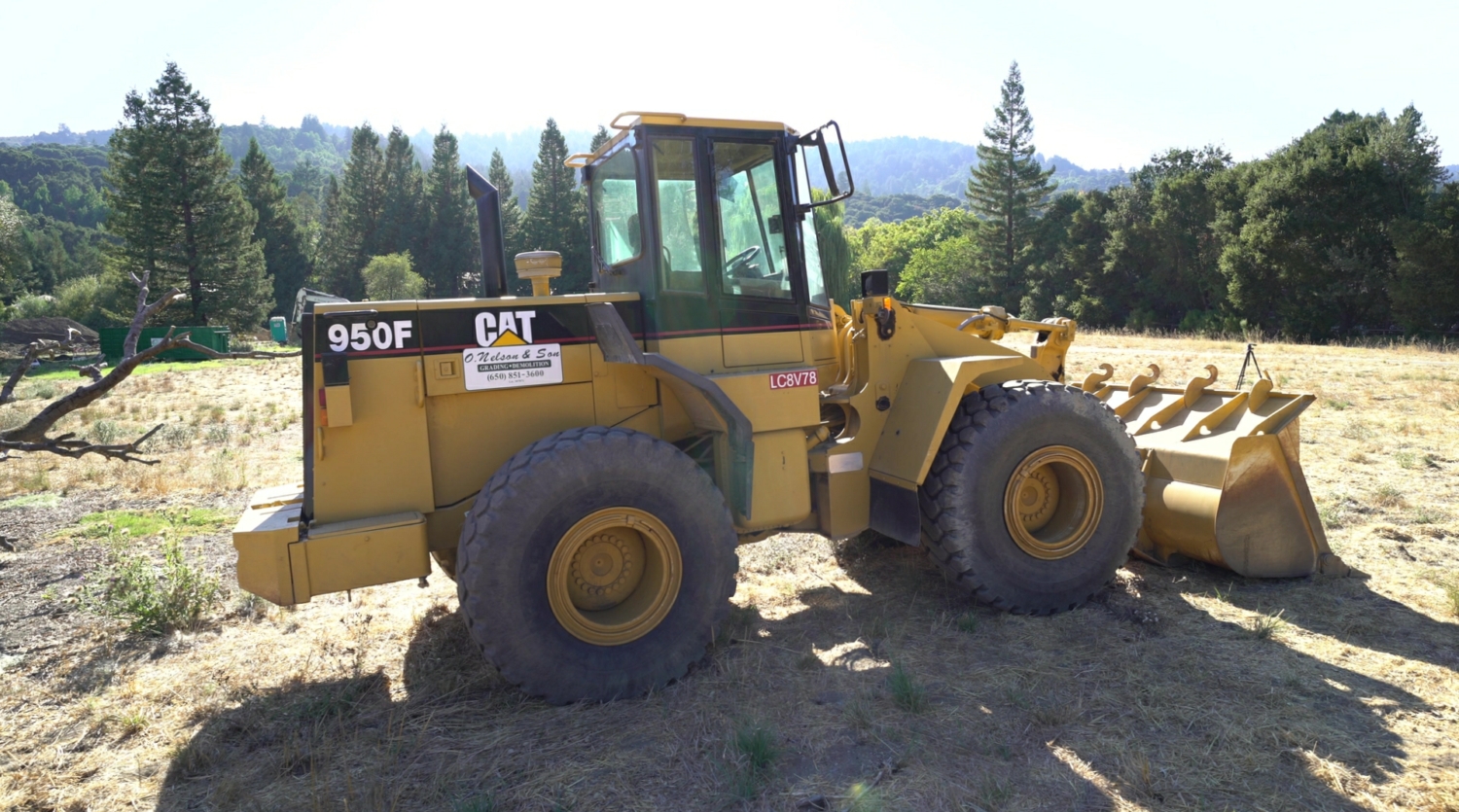}
    \includegraphics[width=0.20\linewidth]{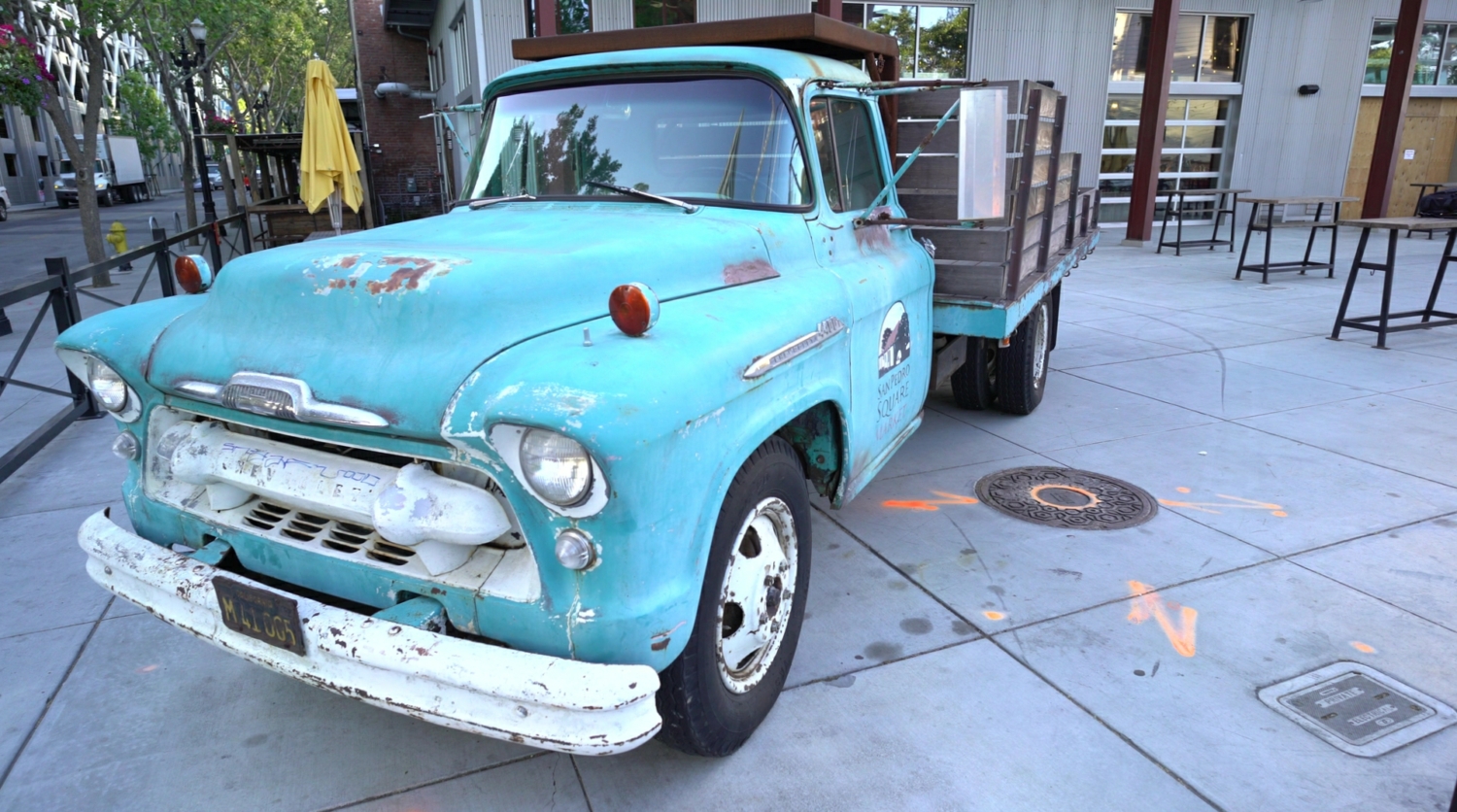}
    \includegraphics[width=0.20\linewidth]{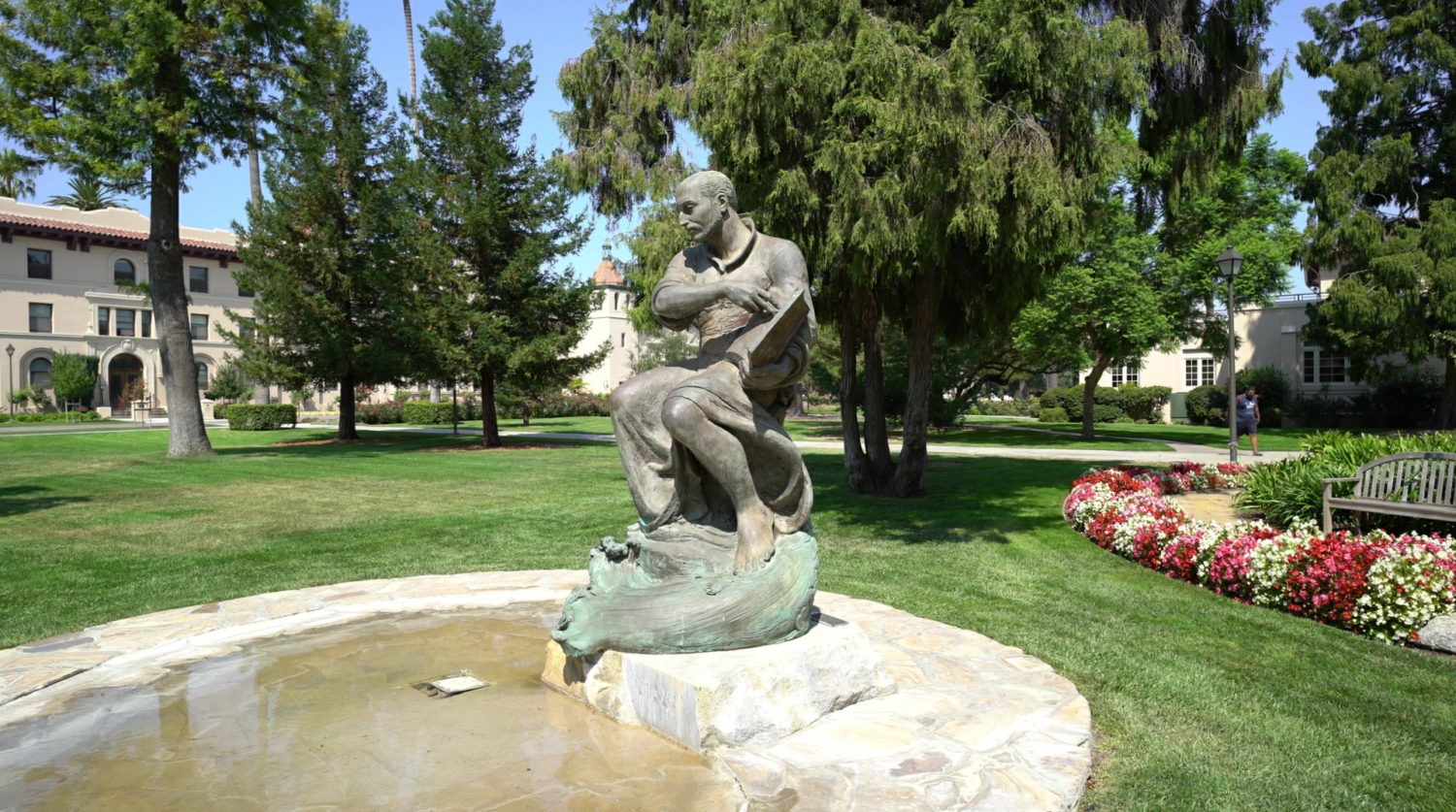}
    \includegraphics[width=0.20\linewidth]{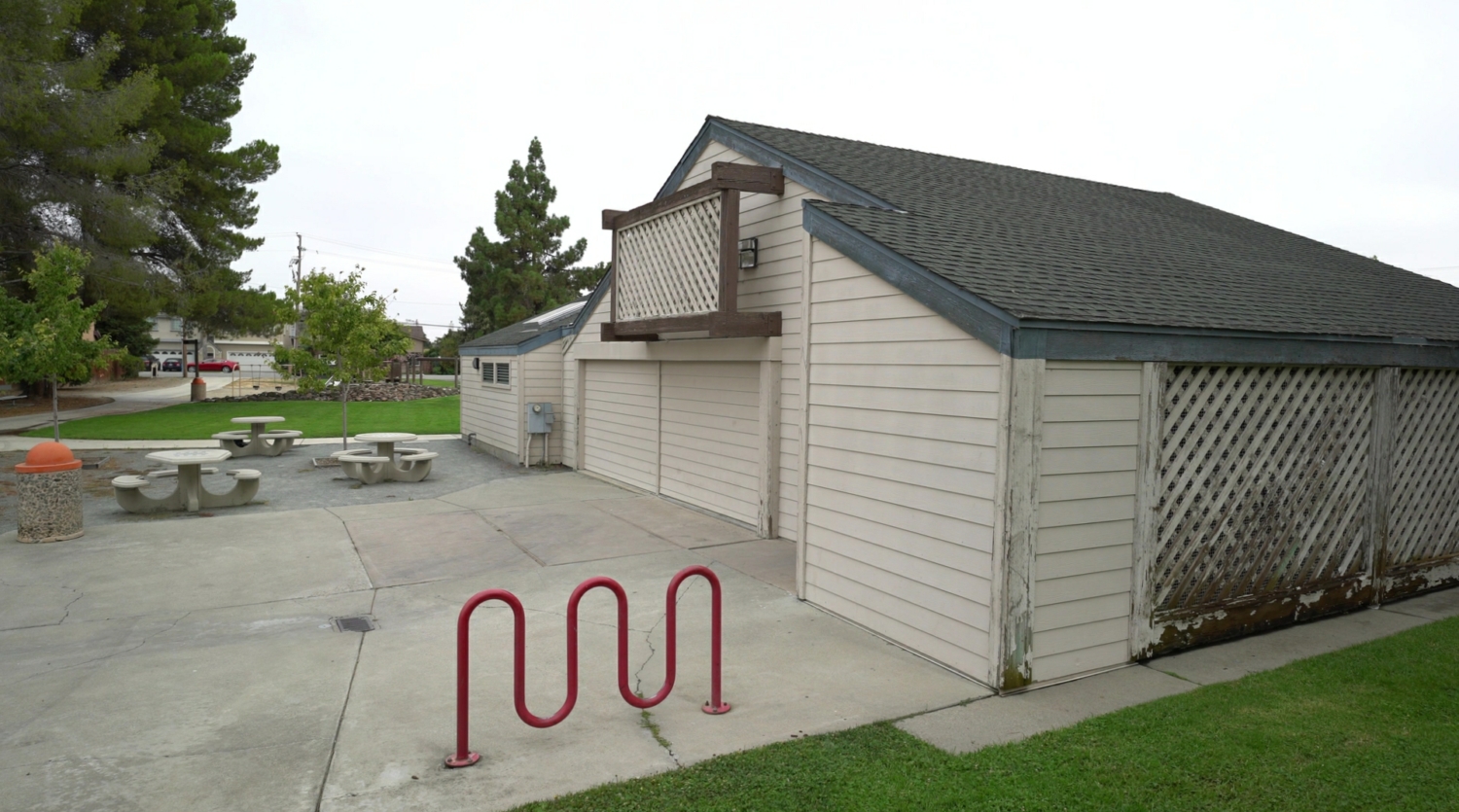}
    \includegraphics[width=0.20\linewidth]{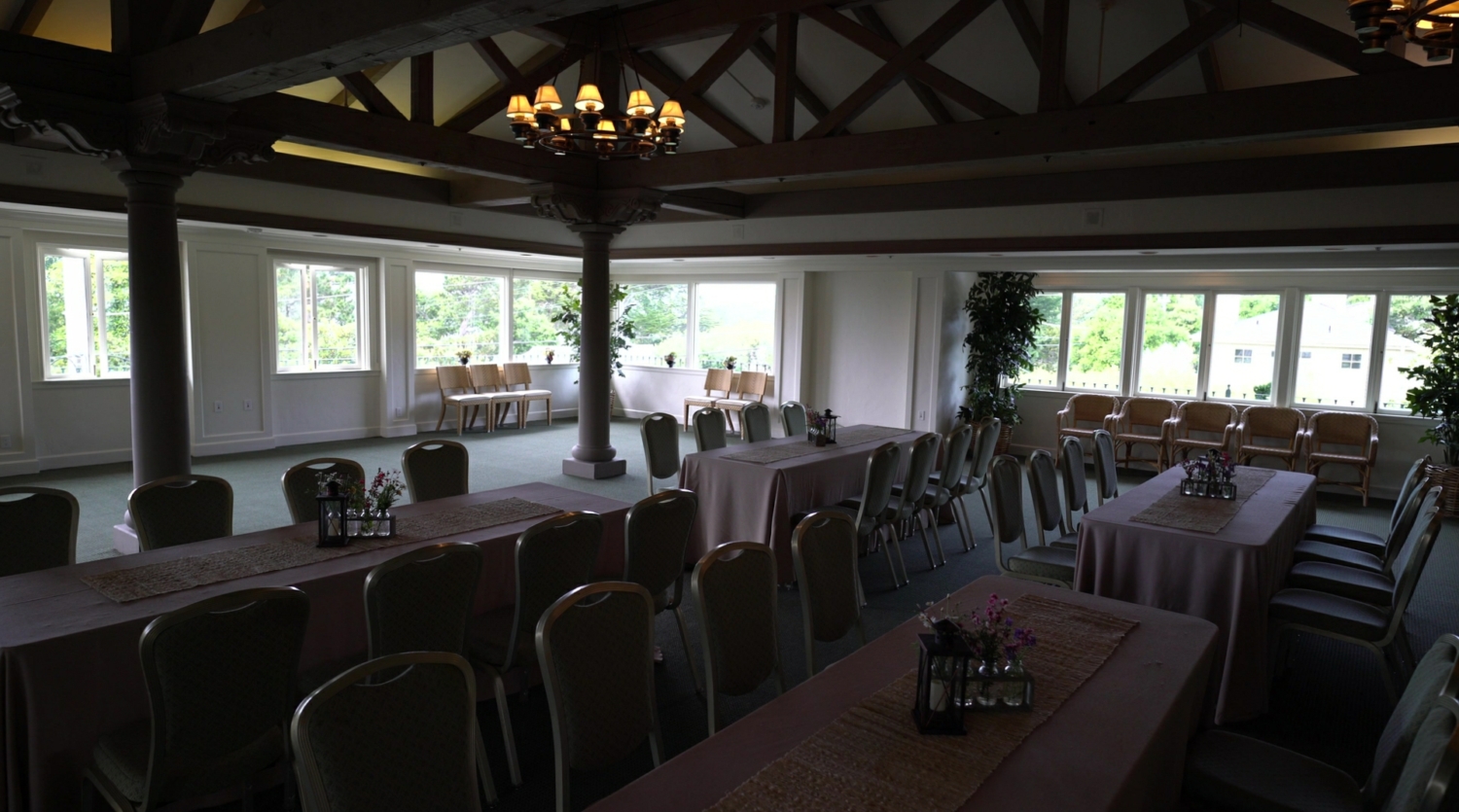}
}
\resizebox{\linewidth}{!}{
    \includegraphics[width=0.20\linewidth]{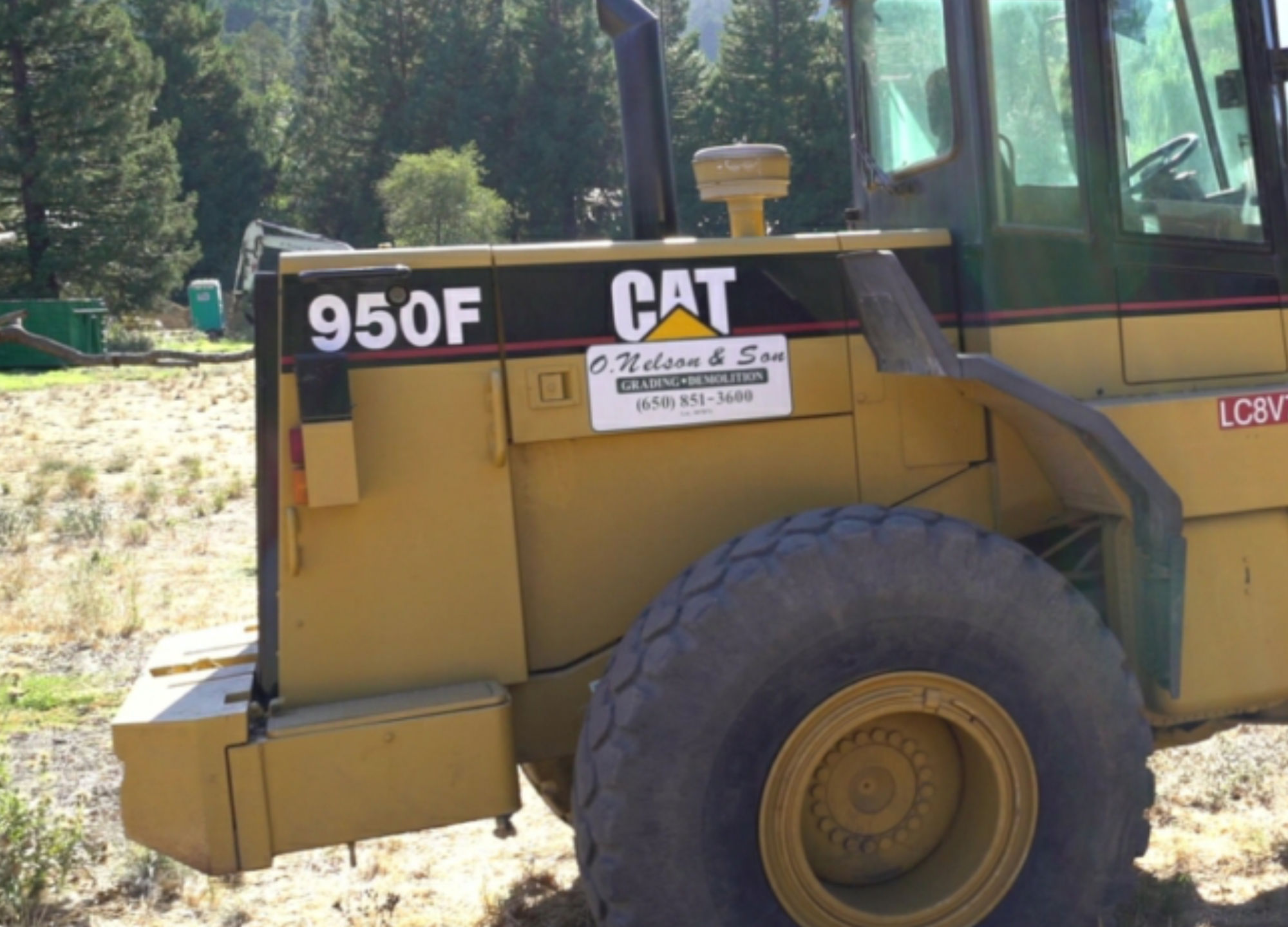}
    \includegraphics[width=0.20\linewidth]{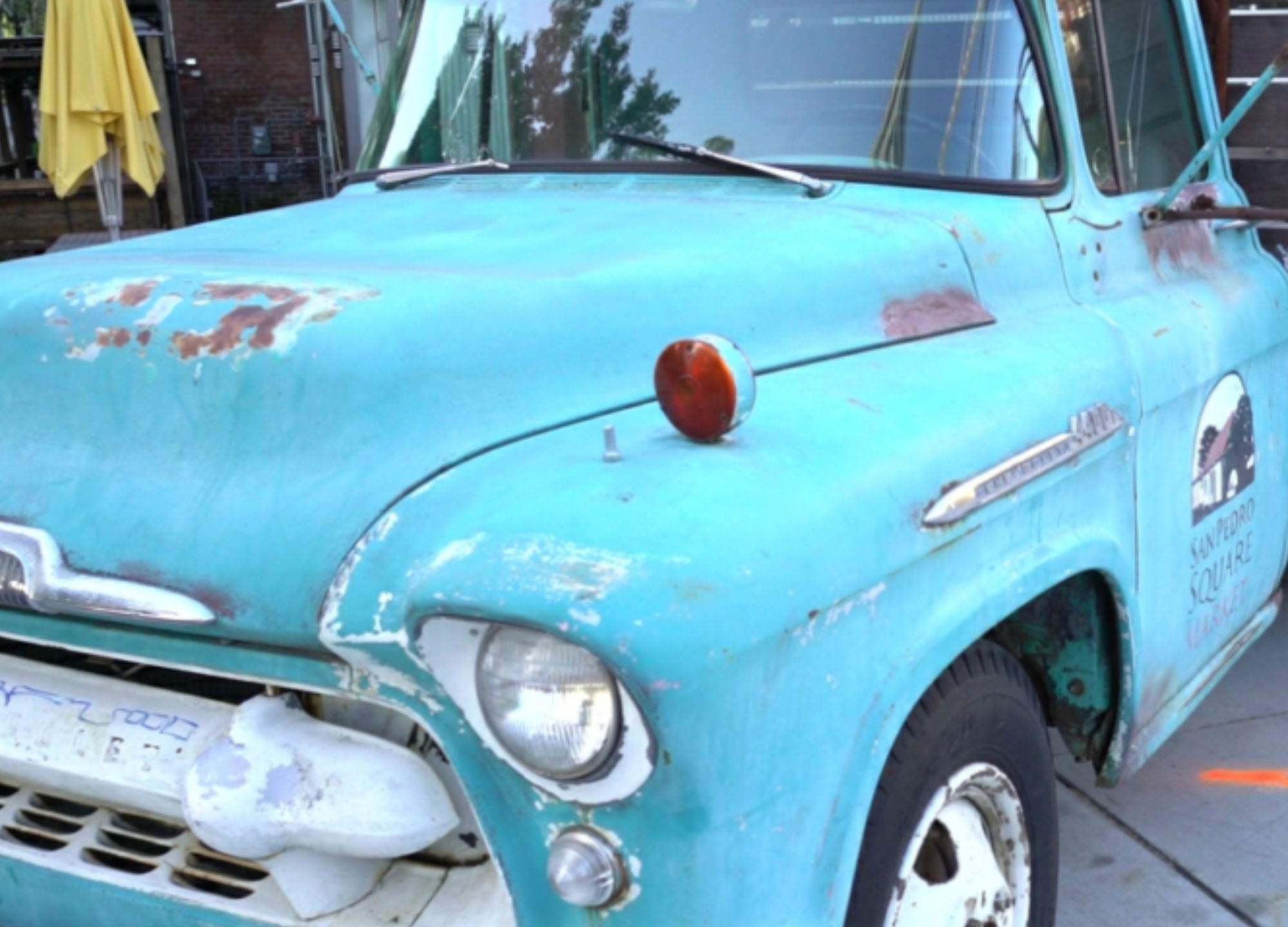}
    \includegraphics[width=0.20\linewidth]{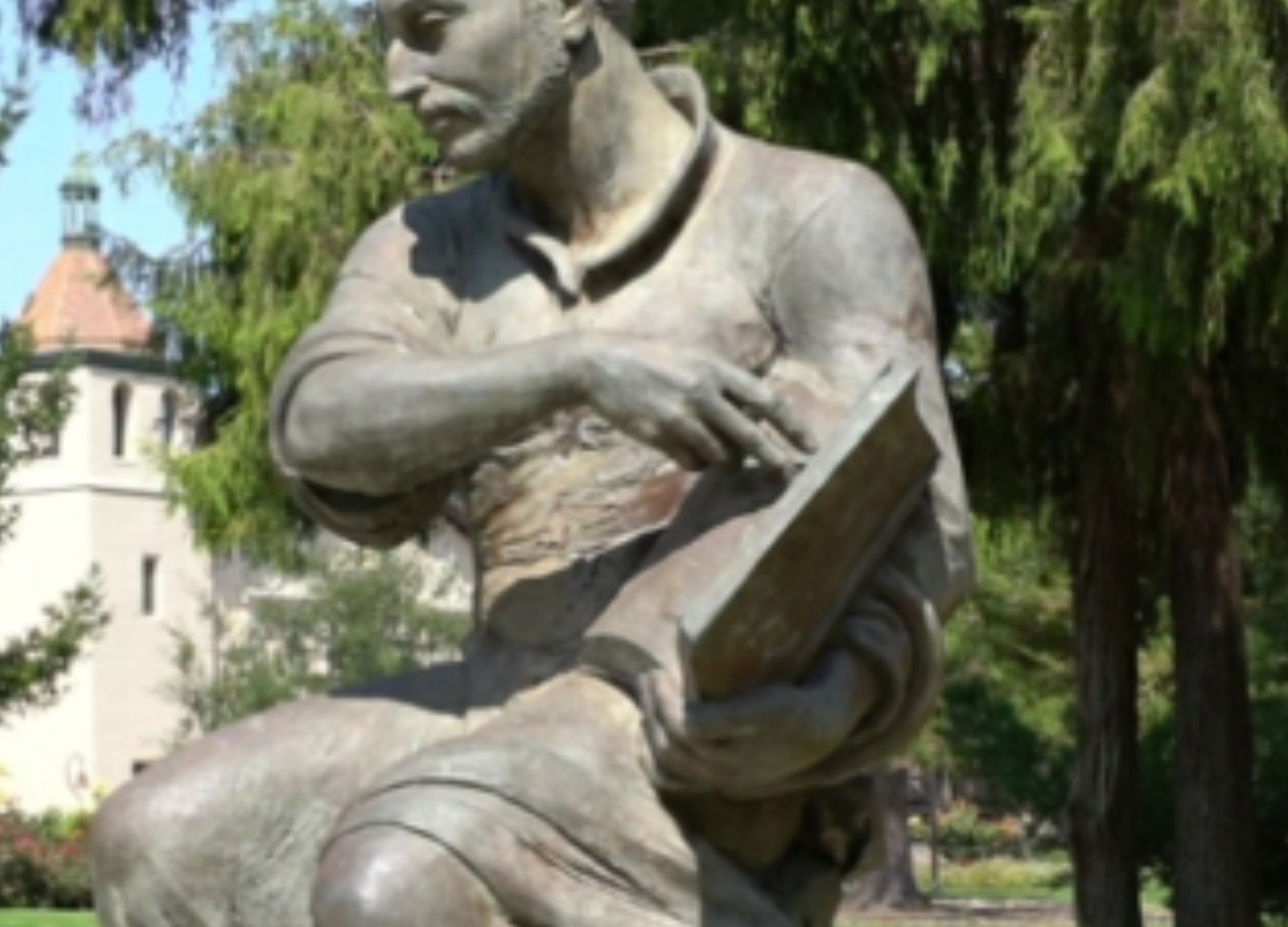}
    \includegraphics[width=0.20\linewidth]{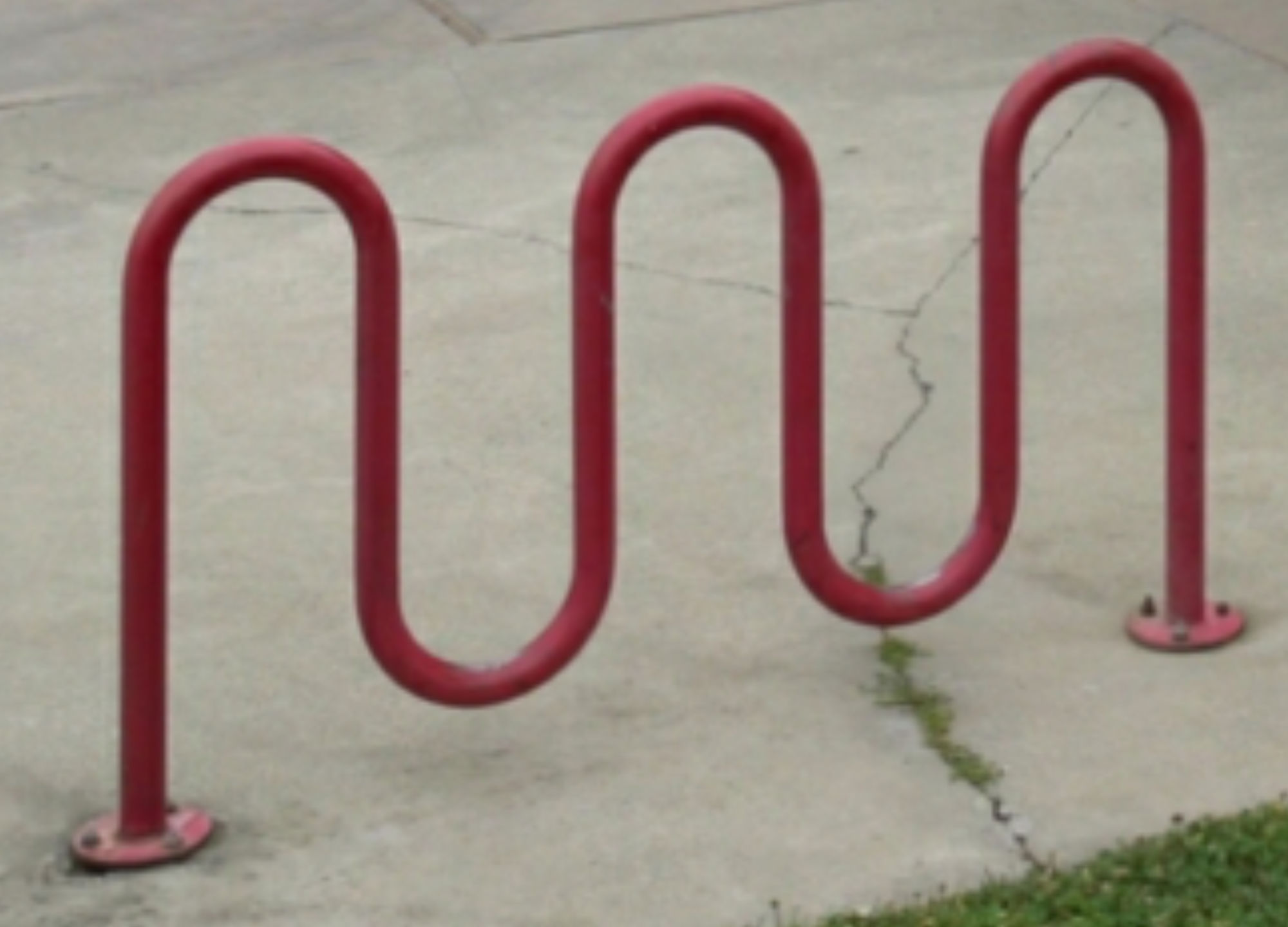}
    \includegraphics[width=0.20\linewidth]{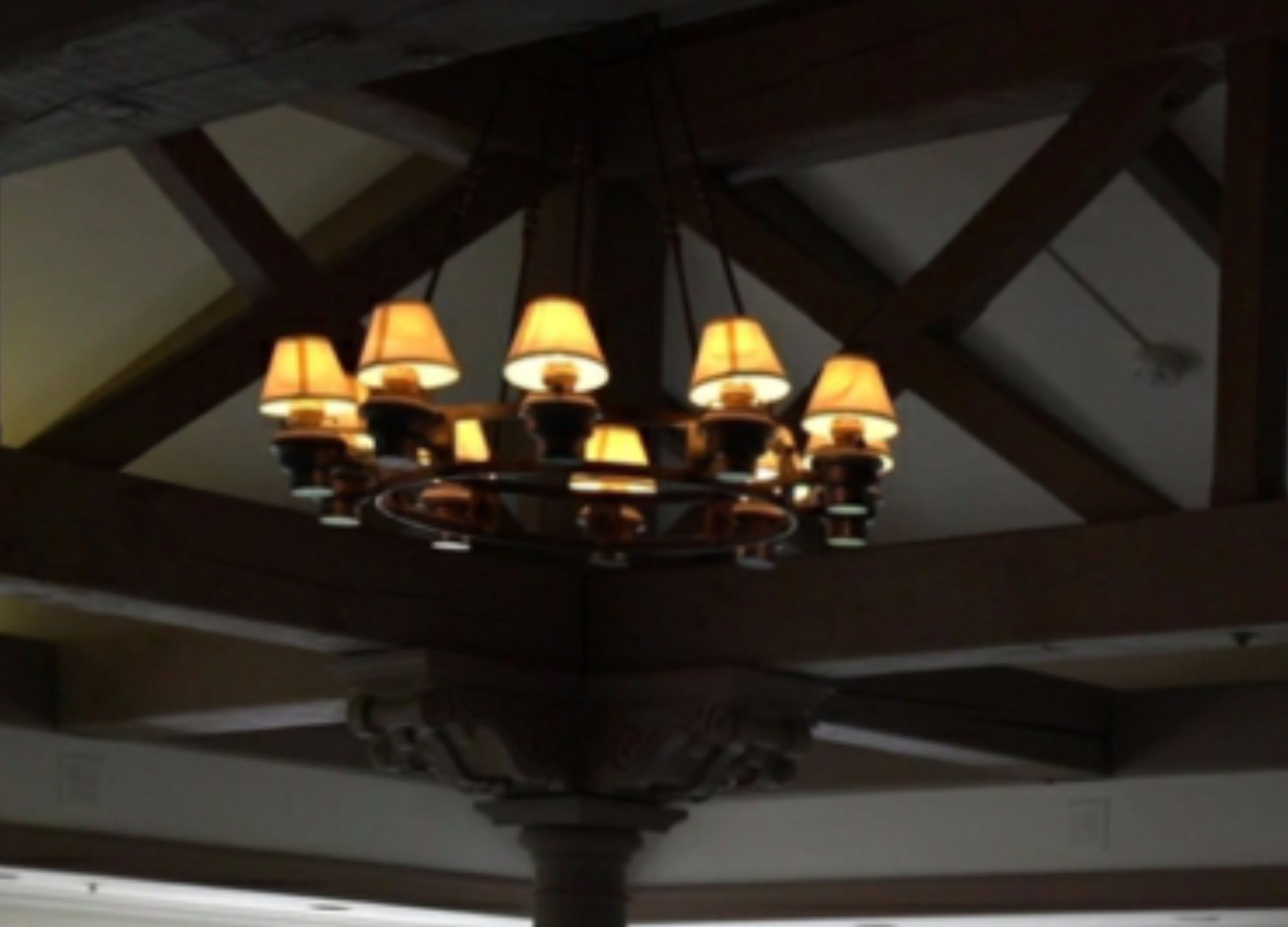}
}
\vspace*{-6mm}
\caption*{Ground Truth Images}
\vspace*{1mm}

\resizebox{\linewidth}{!}{
    \includegraphics[width=0.20\linewidth]{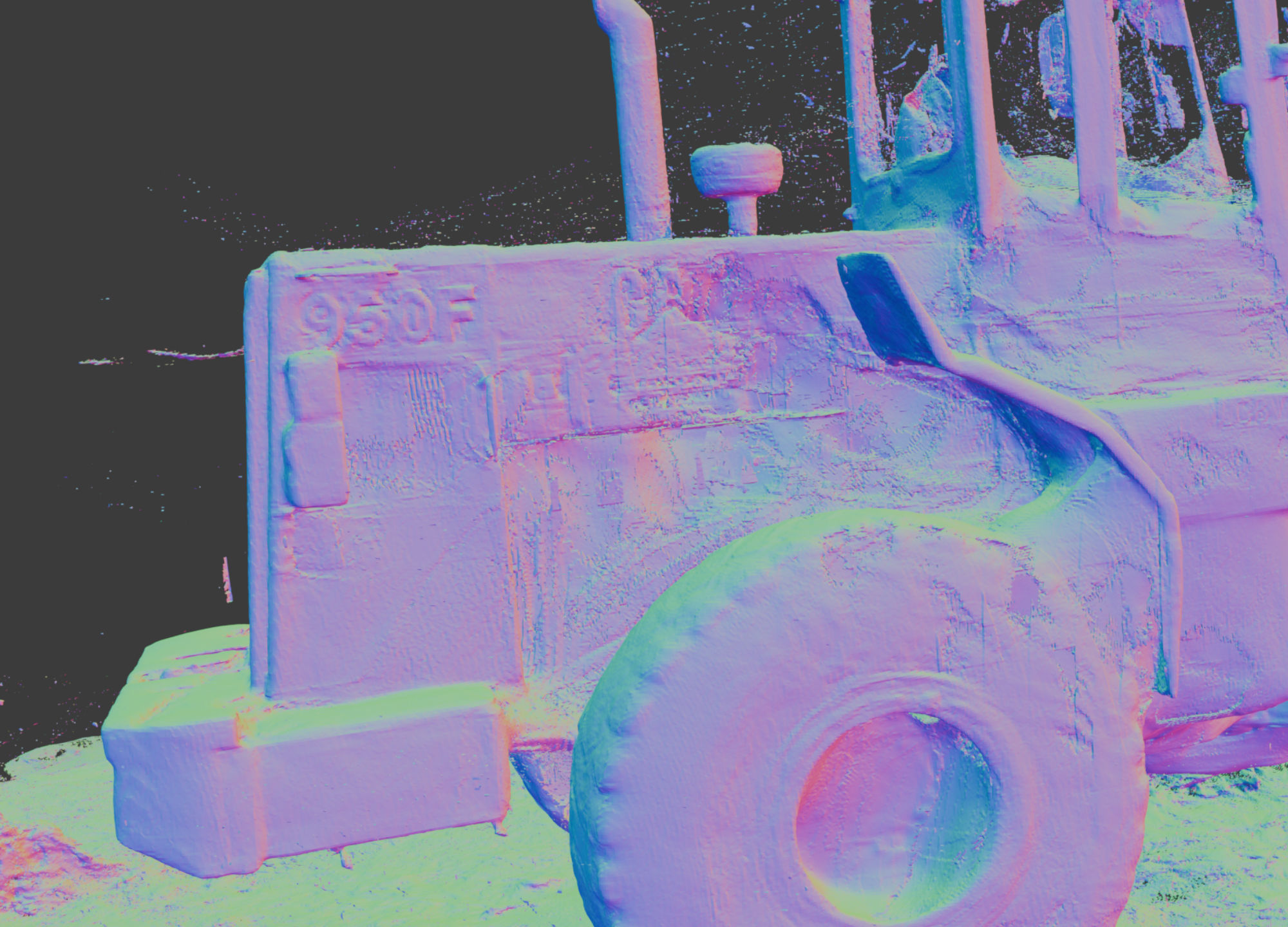} 
    \includegraphics[width=0.20\linewidth]{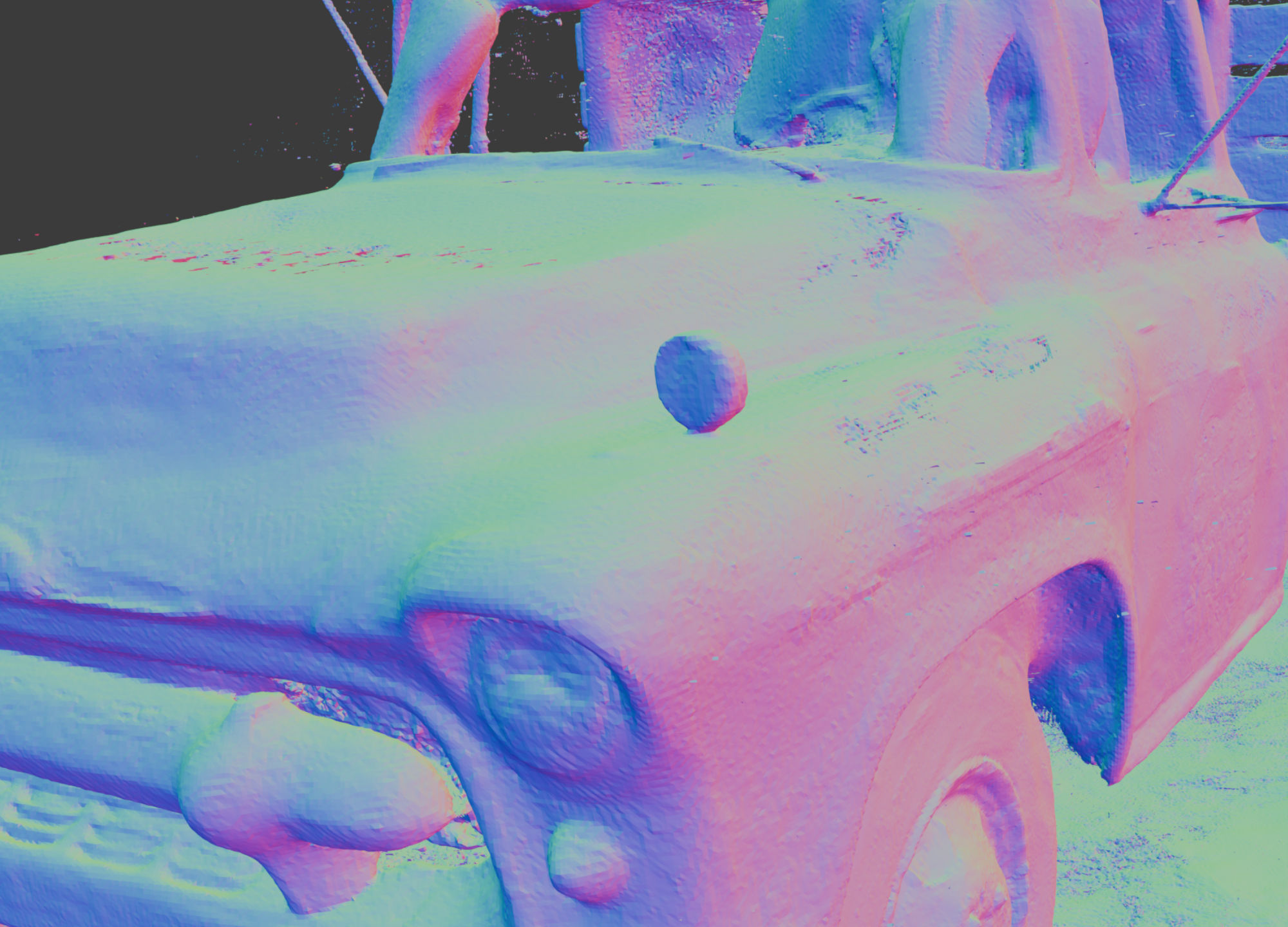} 
    \includegraphics[width=0.20\linewidth]{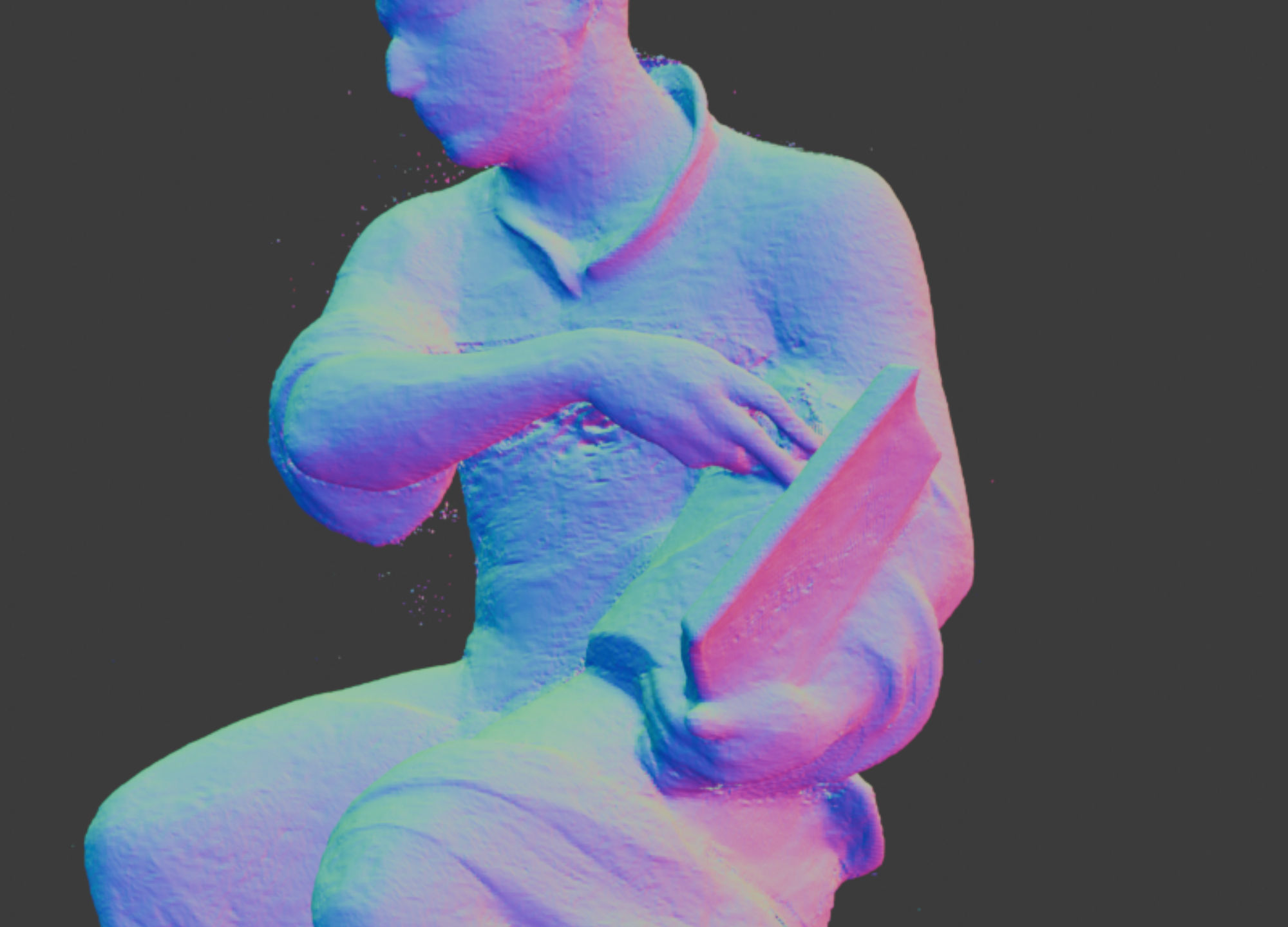} 
    \includegraphics[width=0.20\linewidth]{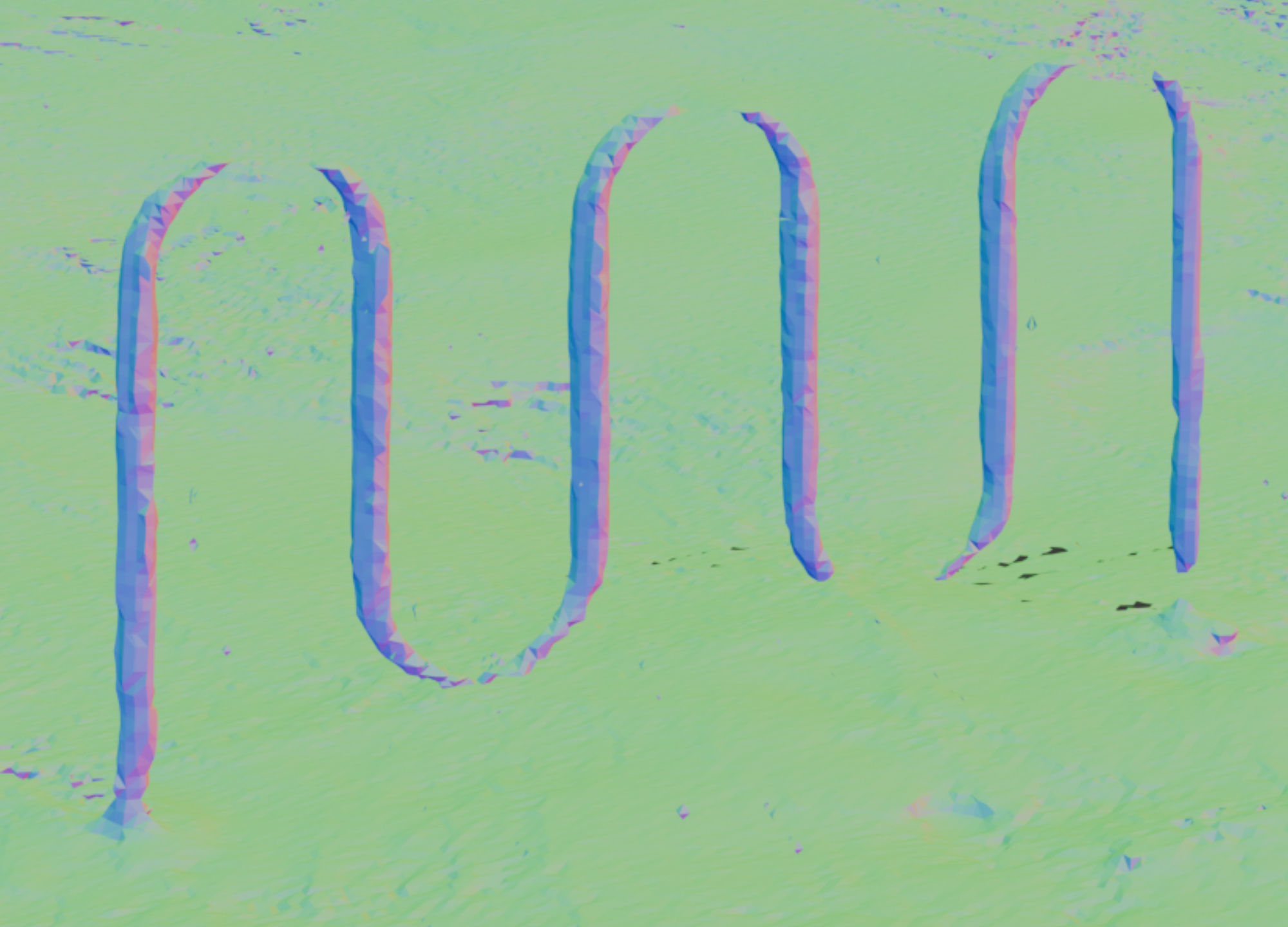}
    \includegraphics[width=0.20\linewidth]{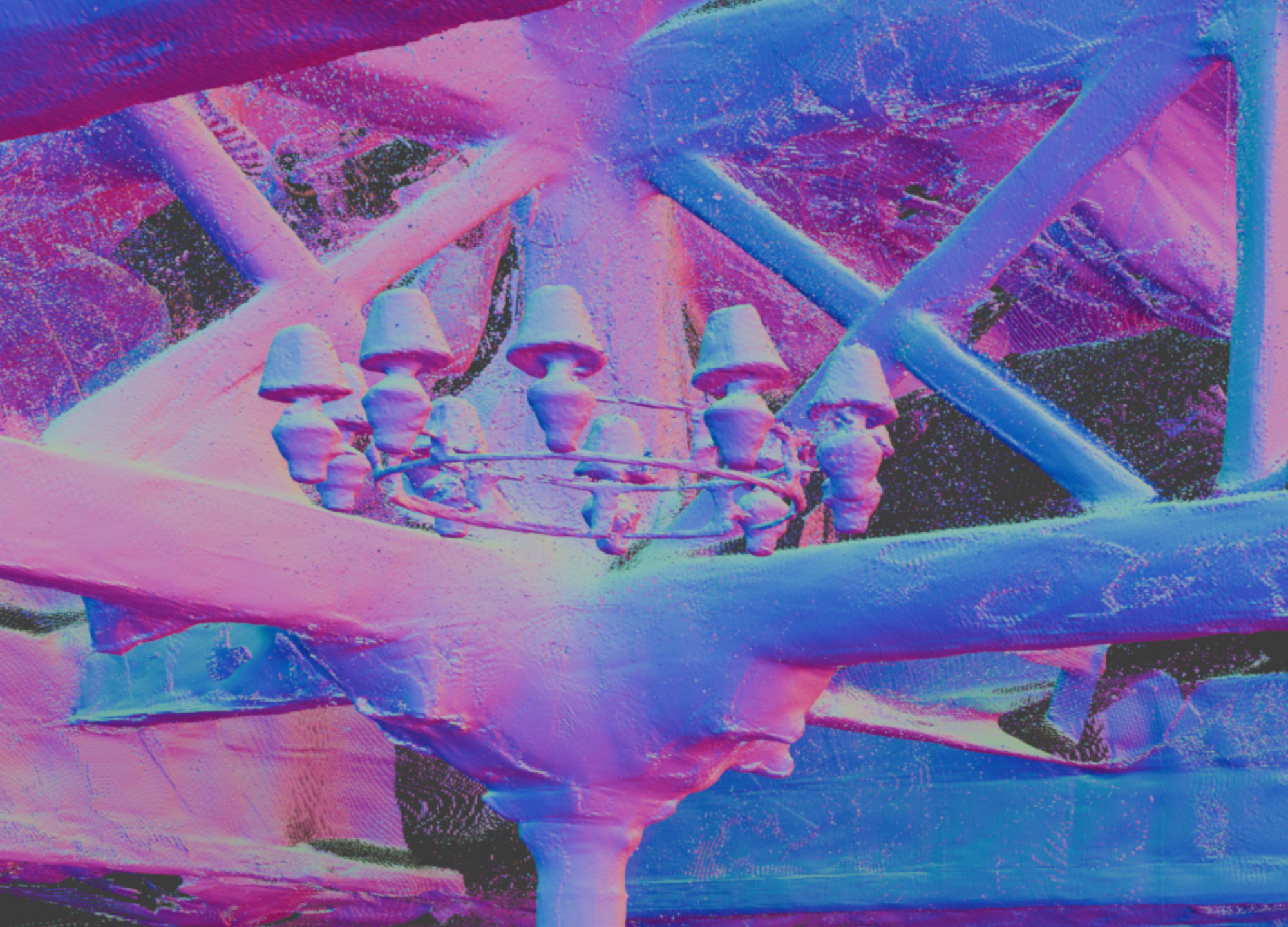}
}
\vspace*{-6mm}
\caption*{(a) 2DGS}
\vspace*{1mm}

\resizebox{\linewidth}{!}{
    \includegraphics[width=0.20\linewidth]{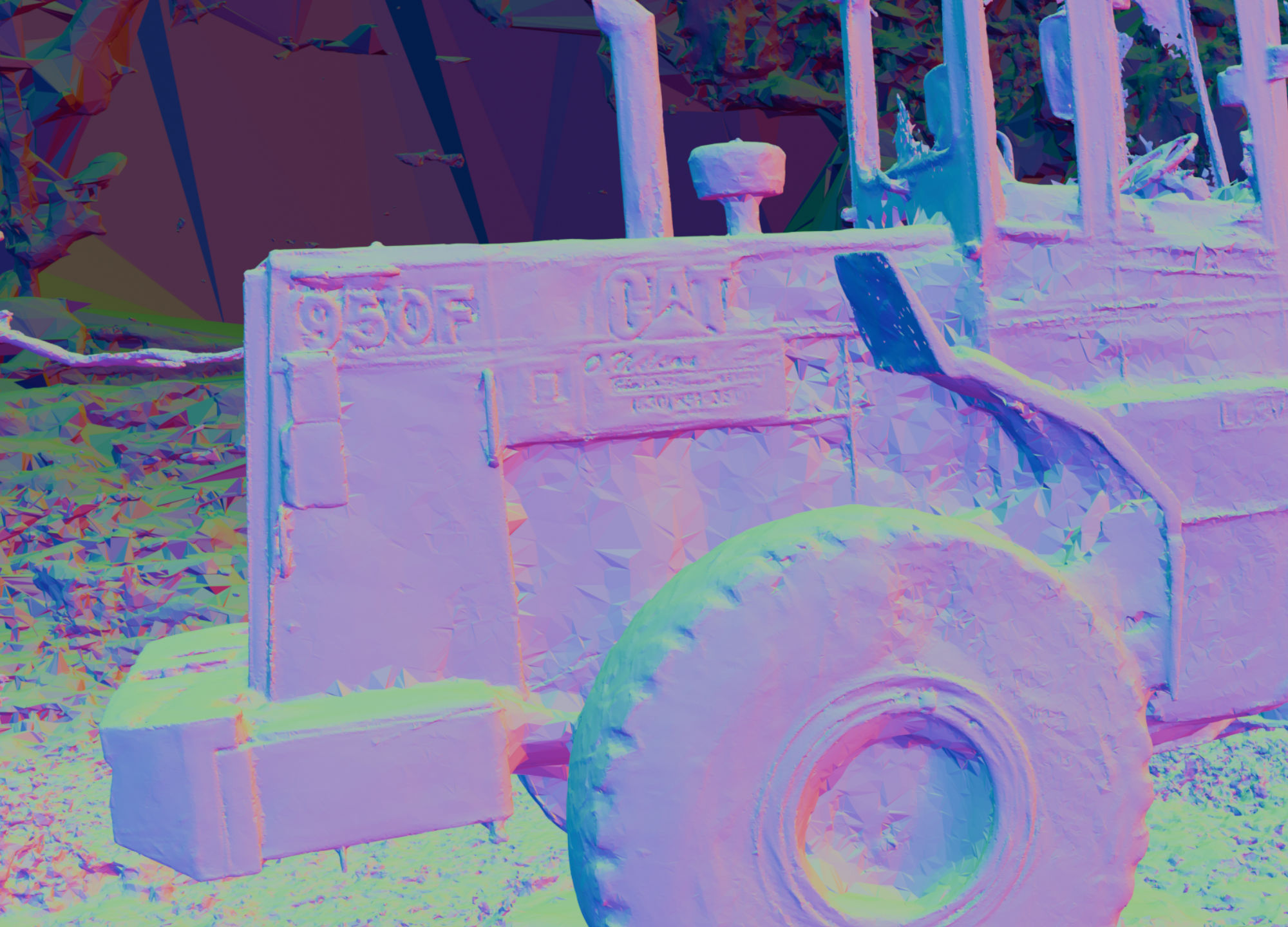}
    \includegraphics[width=0.20\linewidth]{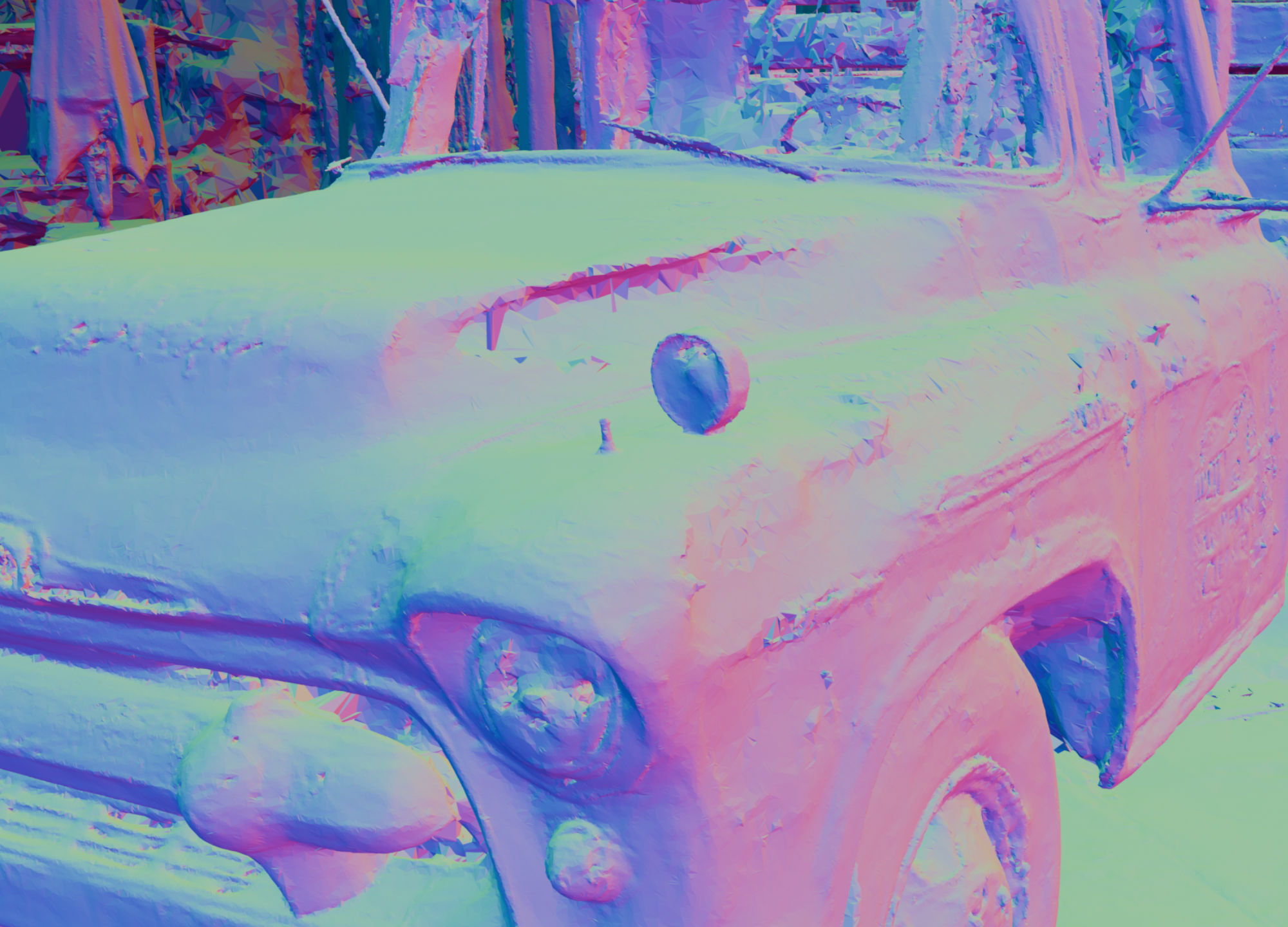} 
    \includegraphics[width=0.20\linewidth]{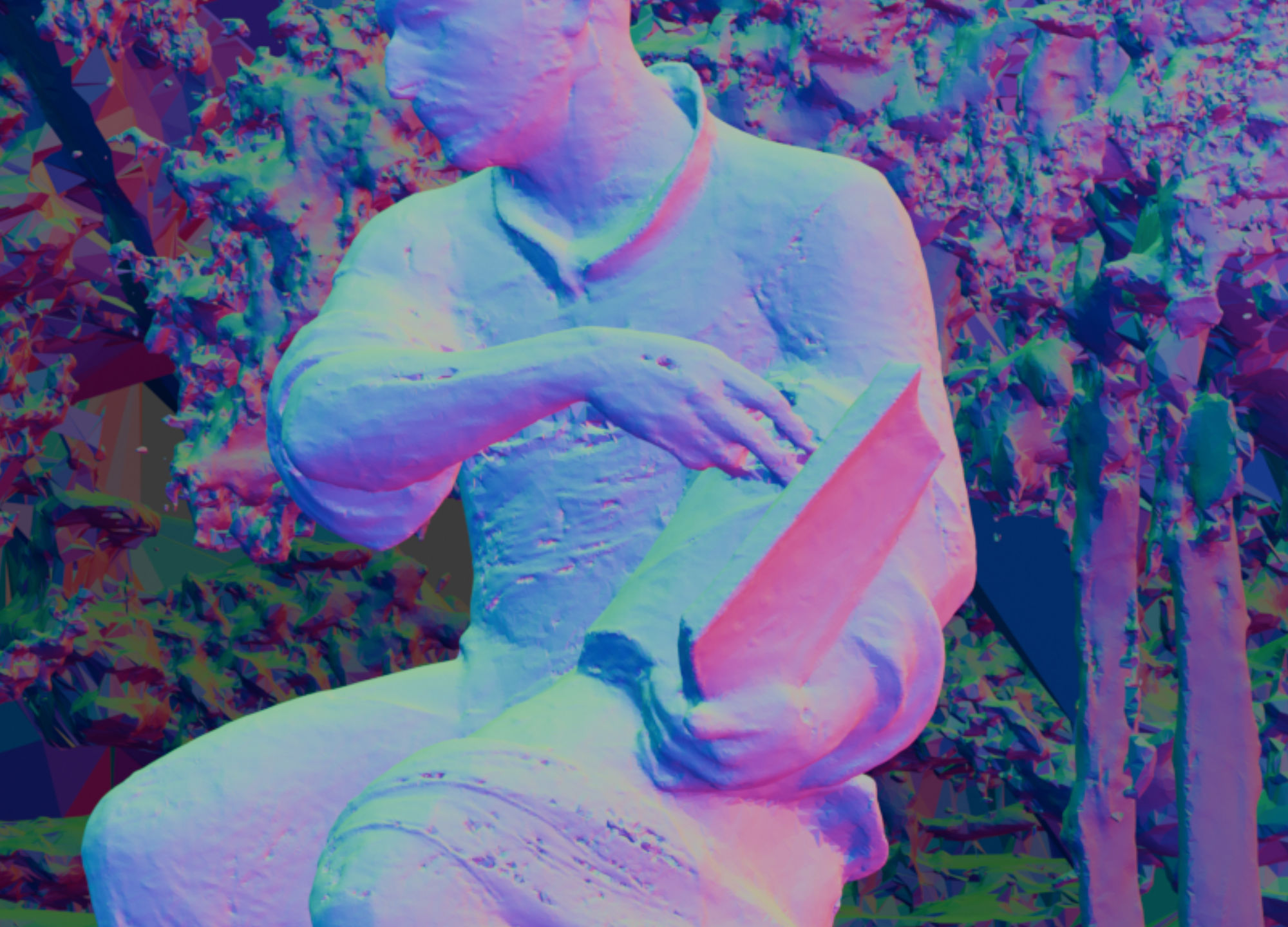} 
    \includegraphics[width=0.20\linewidth]{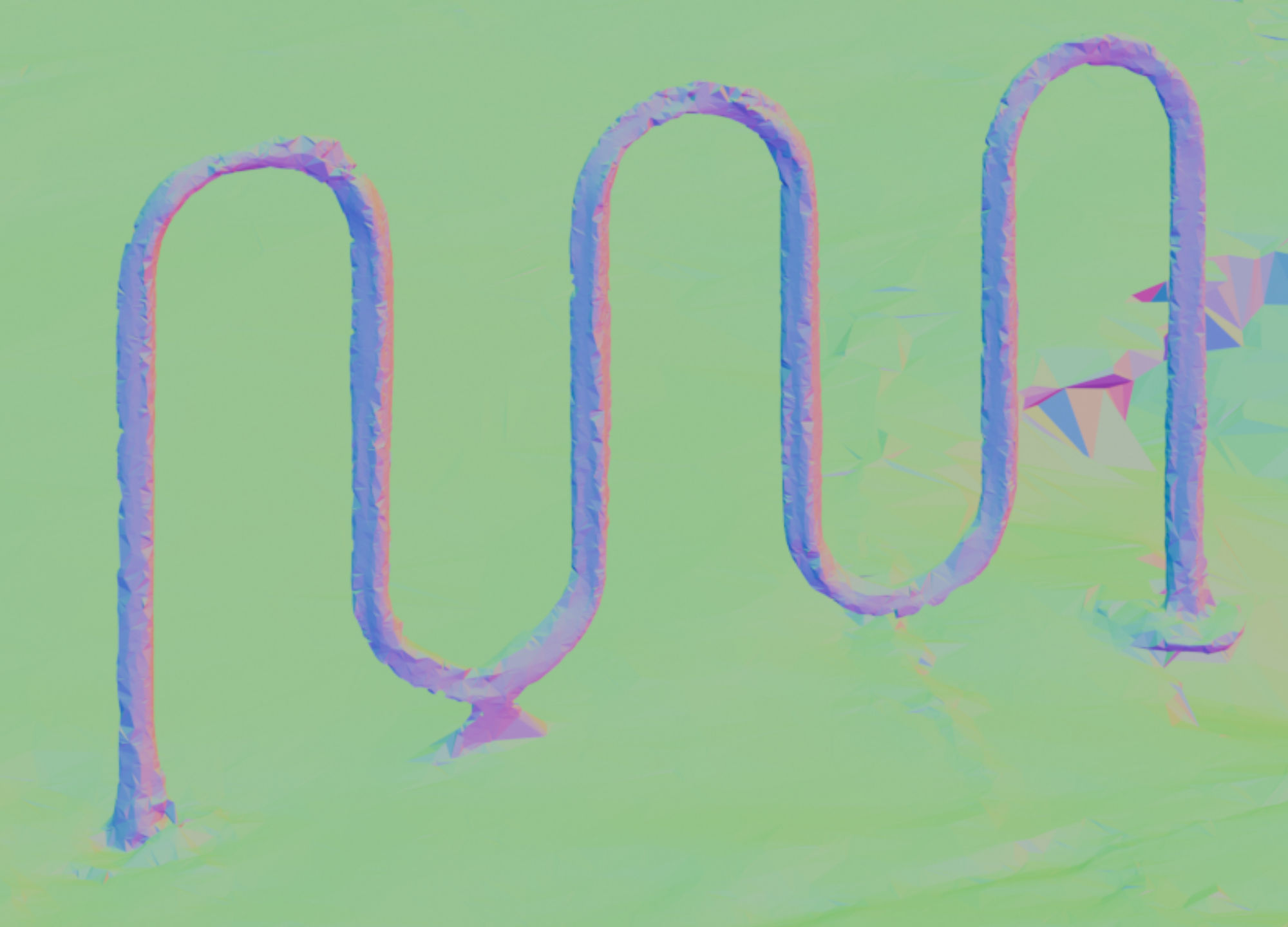}
    \includegraphics[width=0.20\linewidth]{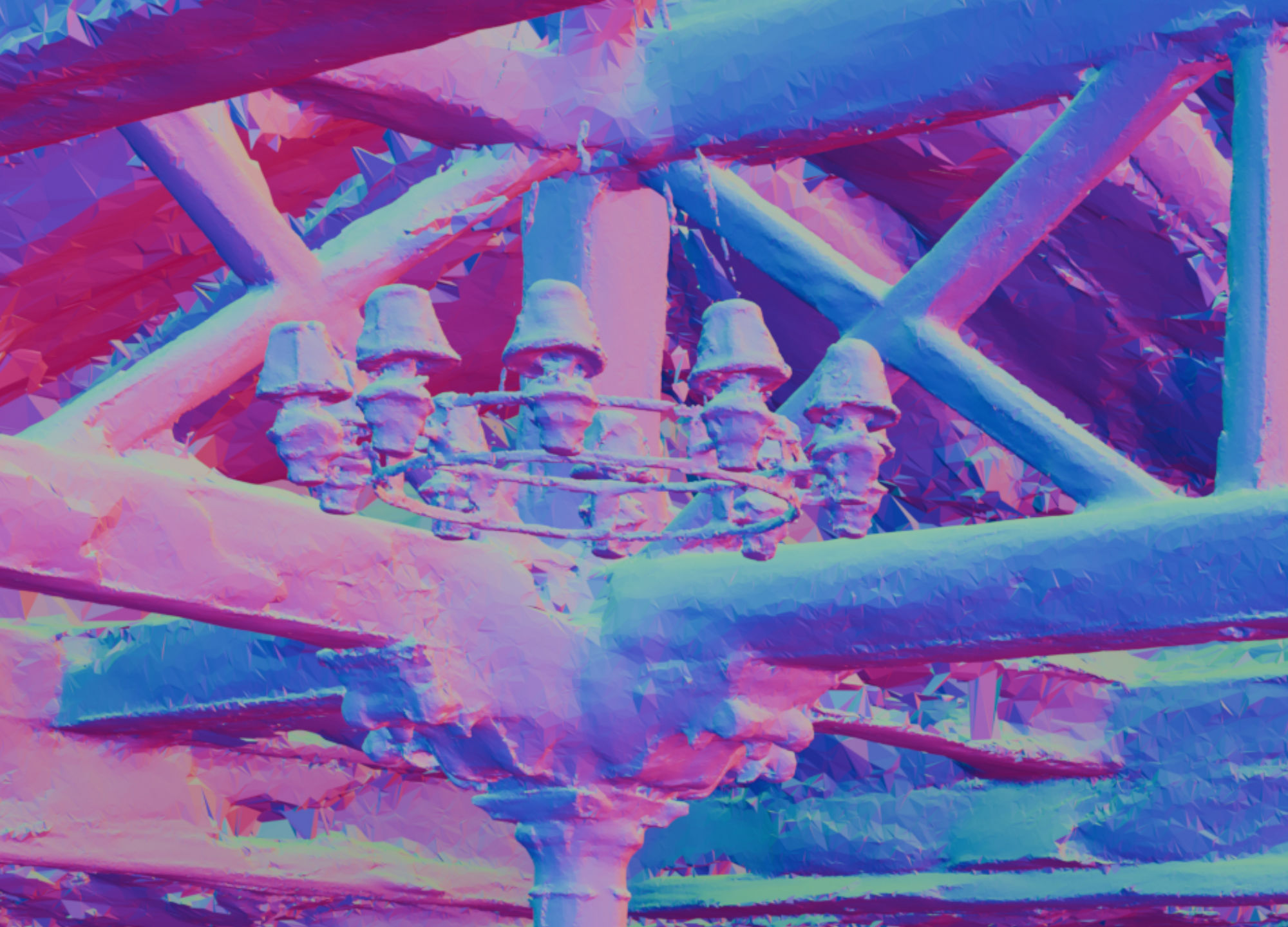}
}
\vspace*{-6mm}
\caption*{(b) GOF} 
\vspace*{1mm}

\resizebox{\linewidth}{!}{
    \includegraphics[width=0.20\linewidth]{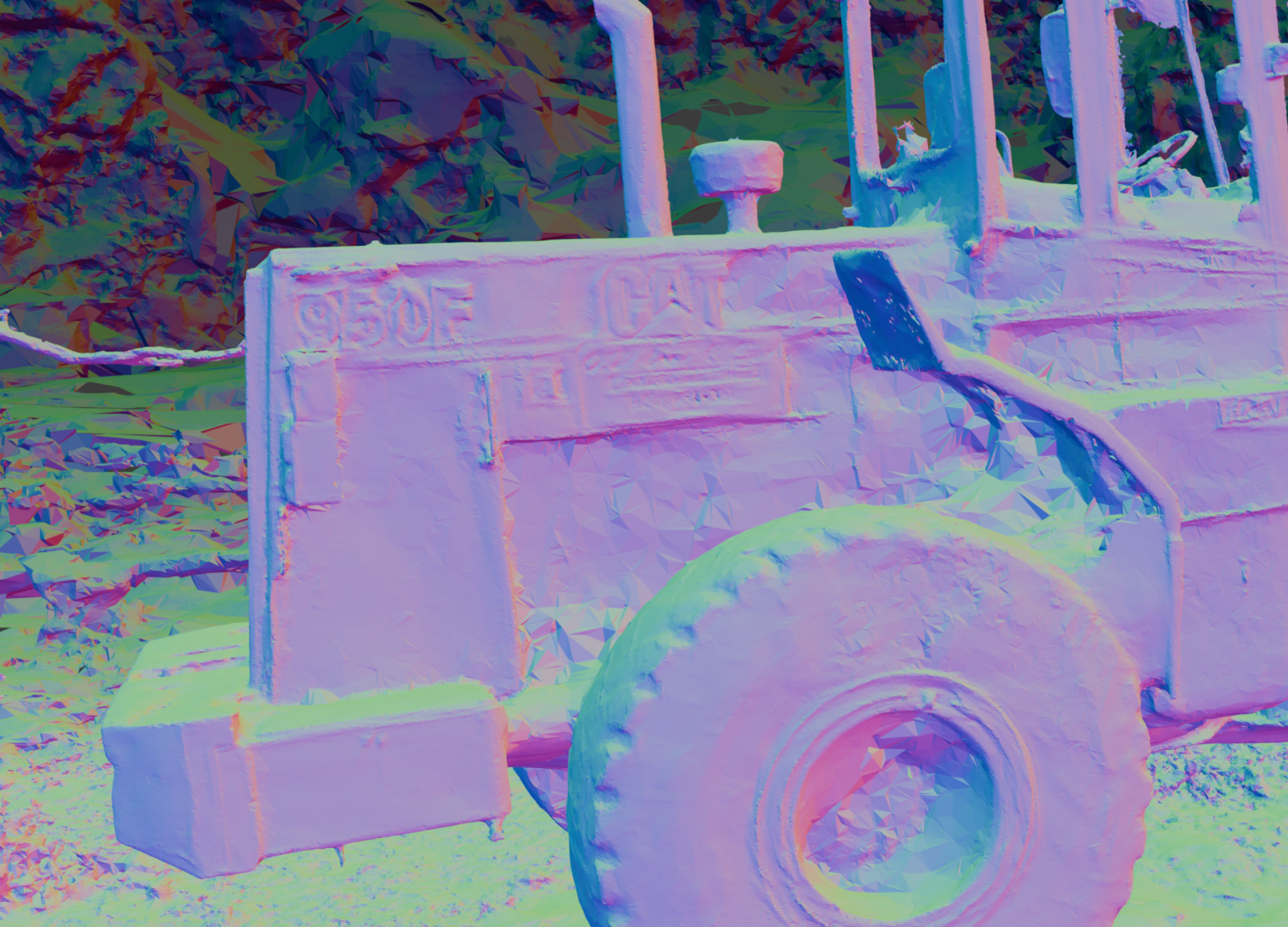} 
    \includegraphics[width=0.20\linewidth]{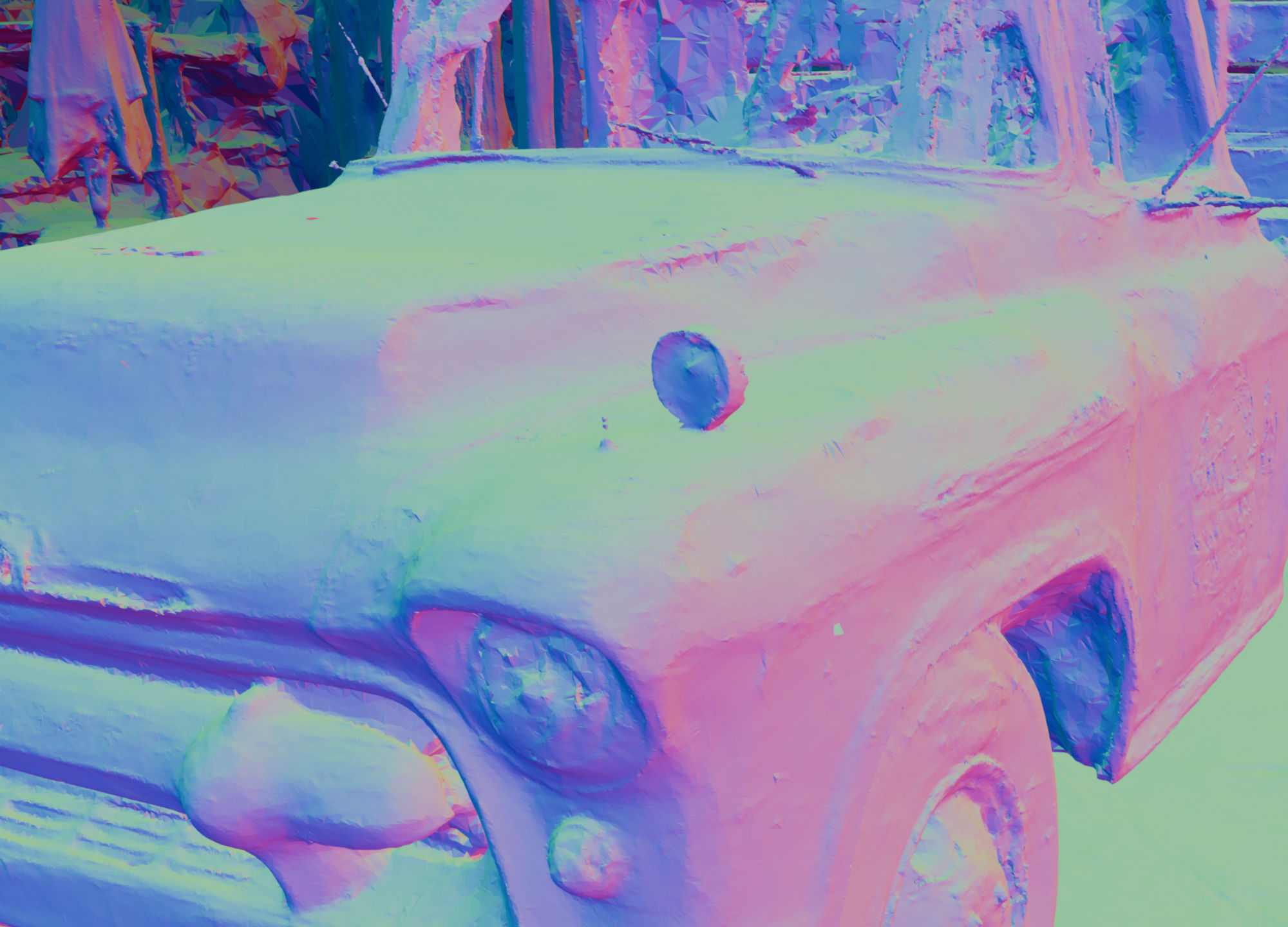} 
    \includegraphics[width=0.20\linewidth]{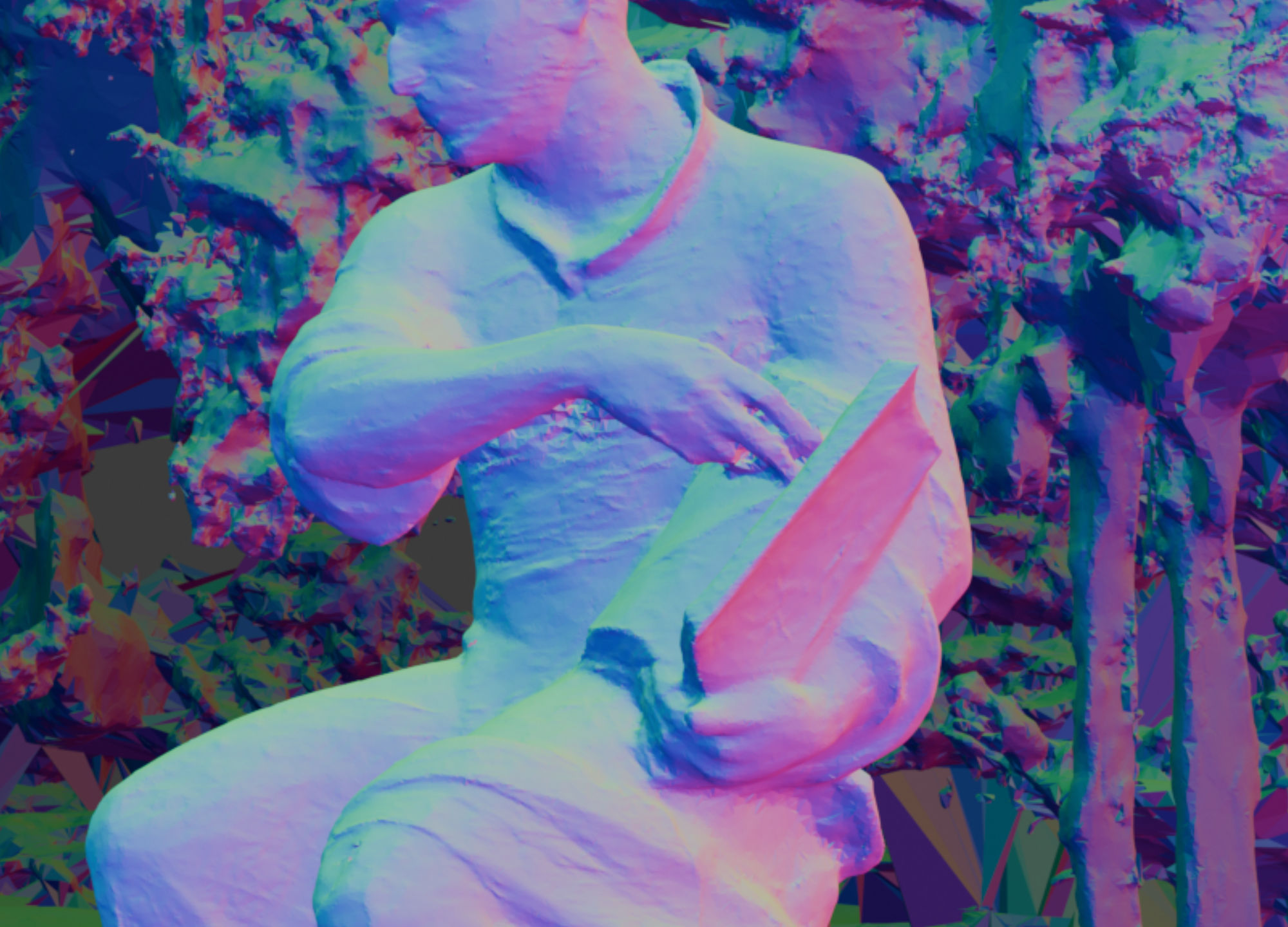}
    \includegraphics[width=0.20\linewidth]{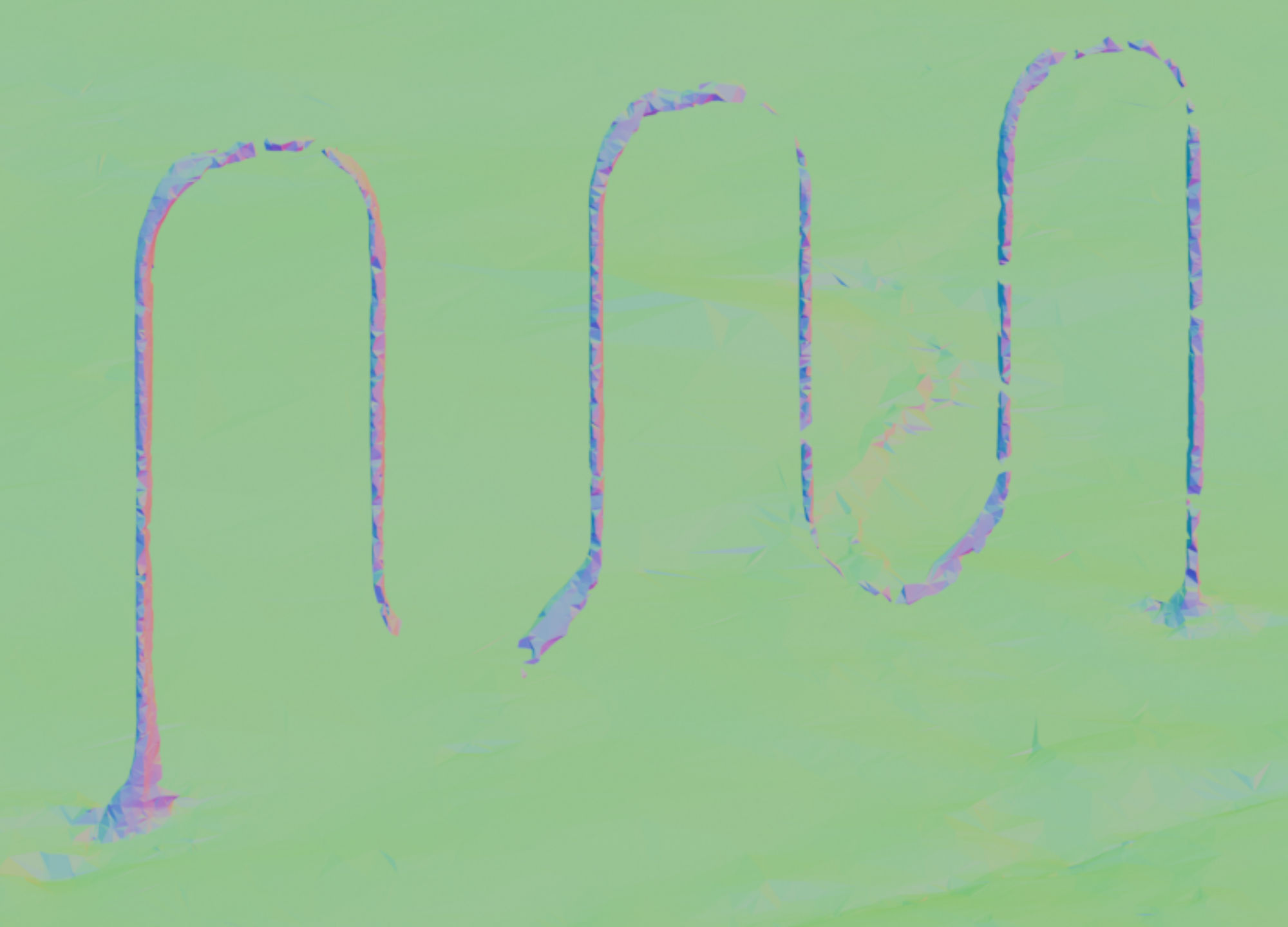}
    \includegraphics[width=0.20\linewidth]{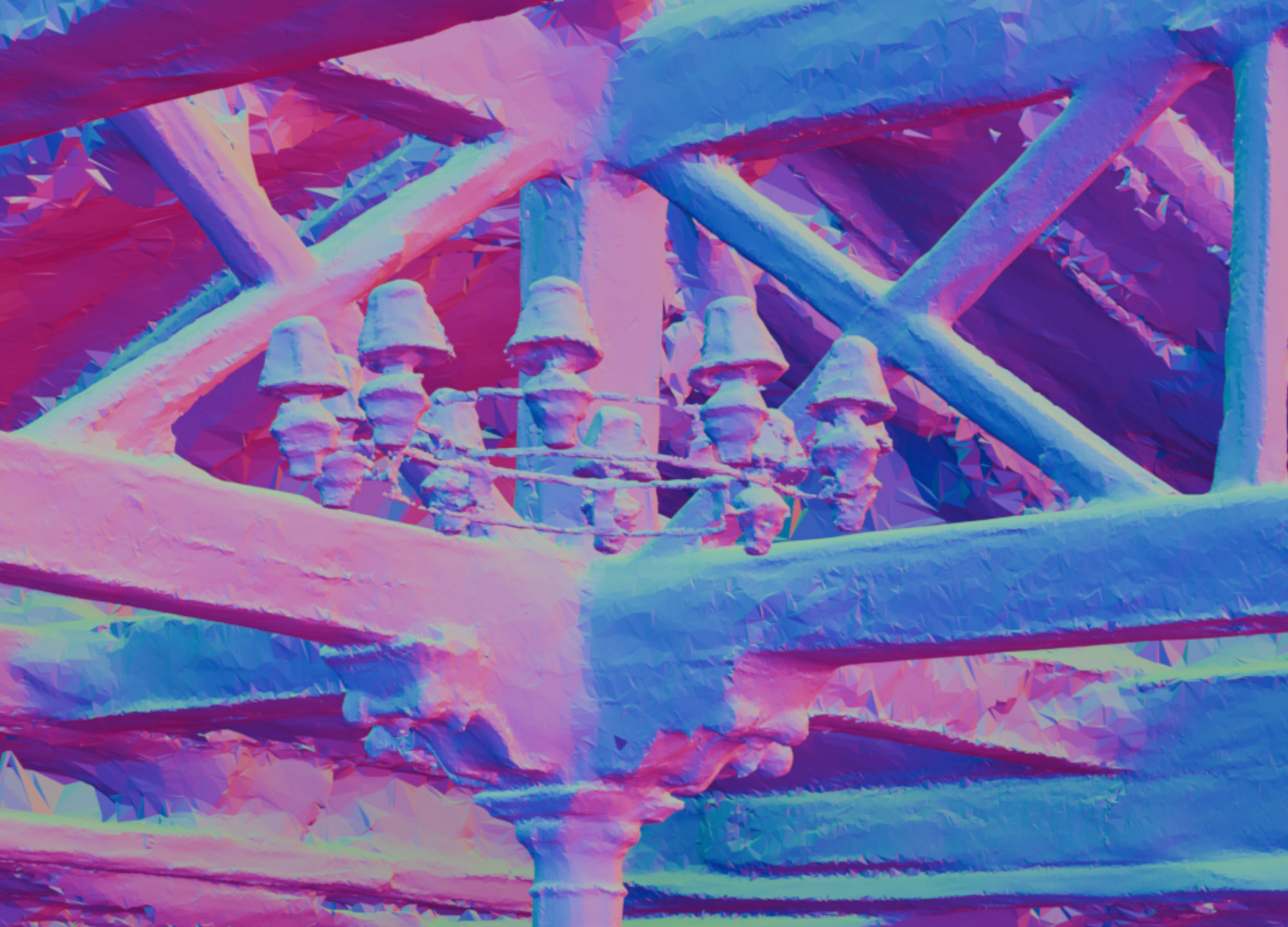}
}
\vspace*{-6mm}
\caption*{(c) RaDe-GS} 
\vspace*{1mm}

\resizebox{\linewidth}{!}{
    \includegraphics[width=0.20\linewidth]{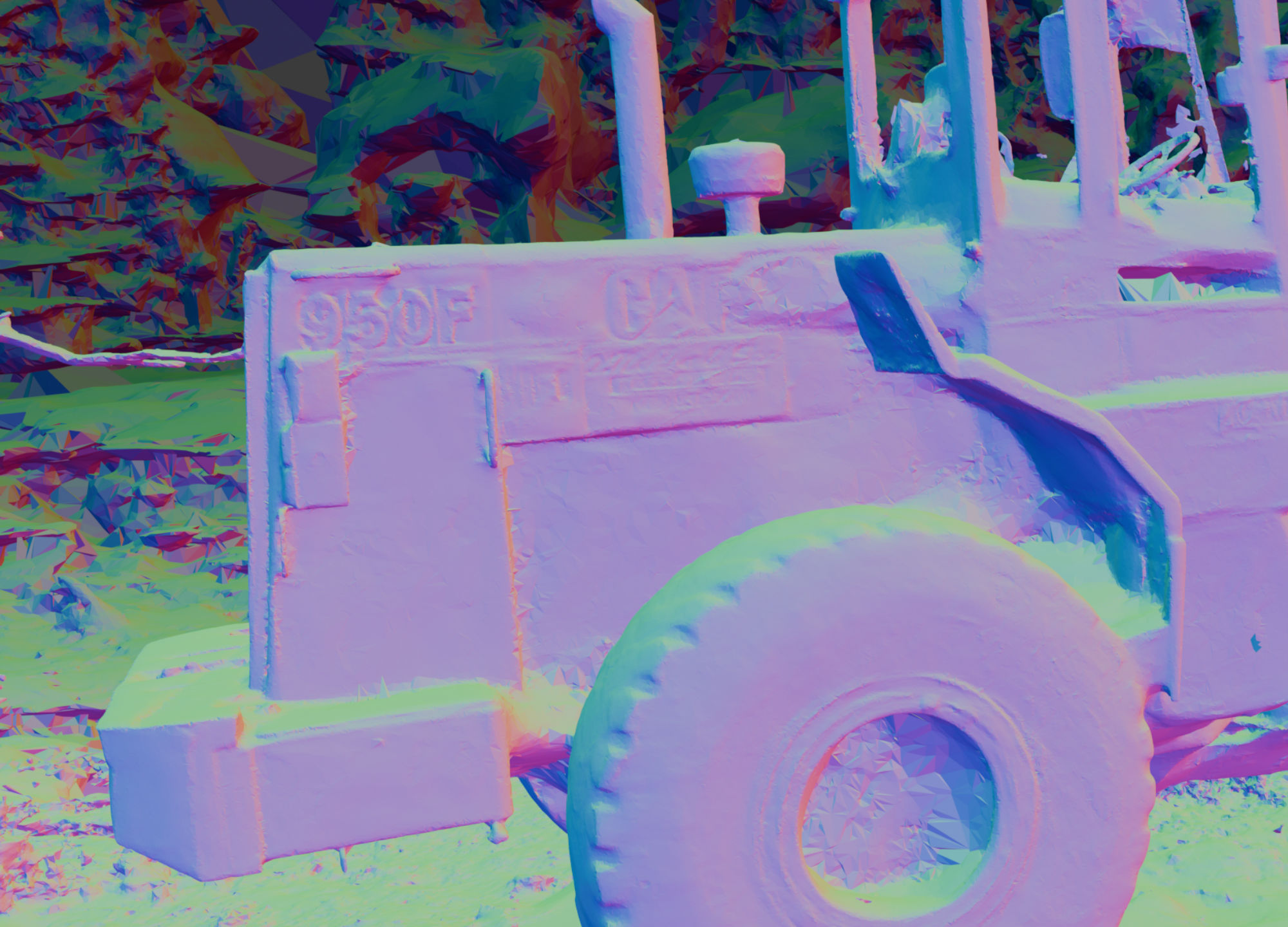} 
    \includegraphics[width=0.20\linewidth]{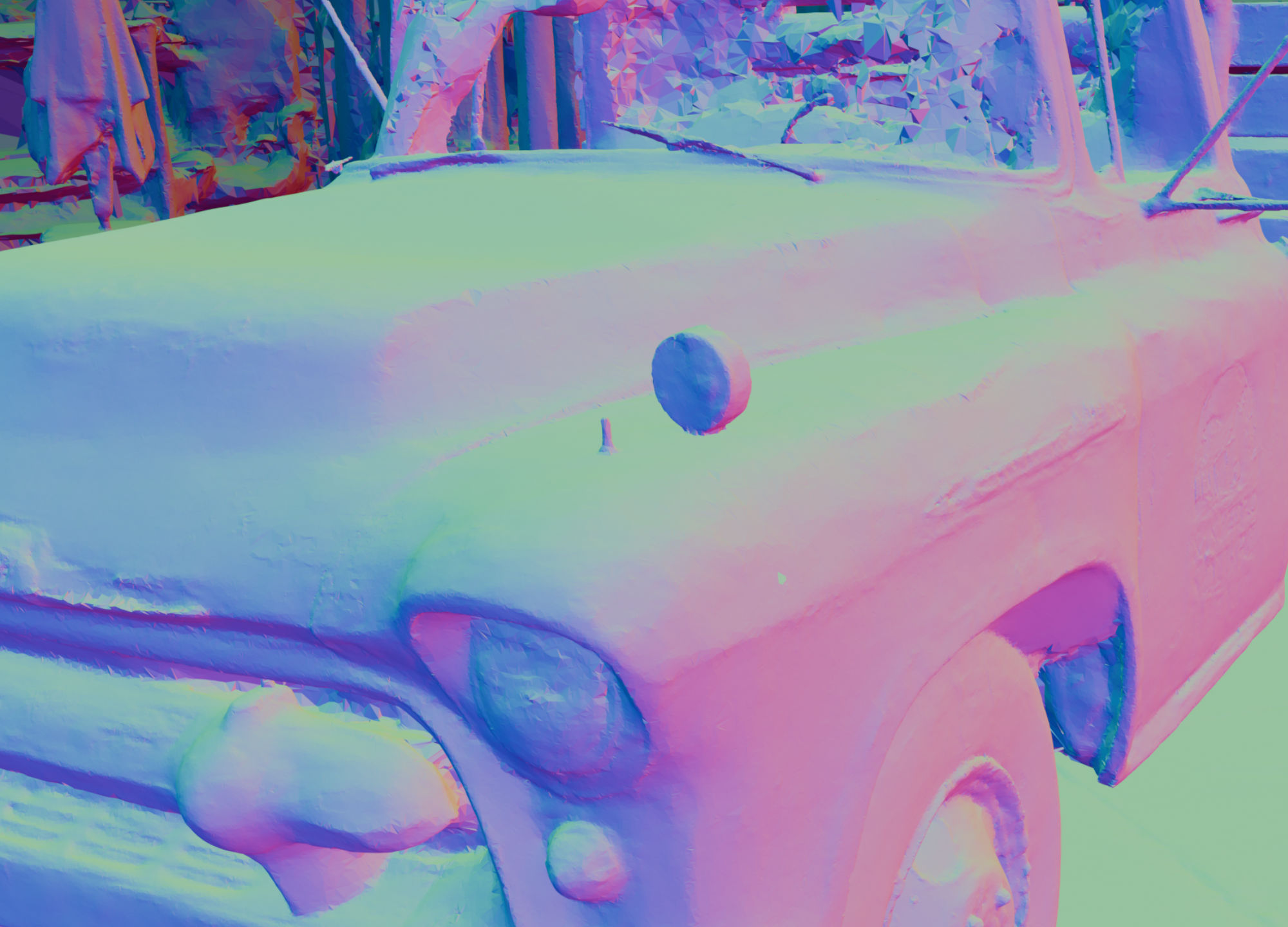} 
    \includegraphics[width=0.20\linewidth]{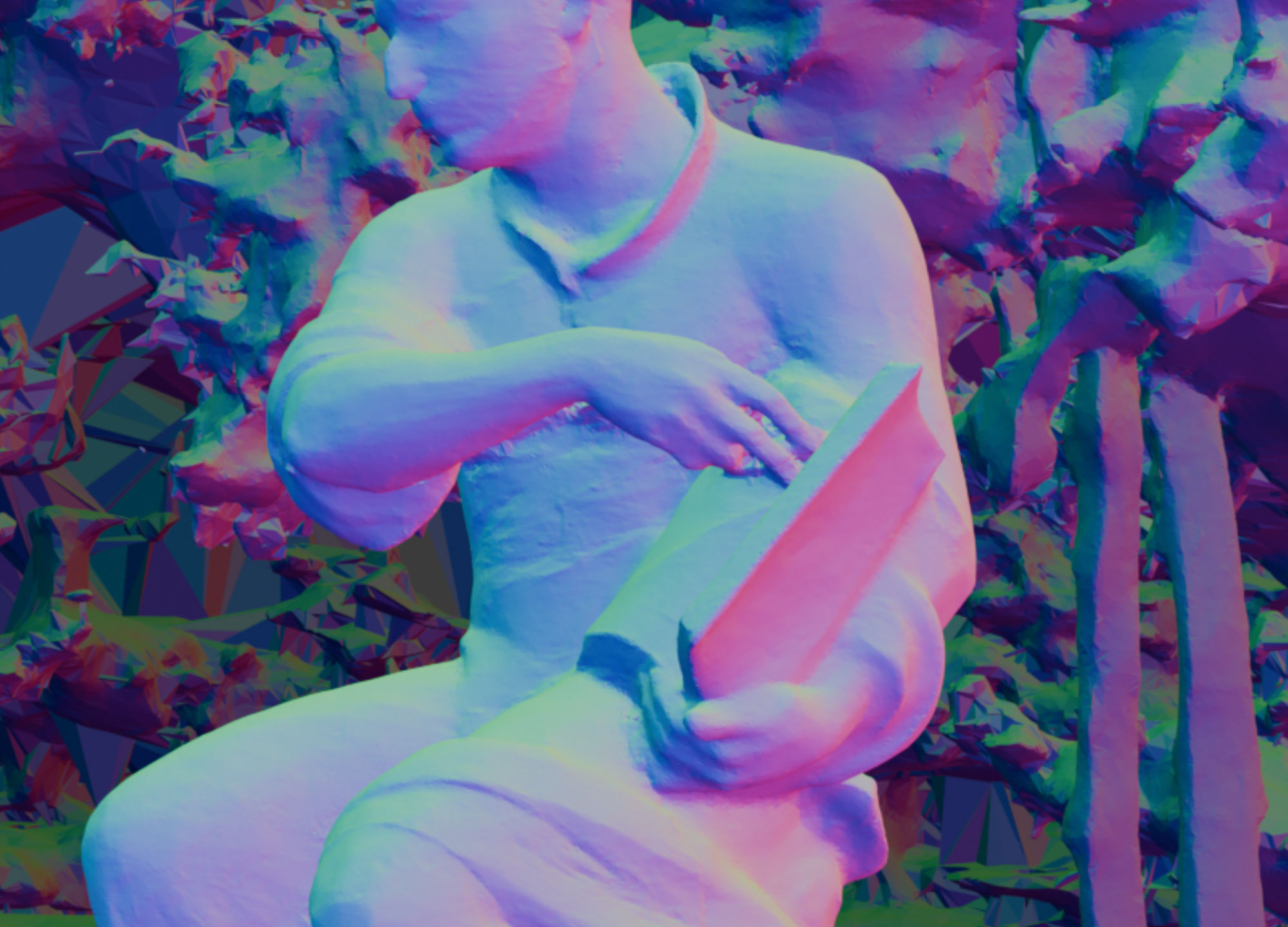} 
    \includegraphics[width=0.20\linewidth]{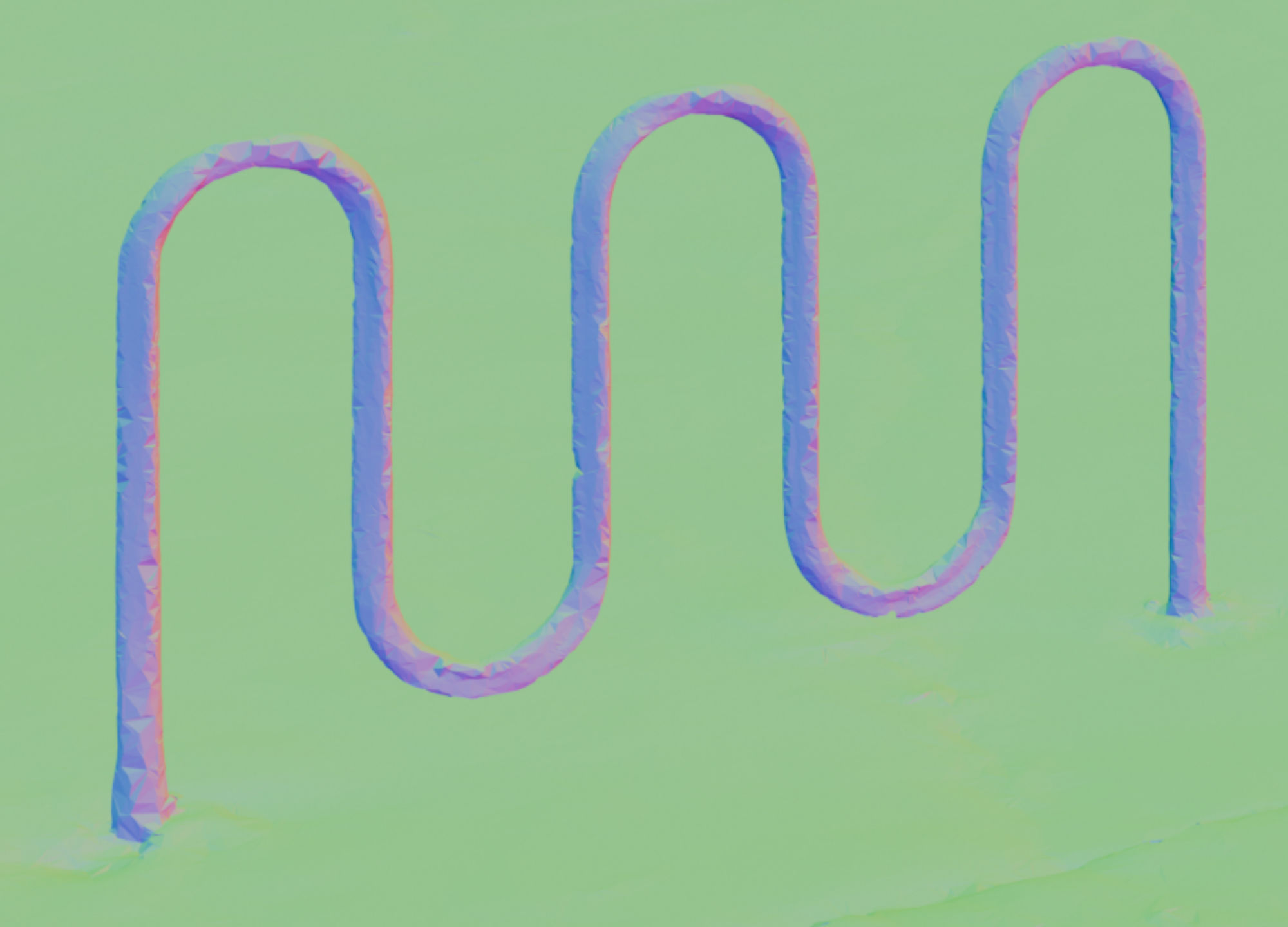}
    \includegraphics[width=0.20\linewidth]{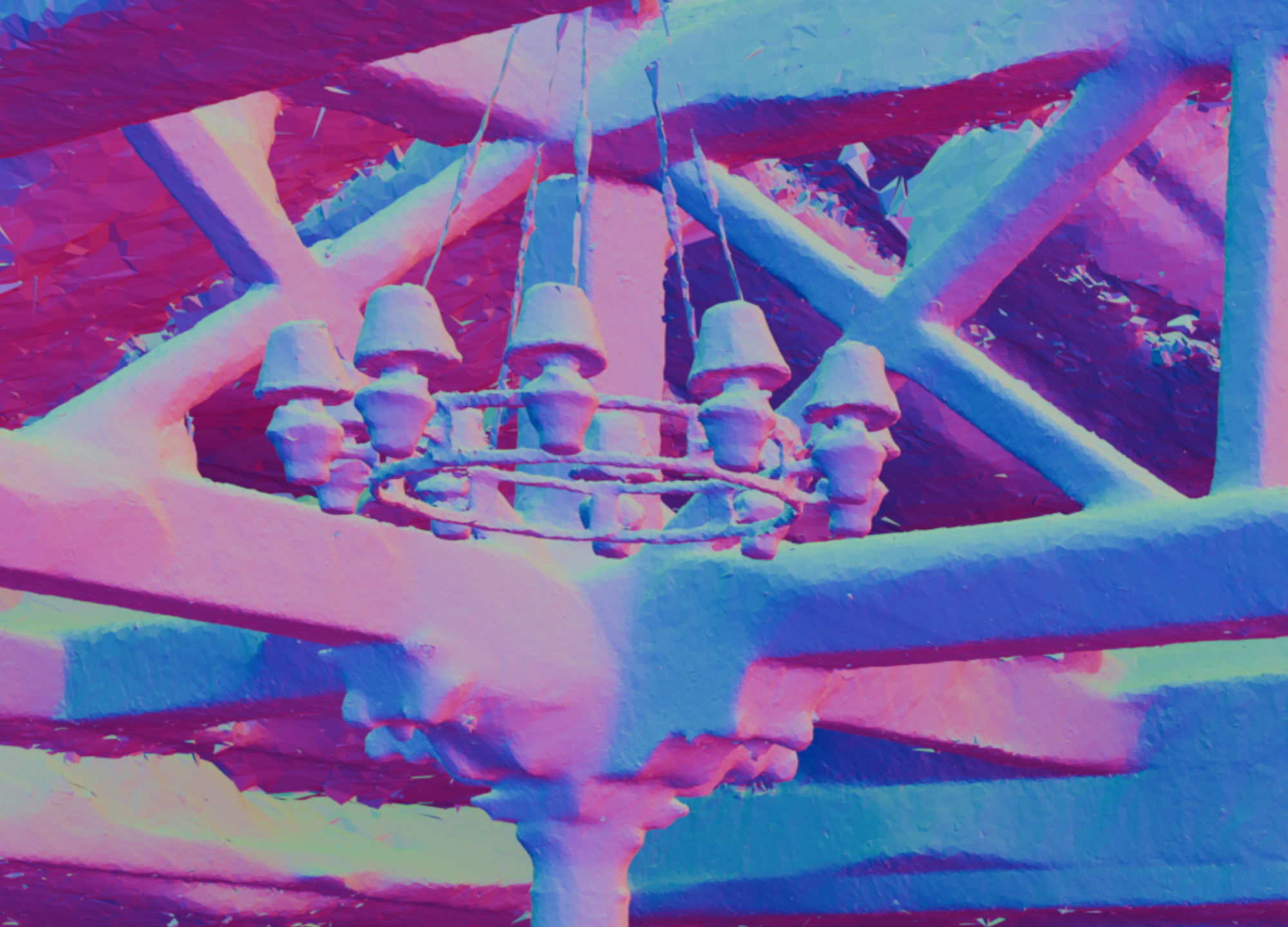}
}
\vspace*{-6mm}
\caption*{(d) ours}

\caption{
\textbf{Qualitative comparison of surface reconstruction results on the Tanks and Temples dataset}. \diego{MILo} produces meshes of full scenes that better fit the target surfaces, with reduced artifacts such as cavities (three left images). Our approach effectively addresses the erosion problem that plagues previous methods (two right images), while producing lighter meshes. \diego{In addition, it is particularly effective at recovering peripheral objects (e.g., the bike rack or chandelier) that are often missed by post-hoc methods.}
}

\vspace*{7mm}

\label{fig:qualitative_comparison}
\end{figure*} 

%% file: tables/surface_metric_dtu.tex
\setlength\tabcolsep{0.5em}
\begin{table*}[!ht]
\centering
\caption{\textbf{Surface reconstruction metrics on the DTU dataset}. We report the Chamfer Distance across 15 scenes. Our method achieves competitive performance with significantly fewer Gaussians and mesh vertices compared to previous approaches.
}
\vspace{-0.2cm}
\resizebox{.98\textwidth}{!}{
\begin{tabular}{@{}llcccccccccccccccclcc}
\hline
 \multicolumn{3}{c}{} & 24 & 37 & 40 & 55 & 63 & 65 & 69 & 83 & 97 & 105 & 106 & 110 & 114 & 118 & 122 & & Mean & Time \\ \cline{4-18} \cline{20-21}
\multirow{4}{*}{\rotatebox[origin=c]{90}{implicit}} & NeRF~\cite{mildenhall2020nerf} & &  1.90 & 1.60 & 1.85 & 0.58 & 2.28 & 1.27 & 1.47 & 1.67 & 2.05 & 1.07 & 0.88 & 2.53 & 1.06 & 1.15 & 0.96 & & 1.49 & \tbestbis $>\text{12h}$\\
& VolSDF~\cite{yariv2021volume_volsdf} & & \tbestbis 1.14 & \sbestbis 1.26 & \sbestbis 0.81 & \tbestbis 0.49 & \tbestbis 1.25 &  \tbestbis 0.70 &  \tbestbis 0.72 &  \sbestbis 1.29 & \tbestbis 1.18 &  \bestbis 0.70 & \tbestbis 0.66 & \sbestbis 1.08 & \tbestbis 0.42 &  \tbestbis 0.61 &  \tbestbis 0.55 & & \tbestbis 0.86 & \tbestbis $>\text{12h}$\\
& NeuS~\cite{wang2021neus} & &  \sbestbis 1.00 & \tbestbis 1.37 & \tbestbis 0.93 & \sbestbis 0.43 & \sbestbis 1.10 &  \sbestbis 0.65 &   \sbestbis 0.57 & \tbestbis 1.48 &  \sbestbis 1.09 &  \tbestbis 0.83 &  \sbestbis 0.52 & \tbestbis 1.20 & \sbestbis 0.35 &  \sbestbis 0.49 &  \sbestbis 0.54 & &  \sbestbis 0.84 & \tbestbis $>\text{12h}$\\
& Neuralangelo~\cite{li2023neuralangelo} & & \bestbis 0.37 & \bestbis 0.72 & \bestbis 0.35 & \bestbis 0.35 & \bestbis 0.87 & \bestbis 0.54 & \bestbis 0.53 &  \bestbis 1.29 & \bestbis 0.97 &  \sbestbis 0.73 & \bestbis 0.47 & \bestbis 0.74 & \bestbis 0.32 & \bestbis 0.41 & \bestbis 0.43 & & \bestbis 0.61 & \tbestbis $>\text{12h}$\\ 
\cline{2-2} \cline{4-18} \cline{20-21}
\multirow{9}{*}{\rotatebox[origin=c]{90}{explicit}} 
&  3D GS~\cite{kerbl3Dgaussians} & & 2.14 & 1.53 & 2.08 & 1.68 & 3.49 & 2.21 & 1.43 & 2.07 & 2.22 & 1.75 &  1.79 & 2.55 & 1.53 & 1.52 & 1.50 & & 1.96 & \best 5.2m\\
&  SuGaR~\cite{guedon2024sugar} & & 1.47 & 1.33 & 1.13 & 0.61 & 2.25 & 1.71 & 1.15 & 1.63 & 1.62 & 1.07 & 0.79 & 2.45 & 0.98 & 0.88 & 0.79 & & 1.33  & 52m \\
& 2D GS~\cite{huang20242d} &&  \tbest 0.48 & 0.91 &  0.39 &  0.39 & \tbest 1.01 &  0.83 &  0.81 &  1.36 & \tbest 1.27 &  0.76  & \sbest 0.70 &  1.40 &  \tbest 0.40 &  \tbest 0.76 &  0.52 &&   0.80 & \tbest 8.9m \\
& GOF~\cite{yu2024gaussian} & &  0.50 & \tbest 0.82 & \tbest 0.37 & \best 0.37 & 1.12 & \best 0.74 & \sbest 0.73 & \best 1.18 & 1.29 & \tbest 0.68 & 0.77 & \tbest 0.90 & 0.42 & \best 0.66 & \tbest 0.49 && \tbest 0.74 & 55m\\
& RaDe-GS~\cite{zhang2024rade} & &  \sbest 0.46 &  \best 0.73 &  \tbest 0.33 &  \tbest 0.38 & \best 0.79 & \tbest 0.75 & \tbest 0.76 & \sbest 1.19 & \best 1.22 &  \best 0.62 & \sbest 0.70 & \best 0.78 & \best 0.36 & \sbest 0.68 &  \best 0.47 && \best 0.68  & \sbest 8.3m\\
& \textbf{Ours (base)} & & \best 0.43 & \sbest 0.74 & \sbest 0.34 & \best 0.37 & \sbest 0.80 & \best 0.74 & \best 0.70 & \tbest 1.21 & \best 1.22 & \sbest 0.66 & \best 0.62 & \sbest 0.80 & \sbest 0.37 & \tbest 0.76 & \sbest 0.48 && \best 0.68 & 25m \\
 \hline
\end{tabular}
}
\label{tab:surface_metrics_dtu}
\vspace{-0.1cm}
\end{table*}

%% file: tables/rendering_metrics_mesh.tex
\begin{table*}[!ht]
    \centering
    \caption{\textbf{Mesh-Based Novel View synthesis metrics on real-world datasets}. We report PSNR, SSIM, and LPIPS metrics, as well as the total number of Gaussians and vertices (in millions). Our method demonstrates superior performance on these challenging unbounded scenes. \antoine{Ours (base) uses the differentiable rasterizer from RaDe-GS.}}
    \vspace{-0.2cm}
    \resizebox{0.99\linewidth}{!}{
    \begin{tabular}{@{}l|ccccc|ccccc|ccccc}
     & \multicolumn{5}{c@{}|}{MipNeRF~360} & \multicolumn{5}{c@{}|}{Tanks~\&~Temples} & \multicolumn{5}{c@{}}{DeepBlending} \\ 
     & PSNR $\uparrow$ & SSIM $\uparrow$ & LPIPS $\downarrow$ & \#Gaussians $\downarrow$ & \#Verts $\downarrow$ & PSNR $\uparrow$ & SSIM $\uparrow$ & LPIPS $\downarrow$ & \#Gaussians $\downarrow$ & \#Verts $\downarrow$ & PSNR $\uparrow$ & SSIM $\uparrow$ & LPIPS $\downarrow$ & \#Gaussians $\downarrow$ & \#Verts $\downarrow$ \\ 
     \hline
     2DGS 
     & 15.36 & 0.4987 & 0.4749 & \sbest 1.88 & \best 4.31
     & 14.23 & 0.5697 & 0.4854 & \sbest 0.98 & 16.39 
     & 17.46 & 0.7133 & 0.5556 & \sbest 1.51 & 16.21 \\
     GOF 
     & \tbest 20.78 & \tbest 0.5730 & \tbest 0.4654 & 2.99 & \tbest 32.80 
     & \best 21.69 & \sbest 0.6904 & \best 0.3261 & 1.25 & \tbest 11.63 
     & \sbest 27.44 & \sbest 0.8078 & \sbest 0.2549 & \tbest 1.64 & \tbest 12.64 \\
     RaDe-GS 
     & \sbest 23.56 & \sbest 0.6685 & \sbest 0.3612 & \tbest 2.91 & 30.85 
     & \tbest 20.51 & \tbest 0.6595 & \sbest 0.3447 & \tbest 1.18 & \sbest 10.06 
     & \tbest 26.74 & \tbest 0.7964 & \tbest 0.2621 &  1.65 & \sbest 11.53 \\
     Ours (base) 
     & \best 24.09 & \best 0.6885 & \best 0.3235 & \best 0.46 & \sbest 6.73 
     & \sbest 21.46 & \best 0.7067 & \tbest 0.3489 & \best 0.28 & \best 4.36 
     & \best 28.04 & \best 0.8336 & \best 0.2285 & \best 0.38 & \best 5.40 \\
    \end{tabular}
    }
    \label{tab:mesh_rendering_metrics}
    \vspace{-0.2cm}
\end{table*}

%% file: figures/tnt_cumulative.tex
\begin{figure}[!ht]
\centering

\begin{tabular}{cc}
    \includegraphics[width=0.42\linewidth]{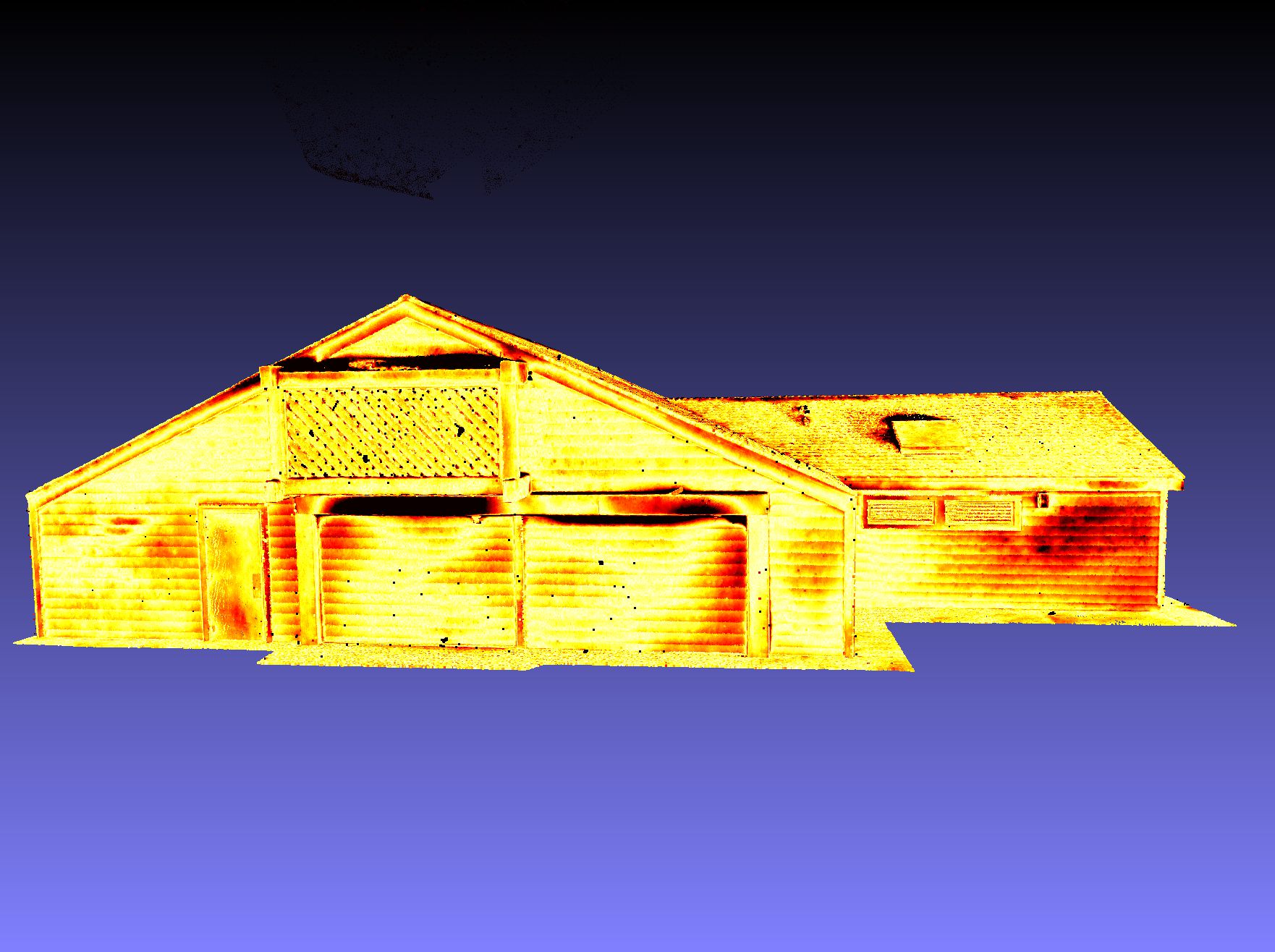} &
    \includegraphics[width=0.42\linewidth]{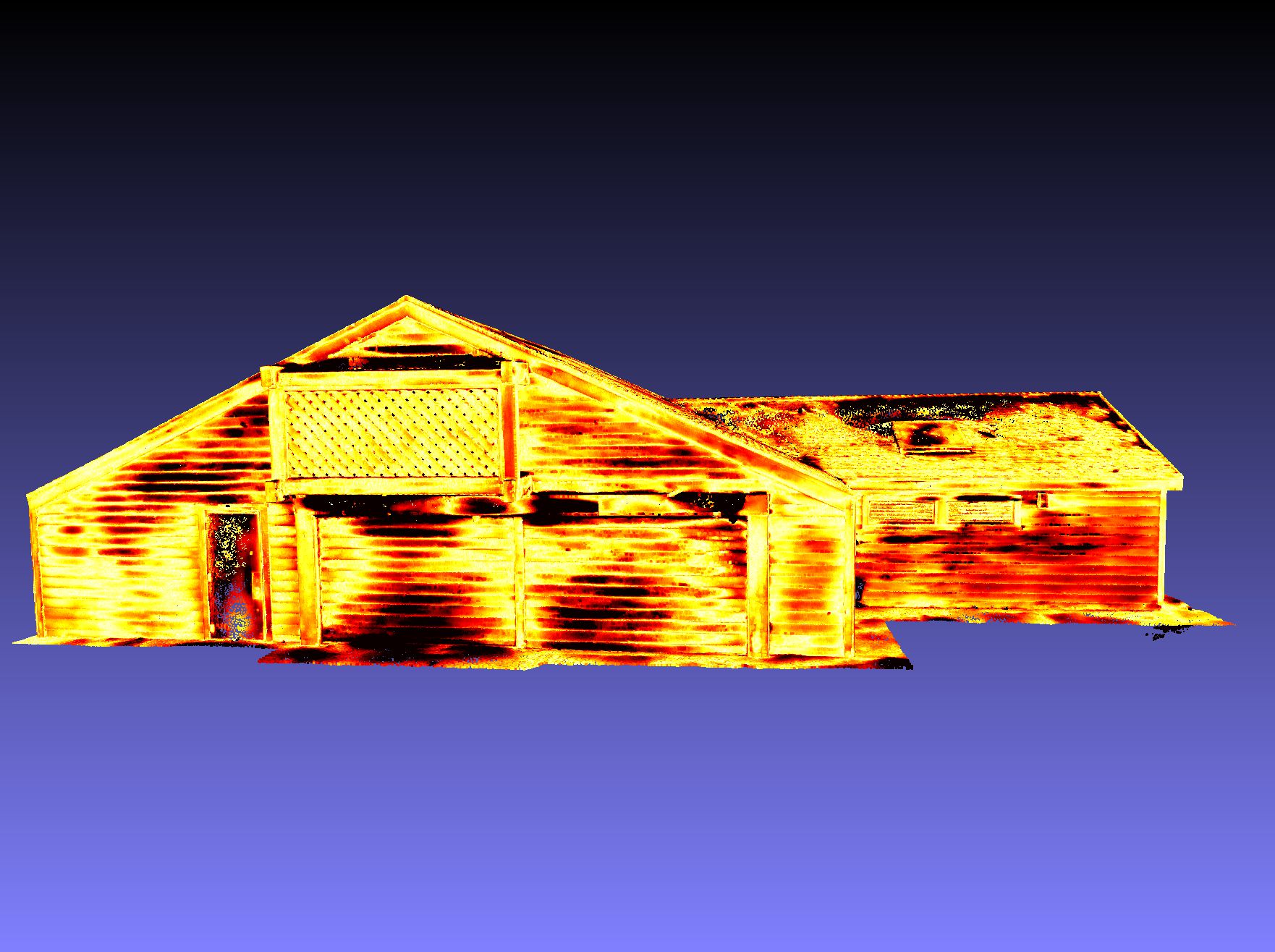} \\
    \small (a) Ours &
    \small (b) GOF \\
\end{tabular}

\includegraphics[width=.6\linewidth]{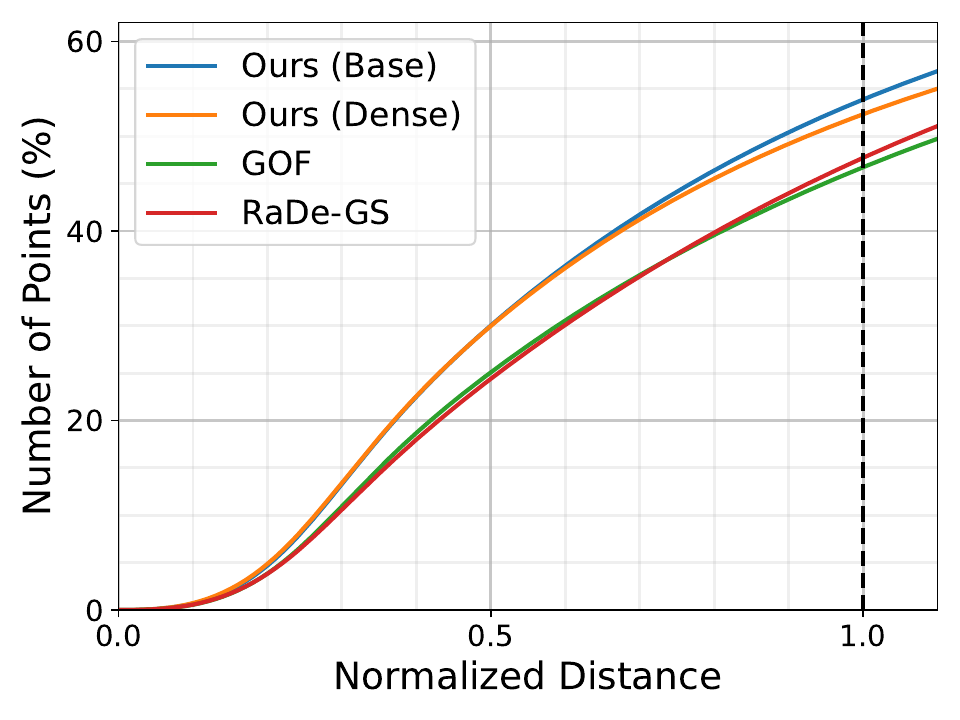}

\small (c) Average cumulative distance distribution on the T\&T dataset

\caption{\textbf{Reconstruction $\rightarrow$ Ground truth Distance Evaluation}. Qualitatively, our reconstructions using the differentiable rasterizer from RaDe-GS (a) are closer to the ground truth point clouds than both GOF (b) and RaDe-GS. The T\&T~\cite{knapitsch2017tanks} evaluation toolkit reports cumulative histograms (c) showing the proportion of reconstructed points within a given distance from the ground truth surface. The dashed line marks the threshold used to compute the F1-score. Our method produces significantly more points close to the true surface (within a normalized distance of $1.0$), demonstrating higher reconstruction accuracy compared to prior approaches.}
\label{fig:tnt_cumulative}
\end{figure} 

%% file: figures/ablation_depth_only_vs_mesh_only.tex
\begin{figure}[!ht]
\centering
\resizebox{\linewidth}{!}{
\begin{tabular}{cc}
    \includegraphics[width=0.49\linewidth]{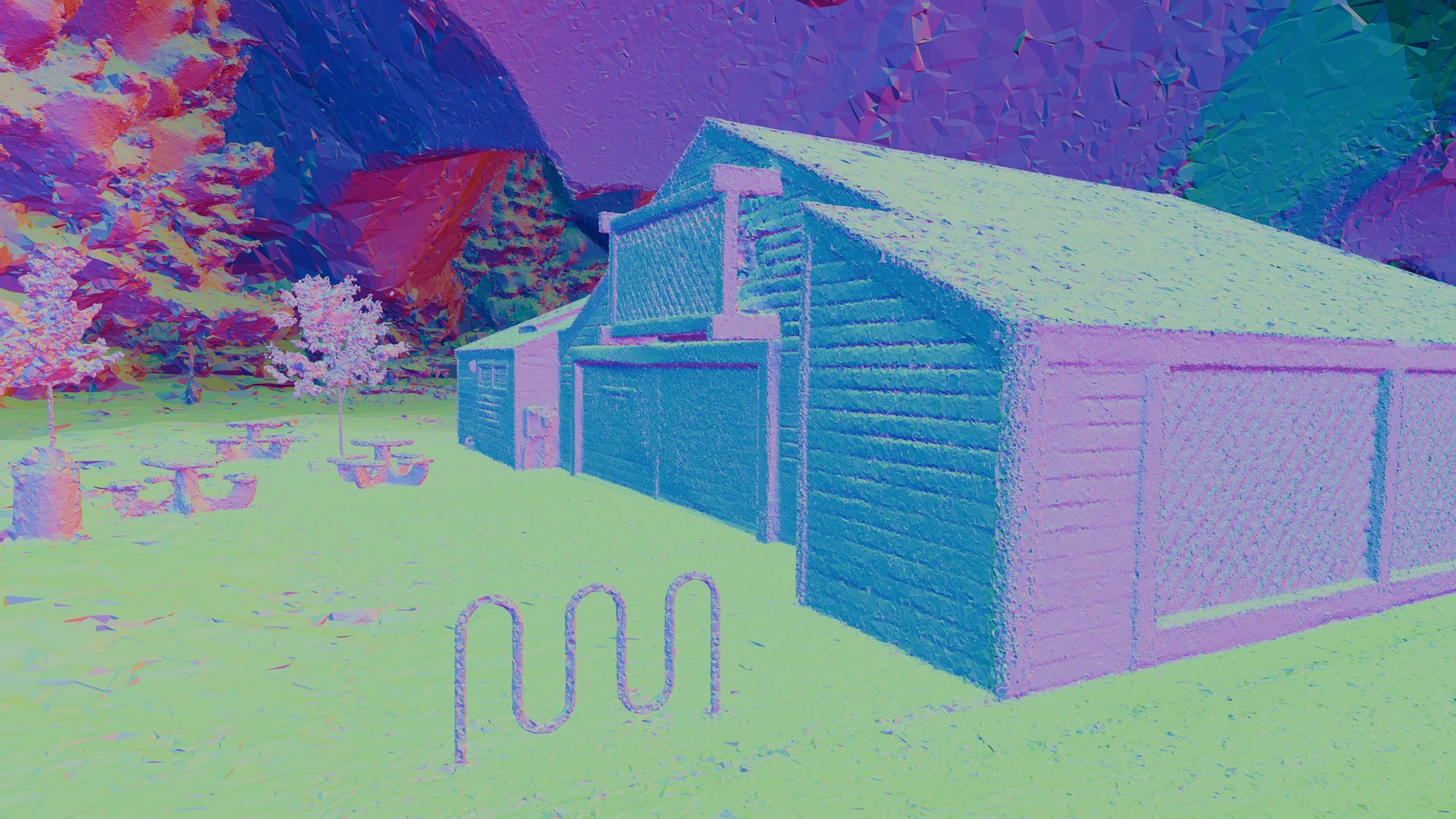} &
    \includegraphics[width=0.49\linewidth]{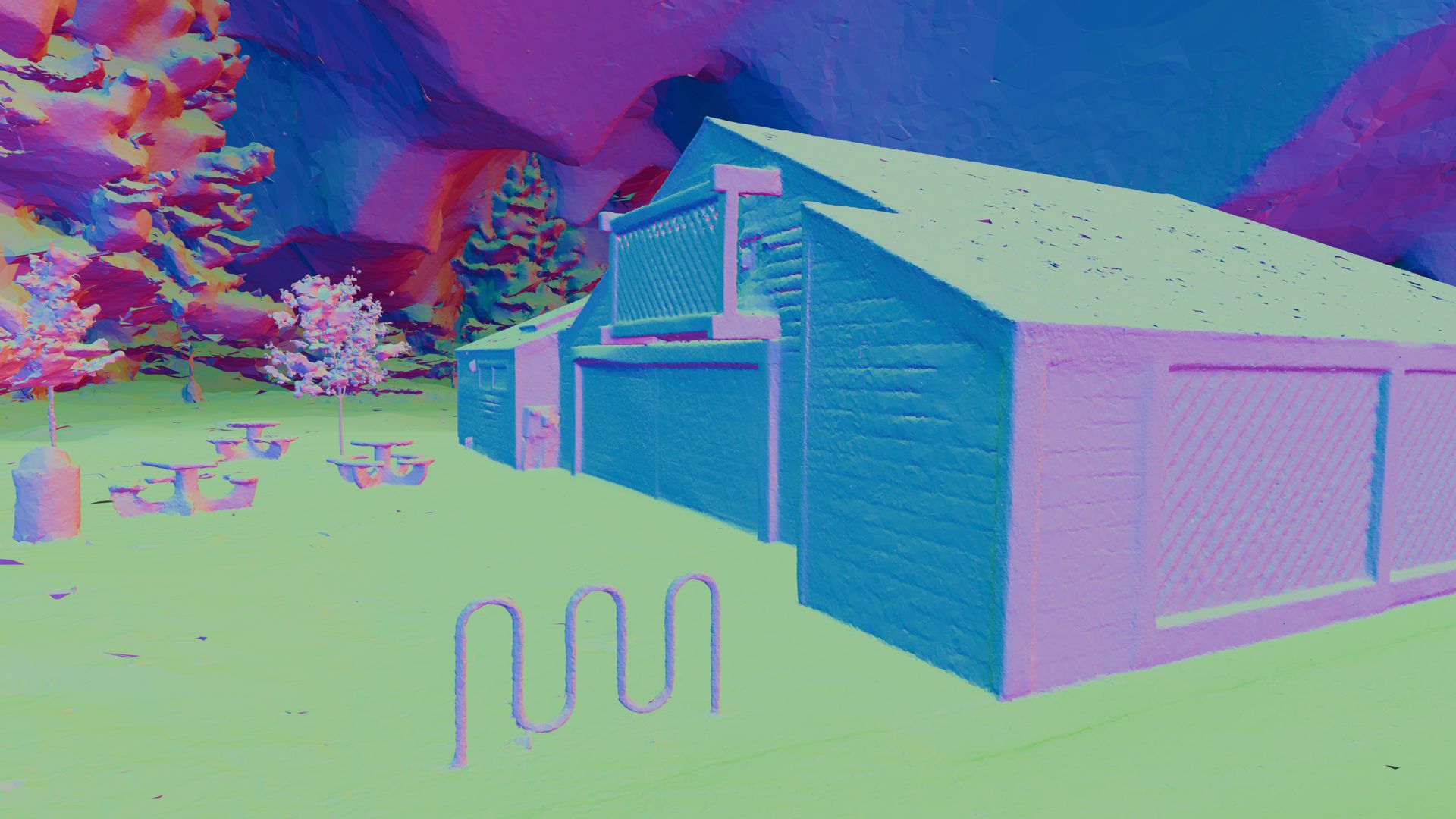} \\
    \small (a) Baseline + $\mathcal{L}_{\text{MD}}$ &
    \small (b) Baseline + $\mathcal{L}_{\text{mesh}}$ \\
\end{tabular}
}
\caption{\textbf{Effect of adding Normal supervision}. Despite showcasing an apparent decrease in performance in our ablation study (see the supplementary material), this figure supports our statement that the combination of depth and normal supervision is essential. The resulting mesh when using only $\mathcal{L}_{\text{MD}}$ is significantly noisier than when supervising with $\mathcal{L}_{\text{mesh}}$.}
\label{fig:depth_only_vs_mesh_only}
\end{figure} 

%% file: tables/ablation_components.tex
\begin{table}[t]
    \small
    \centering
    \caption{\textbf{Ablation study on the Tanks and Temples dataset.} We report only the average $F_1$-score across all scenes to emphasize the impact of each component.}
    \label{tab:ablation_f1_only}
    \begin{tabular}{l|c}
        Method & Average $F_1$-score $\uparrow$  \\
        \hline
        Baseline & 0.41  \\
        + $\calL_{\text{MD}}$ & 0.46 \\
        + $\calL_{\text{mesh}}$ & 0.44  \\
        + $\calL_{\text{mesh}}$ + $\calL_{\text{erosion}}$ & 0.44  \\
        \hline
        \textbf{Ours} (Baseline + $\calL_{\text{mesh}} + \calL_{\text{reg}}$) & \textbf{0.47}  \\
        \textbf{Ours (GOF)} & \textbf{0.49} \\
    \end{tabular}
    
\end{table}

%% file: figures/interior_comparison.tex
\begin{figure}[t]
\centering
\resizebox{\linewidth}{!}{
\begin{tabular}{cc}
    \includegraphics[width=0.49\linewidth]{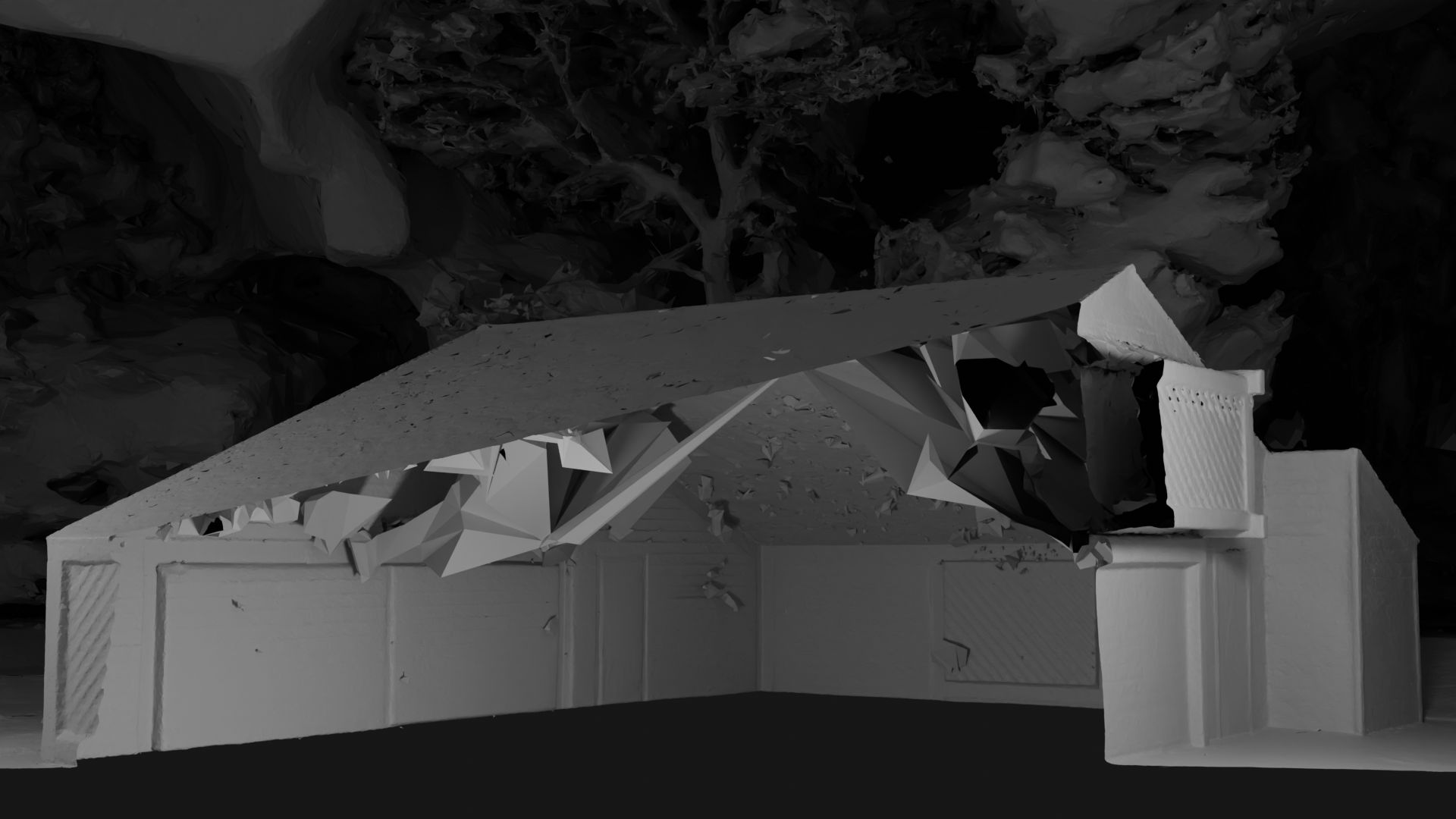} &
    \includegraphics[width=0.49\linewidth]{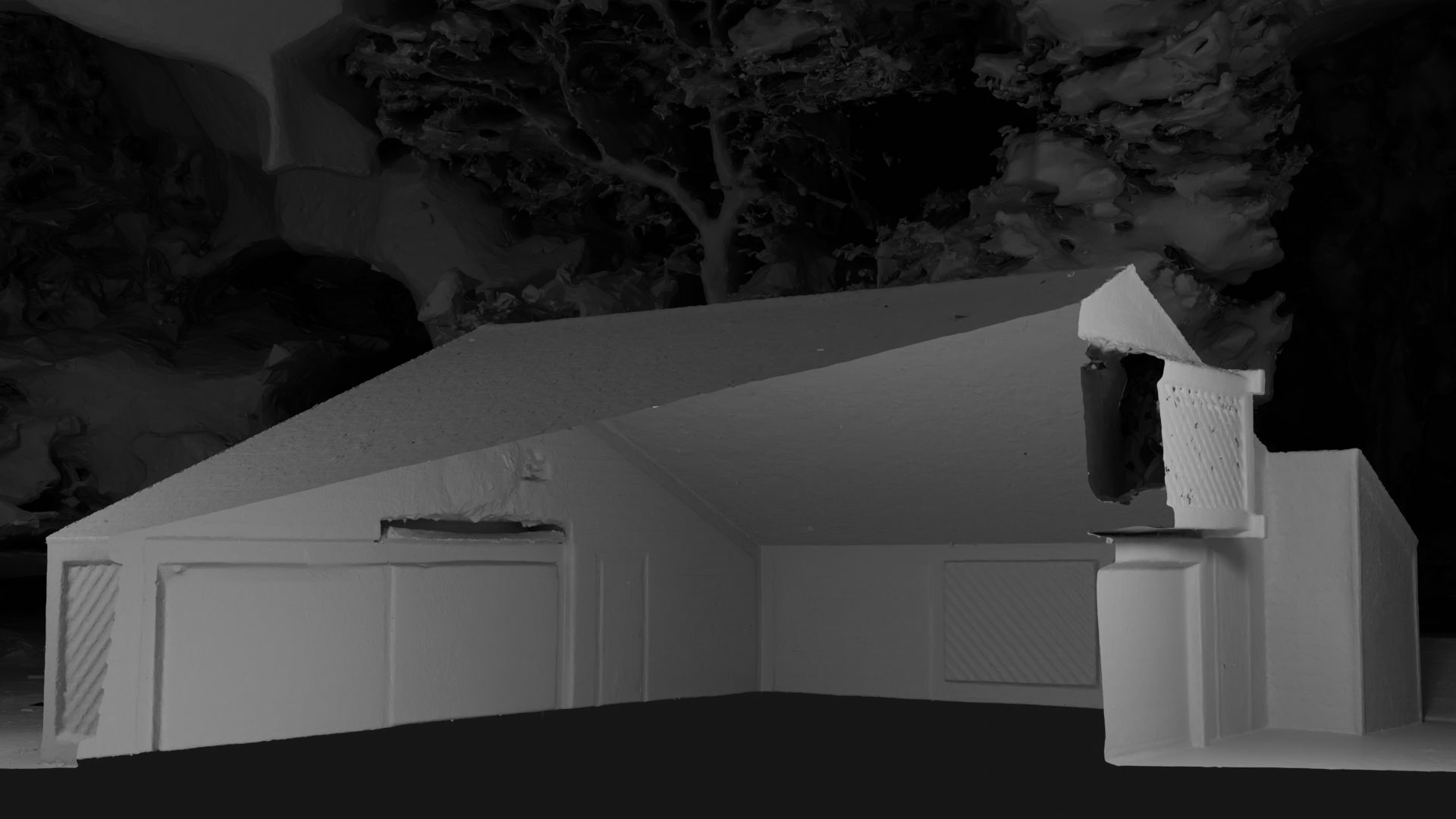} \\
    \small (a) Without interior regularization &
    \small (b) With interior regularization \\
\end{tabular}
}
\caption{\textbf{Comparison of mesh interiors}. We show a cross-section of the reconstructed mesh of the \textit{Barn} scene to reveal its internal structure. Our interior regularization $\calL_{\text{interior}}$ effectively eliminates internal artifacts, producing clean, watertight meshes with empty interiors, which is crucial for downstream applications such as physics simulations and animation.}
\label{fig:interior_comparison}
\end{figure} 

%% file: figures/mesh_extraction_comparison.tex
\begin{figure}[!ht]
\centering
\includegraphics[width=0.9\linewidth]{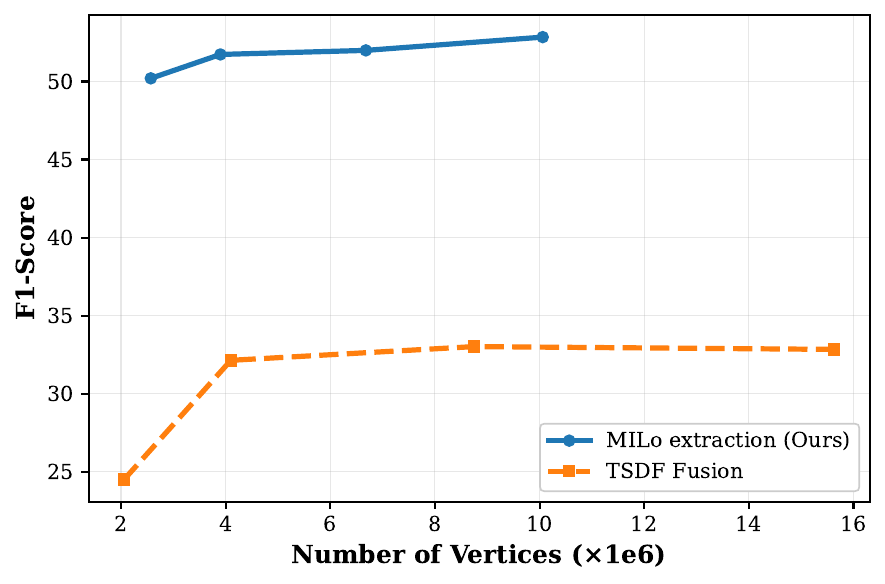}
\caption{
Quantitative evaluation of mesh extraction methods on the Tanks \& Temples dataset across different mesh resolutions. We compare (a) traditional TSDF fusion, which relies on a regular 3D grid, with (b) our mesh extraction method, which learns SDF values at Gaussian pivots.
}
\label{fig:mesh_extraction_comparison}
\end{figure}

%% file: sections/5_conclusions_and_limitations.tex
\vspace{25pt}
\subsection{Limitations and Conclusions}

While our method significantly improves surface reconstruction quality, it still has some limitations. First, the computational cost of extracting and processing the mesh at every iteration increases the training time compared to standard Gaussian Splatting, although it still remains manageable compared to baselines.
Second, the quality of the reconstruction still depends on the initial distribution of Gaussians, which may not be optimal for all scenes.

Conceptually, our framework unlocks many surface-based processing tools for Gaussians that can be applied during training. This integration of mesh-based operations with Gaussian representations opens new possibilities for geometry manipulation, regularization, and enhancement throughout the optimization process, not just as a post-processing step.

Future work could explore novel adaptive sampling strategies for Delaunay sites, and improved initialization techniques. Additionally, extending our approach to \textit{dynamic} scenes and exploring its applications in real-time rendering and interactive applications would be promising directions for future research. 

Further exploration of surface-based processing tools that can be integrated into the training pipeline could also yield significant improvements in reconstruction quality and versatility. 

Finally, through a a mesh-based novel view synthesis evaluation we highlightM the lack of standardized quantitative protocols to evaluate the alignment of the reconstructed surface with the training images. This is true on widely used datasets such as DTU, MipNeRF 360, and Tanks \& Temples. While this approach offers a step toward addressing this gap, it remains an initial effort; future work could focus on refining the evaluation methodology to better capture reconstruction quality in challenging real-world scenes.

\section{Acknowledgements}

Parts of this work were supported by the ERC Consolidator Grant ``VEGA'' (No. 101087347), the ERC Advanced Grant ``explorer'' (No. 101097259), the ANR AI Chair AIGRETTE. Co-funded by the European Union (EU)  (ERC Advanced grant FUNGRAPH No 788065 and ERC Advanced Grant NERPHYS No 101141721). Views and opinions expressed are however those of the author(s) only and do not necessarily reflect those of the EU or the European Research Council. Neither the EU nor the granting authority can be held responsible for them.


%% file: sections/X_supplemental.tex
\appendix
\section*{Supplementary Material}
\label{sec:supp}

\section{Preliminaries}
\label{sec: preliminaries}

In this supplementary material, we start by briefly reviewing the key concepts that form the foundation of our approach: 3D Gaussian Splatting and Delaunay tetrahedralization.

\subsection{3D Gaussian Splatting}

3D Gaussian Splatting (3DGS)~\cite{kerbl3Dgaussians} has emerged as a powerful technique for novel view synthesis, offering high-quality rendering with real-time performance. 
In this representation, a 3D scene is modeled as a collection of 3D Gaussians, each defined by its position~$\mu \in \mathbb{R}^3$, covariance matrix~$\Sigma \in \mathbb{R}^{3 \times 3}$ (typically parameterized by a scaling vector~$s \in \mathbb{R}^3$ and a rotation matrix~$R \in \mathbb{R}^{3 \times 3}$ encoded using a quaternion~$q \in \mathbb{R}^4$), opacity~$\alpha \in (0, 1)$, and appearance attributes (such as spherical harmonics coefficients $c \in \mathbb{R}^{S}$ for view-dependent color). During rendering, these 3D Gaussians are projected onto the image plane as 2D Gaussians and composited in a front-to-back order to produce the final image.

\subsection{Delaunay Triangulation}

Given a set of 3D points, the Delaunay triangulation divides their convex hull into tetrahedra such that no point lies inside the circumsphere of any tetrahedron. This structure has been extensively studied due to its desirable theoretical and practical properties; see~\cite{aurenhammer1991voronoi} for more details.

In MILo, we rely on the Delaunay triangulation as it naturally adapts to the local density of the input point cloud. Specifically, it creates smaller tetrahedra in regions with higher point density and larger ones in sparser regions. This adaptive nature makes it more efficient than uniform grid-based approaches for extracting surfaces from non-uniform point distributions, such as those formed by optimized 3D Gaussians. Furthermore, the regularity of the Delaunay structure facilitates GPU-accelerated mesh extraction via the marching tetrahedra algorithm~\cite{Doi1991marchingtet}, which is the cornerstone of our differentiable mesh extraction process.

\section{Experiments}

\subsection{Implementation Details}

\paragraph{Optimization.} For optimization, we use the Adam optimizer with the same learning rates for the Gaussian parameters as~\cite{zhang2024rade}. We use a learning rate of $0.025$ for the SDF values. We densify Gaussians for 3,000 iterations using aggressive densification~\cite{fang2024mini}, during which we only rely on our photometric loss. We introduce the volumetric rendering regularization loss $\calL_{\text{N}}$ after densification, at iteration 3,000.
We let Gaussians populate the scene and refine their parameters for 5,000 additional iterations, during which we rely on our volumetric rendering loss $\calL_{\text{vol}}$. \diego{After this point we stop the densification and pruning procedures. Thus, our method does not introduce additional sensitivity to initialization beyond what is already present in standard 3DGS pipelines.}

We introduce our mesh extraction pipeline at iteration 8,000. We extract a mesh at every iteration and apply our full loss $\calL$ to our representation. The mesh-in-the-loop optimization continues for an additional 10,000 iterations, for a total of 18,000 iterations.
For our base model, we prune Gaussians at iteration 8,000 with importance-weighted sampling~\cite{fang2024mini} to maintain only the most important Gaussians. For the dense model, we maintain a larger set of Gaussians but compute the importance scores at iteration 8,000 and sample Delaunay sites only from the most important Gaussians. Our weight parameters are~$\lambda_{\text{RGB}} = 0.2$, $\lambda_{\text{N}} = \lambda_{\text{MD}} = \lambda_{\text{MN}} = 0.05$, and~$\lambda_{\text{erosion}} = \lambda_{\text{interior}} = 0.005$.

\paragraph{Delaunay triangulation.} For the Delaunay triangulation, we use the CGAL library's 3D Delaunay triangulation implementation, which provides robust and efficient computation of the tetrahedralization. Although Delaunay triangulation is inherently non-differentiable, this does not pose a problem in our setting, as gradients propagate from the mesh back to the Gaussians via the Gaussian pivots. We found that updating the Delaunay triangulation at every iteration is not necessary for stable optimization; We therefore only update the Delaunay triangulation every 500 iterations.

\paragraph{SDF normalization.} In practice, for more stable optimization, we optimize Truncated SDF values normalized to be in the range $[-1, 1]$ using a tanh function. When introducing our Mesh-in-the-Loop optimization, the initial SDF values are computed using a custom, scalable depth-fusion algorithm that operates directly on our Delaunay sites rather than on a regular grid: For each Delaunay site, we compute the signed difference between the depth map rendered from the Gaussians and the depth of the site, and fuse these distances by averaging across all views. This initialization provides a good starting point for the SDF optimization; While being over-smoothed and missing many fine details, the initial mesh is already a reasonable approximation of the surface encoded by the Gaussians.

\paragraph{Mesh rendering.} For differentiable mesh rendering, we use nvdiffrast~\cite{Laine2020nvdiffrast}, which provides efficient GPU-accelerated rasterization with gradient support. We render the mesh at the same resolution as the Gaussian rendering, using the same camera parameters. To smooth out discontinuities and ensure the propagation of gradients between neighbor pixels, we apply antialiasing smoothing to the depth maps rendered from the mesh.

\subsection{Mesh-Based Novel View Synthesis}

\input{figures/mesh_based_evaluation_qualitative}

We showcase a qualitative comparison of our mesh-based novel view synthesis evaluation in Fig.~\ref{fig:mesh_based_evaluation_qualitative}.

\subsection{Novel View Synthesis}

\input{tables/nvs_metrics}

While our primary focus is on surface reconstruction, we also evaluate the novel view synthesis quality of our optimized Gaussians. Table~\ref{tab:nvs_metrics} presents the PSNR, SSIM, and LPIPS metrics on the MipNeRF~360~\cite{barron22mipnerf360} dataset. Our method maintains competitive rendering quality compared to previous approaches, demonstrating that our mesh-in-the-loop optimization does not compromise the visual fidelity of the Gaussian representation.

%% file: figures/mesh_based_evaluation_qualitative.tex
\begin{figure*}[t]
\centering
\includegraphics[width=\linewidth]{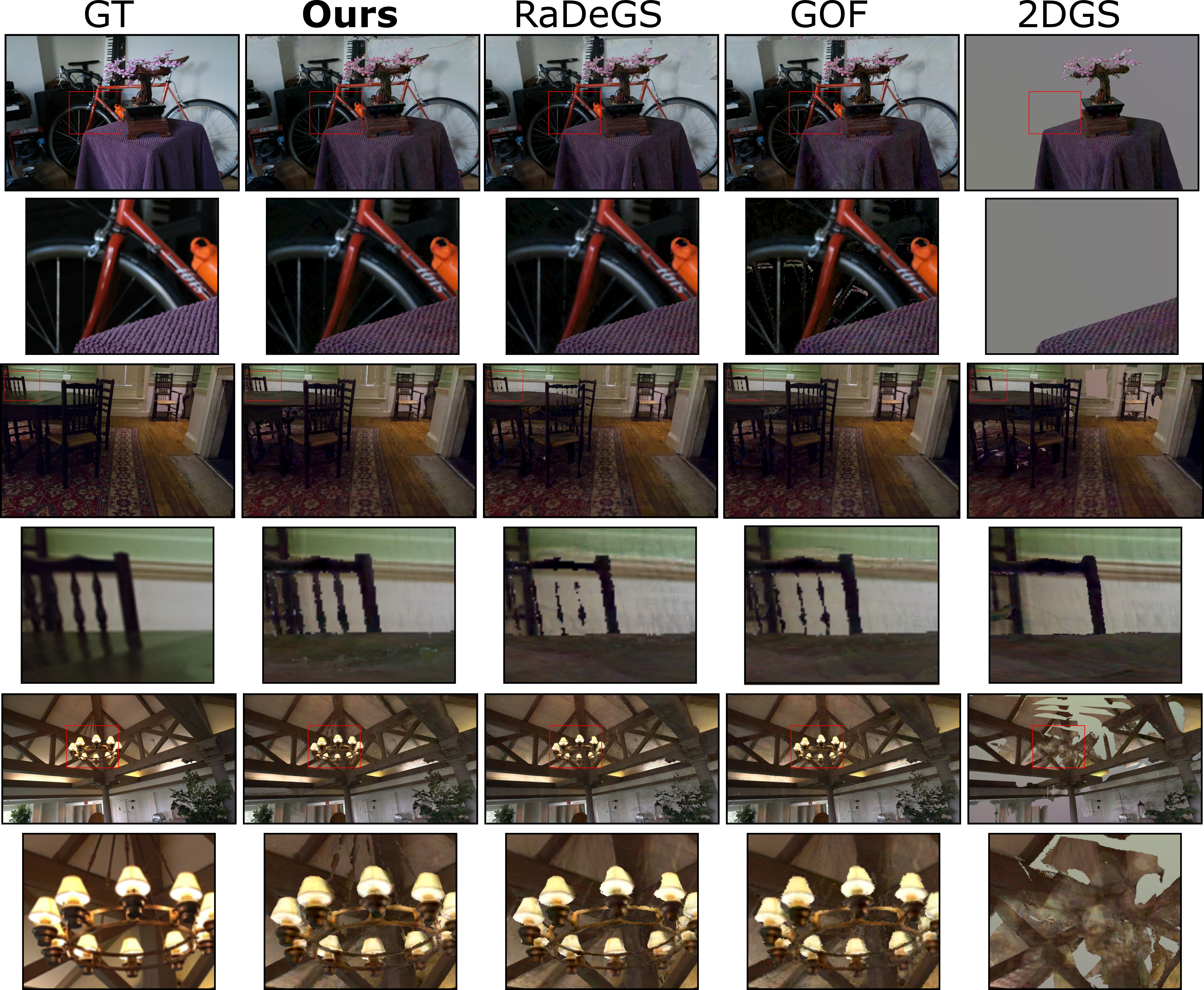}
\caption{\textbf{Our mesh-based evaluation serves as a proxy for quantitatively assessing geometric artifacts, mesh completeness, and background reconstruction. MILo demonstrates superior performance in capturing fine details---such as the intricate chandelier stem and chair spindles---and consistently recovers high-quality background geometry. }}
\label{fig:mesh_based_evaluation_qualitative}
\end{figure*} 

%% file: tables/nvs_metrics.tex
\begin{table}[t]
    \centering
    \caption{\textbf{Quantitative results for novel view synthesis on MipNeRF~360~\cite{barron22mipnerf360}}. We report PSNR, SSIM, and LPIPS. Our method maintains competitive rendering quality while significantly improving surface reconstruction.}
    \vspace{-0.2cm}
    \resizebox{0.99\linewidth}{!}{
    \begin{tabular}{@{}l|ccc|ccc}
     & \multicolumn{3}{c@{}|}{Indoor Scenes} & \multicolumn{3}{c@{}}{Outdoor Scenes}\\ 
     & PSNR $\uparrow$ & SSIM $\uparrow$ & LPIPS $\downarrow$
     & PSNR $\uparrow$ & SSIM $\uparrow$ & LPIPS $\downarrow$ 
     \\ 
     \hline
    3DGS 
    & 30.41 & 0.920 & 0.189
    & 24.64 & 0.731 & 0.234
    \\
    Mip-Splatting 
    & 30.90 & 0.921 & 0.194 
    & 24.65 & 0.729 & 0.245 
    \\
    BakedSDF 
    & 27.06 & 0.836 & 0.258
    & 22.47 & 0.585 & 0.349 
    \\
    SuGaR
    & 29.43 & 0.906 & 0.225
    & 22.93 & 0.629 & 0.356 
    \\
    2DGS 
    & 30.40 & 0.916 & 0.195 
    & 24.34 & 0.717 & 0.246 
    \\
    GOF 
    & \best 30.79 & \tbest 0.924 & \tbest 0.184 
    & \sbest 24.82 & \sbest 0.750 & \sbest 0.202 
    \\
    RaDe-GS 
    & \sbest 30.74 & \sbest 0.928 & \sbest 0.165 
    & \best 25.17 & \best 0.764 & \best 0.199 
    \\
    Ours (base) 
    &  29.96 &  0.920 &  0.191 
    &  24.47 &  0.718 &  0.290 
    \\
    Ours (dense) 
    & \tbest 30.76 & \best 0.934 & \best 0.155 
    & \tbest 24.81 & \tbest 0.744 & \tbest 0.229 
    \\
    \end{tabular}
    }
    \label{tab:nvs_metrics}
    \vspace{-0.2cm}
\end{table}